\documentclass{article}

\usepackage[utf8]{inputenc} 
\usepackage[T1]{fontenc}    
\usepackage{booktabs}       
\usepackage{nicefrac}       
\usepackage{microtype}      
\usepackage{times}             

\usepackage[round]{natbib}
\usepackage{amssymb,amsmath,amsthm,bbm}
\usepackage[margin=1in]{geometry}
\usepackage{verbatim,float,url,dsfont}
\usepackage{graphicx,subfigure,psfrag}
\usepackage{algorithm,algorithmic}
\usepackage{mathtools,enumitem}
\usepackage[colorlinks,citecolor=blue]{hyperref}
\usepackage[utf8]{inputenc} 
\usepackage[T1]{fontenc}    
\usepackage{hyperref}       
\usepackage{url}            
\usepackage{booktabs}       
\usepackage{amsfonts}       
\usepackage{nicefrac}       
\usepackage{microtype}      

\usepackage{times}             
\usepackage{amsthm}

\usepackage{bbm}
\usepackage{verbatim,float,dsfont}
\usepackage{psfrag}
\usepackage{algorithm,algorithmic}
\usepackage{mathtools,enumitem}
\mathtoolsset{showonlyrefs}
\usepackage{tcolorbox}
\usepackage{breqn,lipsum}
\usepackage{thmtools,thm-restate}
\usepackage{graphicx,stackengine}
\usepackage{array}
\usepackage{caption}
\usepackage{boldline}
\usepackage{amssymb}
\usepackage{multirow}

\usepackage{graphicx}
\usepackage[nocomma]{optidef}
\usepackage{mdframed}

\usepackage{tikz}
\usetikzlibrary{shapes,arrows}

\usetikzlibrary{positioning, quotes}

\newtheorem{theorem}{Theorem}
\newtheorem{lemma}[theorem]{Lemma}
\newtheorem{corollary}[theorem]{Corollary}
\newtheorem{proposition}[theorem]{Proposition}
\newtheorem{remark}[theorem]{Remark}
\newtheorem{definition}[theorem]{Definition}

\newtheorem{example}[theorem]{Example}

\def\shownotes{1}  
\ifnum\shownotes=1
\newcommand{\authnote}[2]{$\ll$\textsf{\footnotesize #1 notes: #2}$\gg$}
\else
\newcommand{\authnote}[2]{}
\fi

\usepackage{accents}
\newcommand{\ubar}[1]{\underaccent{\bar}{#1}}

\makeatletter
\newcommand*\rel@kern[1]{\kern#1\dimexpr\macc@kerna}
\newcommand*\widebar[1]{%
  \begingroup
  \def\mathaccent##1##2{%
    \rel@kern{0.8}%
    \overline{\rel@kern{-0.8}\macc@nucleus\rel@kern{0.2}}%
    \rel@kern{-0.2}%
  }%
  \macc@depth\@ne
  \let\math@bgroup\@empty \let\math@egroup\macc@set@skewchar
  \mathsurround\z@ \frozen@everymath{\mathgroup\macc@group\relax}%
  \macc@set@skewchar\relax
  \let\mathaccentV\macc@nested@a
  \macc@nested@a\relax111{#1}%
  \endgroup
}
\makeatother

\newcommand{\argmin}{\mathop{\mathrm{argmin}}}

\makeatletter
\newcommand{\setword}[2]{%
  \phantomsection
  #1\def\@currentlabel{\unexpanded{#1}}\label{#2}%
}
\makeatother

\def\E{\mathbb{E}}

\def\sign{\mathrm{sign}}

\def\cA{\mathcal{A}}
\def\cB{\mathcal{B}}
\def\cD{\mathcal{D}}

\def\cF{\mathcal{F}}

\def\cH{\mathcal{H}}

\def\cP{\mathcal{P}}
\def\cS{\mathcal{S}}
\def\cT{\mathcal{T}}
\def\cW{\mathcal{W}}

\def\bs{\ensuremath\boldsymbol}
\def\code#1{\texttt{#1}}

\newcommand{\ALG}{\textsc{FLH}}

\newcommand{\grad}{\nabla}
\newcommand{\lamda}{\lambda}

\title{Optimal Dynamic Regret in Exp-Concave Online Learning}  

\author{Dheeraj Baby \\dheeraj@ucsb.edu \and Yu-Xiang Wang \\yuxiangw@cs.ucsb.edu}

\date{Dept. of Computer Science \\ UC Santa Barbara}

\begin{document}

\maketitle

\begin{abstract}%
We consider the problem of the \citet{zinkevich2003online}-style dynamic regret  minimization in online learning  with \emph{exp-concave} losses. We show that whenever improper learning is allowed, a Strongly Adaptive online learner achieves the dynamic regret of $\tilde O^*(n^{1/3}C_n^{2/3} \vee 1)$ where $C_n$ is the \emph{total variation} (a.k.a. \emph{path length}) of the an arbitrary sequence of comparators that may not be known to the learner ahead of time. Achieving this rate was highly nontrivial even for square losses in 1D where the best known upper bound was $O(\sqrt{nC_n} \vee \log n)$ \citep{yuan2019dynamic}.
Our new proof techniques make elegant use of the intricate structures of the primal and dual variables imposed by the KKT conditions and could be of independent interest. Finally, we apply our results to the classical statistical problem of  \emph{locally adaptive non-parametric regression} \citep{mammen1991,donoho1998minimax} and obtain a stronger and more flexible algorithm that do not require any statistical assumptions or any hyperparameter tuning.
\end{abstract}

\section{Introduction} \label{sec:intro}
We consider a generic online learning framework which is modelled as an  interactive $n$ step game between a learner and adversary. At each time step $t$, the learner predicts a $\bs p_t \in \cD \subseteq \mathbb{R}^d$. Then the adversary reveals a loss function $f_t:\mathbb{R}^d \rightarrow \mathbb{R}$. The objective of the learner is to minimise its regret against a predefined set of strategies $\cW$ that is known to the learner before the start of the game.  We call a learning algorithm to be \emph{proper} when $\cD = \cW$. Further when $\cD = \cW$ are convex sets and the losses $f_t$ are convex in $\cD$, the generic learning framework reduces to the one studied in Online Convex Optimization (OCO) \citep{hazan2016introduction}. On the other hand, we call the learning algorithm to be \emph{improper} when $\cD \supset \cW$.
A commonly used metric to measure the performance of the learner is its \emph{static} regret defined as
\begin{align}
    R_n
    &= \sum_{t=1}^{n} f_t(\bs p_t) - \inf_{\bs w \in \cW} \sum_{t=1}^{n} f_t(\bs w).
\end{align}
A sub-linear static regret implies that the average loss incurred by the learner converges to that of the best comparator strategy in hindsight. 

A canonical example of an improper algorithm can be found in an online linear regression setting where $f_t(\bs u) = (y_t - \bs x_t^T \bs u)^2$ with $|y_t| \le 1,\: \| \bs x_t\| _2 \le 1$ and we are interested in controlling the static regret against against a set of linear predictors with bounded norm, $\cW = \{\bs w \in \mathbb{R}^d: \| \bs w\|_2 \le 1 \}$. One popular learning algorithm in this framework is the Vovk-Azoury-Warmuth (VAW) forecaster \citep{Vovk1997CompetitiveOL,Azoury2004RelativeLB} (or see Section 11.8 in \citep{BianchiBook2006}). The VAW forecaster attains an $O(d \log n)$ static regret against $\cW$. However predictions of VAW at time $t$ denoted by $\bs u_t$ may not necessarily satisfy $\| \bs u_t\|_2 \le 1$ hence making it an improper algorithm.

The notion of static regret is not befitting for non-stationary environments -- such as financial markets -- where it could be inappropriate to compete against a fixed comparator due to the changes in the dynamics of the environment. The work of \citep{zinkevich2003online} introduces the notion of \emph{dynamic} regret defined as
\begin{equation}
    R^n_{\bs w_1,\ldots,\bs w_n}
    := \sum_{t=1}^n f_t(\bs p_t) - f_t(\bs w_t), \label{eq:d-regret}
\end{equation}
for \emph{any} sequence of comparators $\bs w_t$ in $\cW$. The dynamic regret bounds are usually expressed in literature as a function of number of time steps and some path variation metric that captures the degree of non-stationarity in the comparator sequence. In this paper, we study the following path variation:
$$
    TV(\bs w_1,\ldots,\bs w_n) := \sum_{t=2}^n \| \bs w_t - \bs w_{t-1}\|_1.
$$
The maximum dynamic regret against all comparator sequences whose path variation is bounded by a number $C_n$ can then be defined as
\begin{align}
    R_n(C_n)
    &:= \sup_{\substack{\bs w_1,\ldots,\bs w_n \\ TV(\bs w_1,\ldots, \bs w_n) \le C_n}}  R^n_{\bs w_1,\ldots,\bs w_n}.
\end{align}

There is a complementary body of work on Strongly Adaptive (SA) algorithms \citep{daniely2015strongly} where the static regret in \emph{any} sub-interval of $[n] := \{1,\ldots,n \}$ is controlled (see Section \ref{sec:lit} for a review). Hence SA algorithms have the nice property of being globally and locally optimal. The work of \citep{zhang2018dynamic} exploits this property of SA algorithms to control their dynamic regret  in terms of a variational metric that measures how much the losses $f_t$ change over time. In particular, whenever the losses have extra curvature properties such as strong convexity or exp-concavity, they show that one can get fast dynamic regret rates. However, it was unclear if SA methods can lead to optimal dynamic regret guarantees in terms of the path length of the comparator sequence --- 
an open question raised in \citep{zhang2018dynamic}.

The works of \citep{zhang2018adaptive} and \citep{yuan2019dynamic} attains a dynamic regret of $O^*(\sqrt{n(1+C_n)})$ and $O^*(\sqrt{nC_n} \vee \log n)$ respectively, where $O^*(\cdot)$ hides dependence on the dimension and $(a \vee b) = \max \{ a, b\}$. However, we show a lower bound of $\Omega^*(n^{1/3}C_n^{2/3} \vee \log n)$ in Proposition \ref{prop:lb-sq}  applicable to the case when losses are strongly convex / exp-concave. Hence, there is a large gap between this lower bound and existing upper bounds. In this work, we show that whenever improper learning is allowed and when the loss functions are strongly convex / exp-concave, one can leverage SA algorithms to attain the \emph{sharp} rate of  $\tilde O^*(n^{1/3}C_n^{2/3} \vee \log n)$ for $R_n(C_n)$ where $\tilde O^*(\cdot)$ hides dependence in the dimension and factors of $\log n$ (see section \ref{sec:ec} for formal statements and complete list of assumptions). Further, the SA algorithms need not require the apriori knowledge of $C_n$ to attain this rate.  

As a concrete use case, we show that our results have interesting implications to the problem of \emph{online} Total Variation (TV) denoising. The offline version of TV-denoising problem has seen many influential applications in the signal processing community. For example, algorithms that use TV-regularization has been deployed in every cellphone, digital camera and medical imaging devices (we refer readers to the book \citep{chambolle2010introduction} and the references therein) as well as other tasks beyond the context of images such as change-point detection, semisupervised learning and graph partitioning.

We proceed to formally introduce the non-paramteric regression problem behind TV-denoising. Define a non-parametric class of TV bounded sequences as
$$\mathcal{TV}(C_n)
    := \left \{(w_1,\ldots,w_n) : \sum_{t=2}^n |w_t - w_{t-1}| \le C_n  \right \},$$
    where $\sum_{t=2}^n |w_t - w_{t-1}| $ is termed as the TV of the sequence $w_{1:n} := (w_1,\ldots,w_n)$. In the offline TV-denoising problem we are given $n$ observations of the form $y_t = w_t + \epsilon_t$ where $\epsilon_t$ are iid zero mean subgaussian noise, $t \in [n]$ and $w_{1:n}$ is an unknown sequence in $\mathcal{TV}(C_n)$. We are interested in coming up with estimates $\hat w_t$ such that $R^{\mathcal{TV}}(C_n) := \E \left [\sum_{t=1}^n (\hat w_t - w_t)^2 \right]$ is controlled. Several non-parametric regression algorithms such as Trend Filtering \citep{tibshirani2014adaptive} are known to achieve a near minimax optimal rate of $\tilde O(n^{1/3}C_n^{2/3})$ for $R^{\mathcal{TV}}(C_n)$ where $\tilde O(\cdot)$ hides dependence on factors of $\log n$.

We can instantiate an online version of the above non-parametric regression problem behind TV-denoising into our learning framework with slight modifications. We consider a TV class with bounded sequences 
\begin{equation}
\mathcal{TV}^B(C_n) := \left \{w_{1:n} : \sum_{t=2}^n |w_t - w_{t-1}| \le C_n, \: |w_t| \le B \: \forall t \in [n] \right \}. \label{eq:tvb}
\end{equation}
When viewed through our online learning framework, we take $f_t(x) = (y_t - x)^2$ where $|y_t| \le B$, $\cD = \cW = [-B,B]$. Labels $y_{1:n}$ is a fixed sequence in contrast to the stochastic noise setting discussed earlier, 
and we are hoping to compete with the best approximation from sequences in $\mathcal{TV}^{B}(C_n)$ for all $C_n\geq 0$ at the same time. We remark that to compete with the entire $\mathcal{TV}(C_n)$ class it is sufficient to compete with $\mathcal{TV}^B(C_n)$ due to the property $|y_t| \le B$.
We show in Section \ref{sec:sq} that by using appropriate SA algorithms, one can attain a dynamic regret of $R_n(C_n) = \tilde O(n^{1/3}C_n^{2/3})$. This in turn implies the minimax estimation rate in the iid stochastic setting (see Appendix \ref{app:lit} for details). Further our results have the added advantage of providing an oracle inequality. We conclude this section by summarizing our key contributions below.

\begin{itemize}
    \item We show that Follow-the-Leading-History (FLH) algorithm \citep{hazan2007adaptive} with Follow The Leader (FTL) as base learners can achieve the optimal minimax regret (modulo $\log n$ factors) of $\tilde O(n^{1/3}C_n^{2/3} B^{4/3} \vee B^2\log n)$ for the problem of online non-parametric regression with TV bounded sequences -- $\mathcal{TV}^B(C_n)$ -- as the reference class. The policy is \emph{adaptive} to the TV budget $C_n$. Further, we demonstrate that the same policy is minimax optimal for smoother non-parametric sequence classes such as Sobolev class or Holder class.
    
    \item When improper learning is allowed and when the loss functions revealed by the adversary are exp-concave, strongly smooth and Lipschitz on a box that encloses the set of comparators $\cW$, (see Section \ref{sec:ec}) we show that FLH with ONS as base learners attains a dynamic regret of $\tilde O\left (d^{3.5}(n^{1/3}C_n^{2/3} \vee 1) \right)$ when $C_n \ge 1/n$ and $O(d^{1.5} \log n)$ otherwise, without prior knowledge of $C_n$ -- the path variation of the \emph{comparator sequence}. This rate is shown to be minimax optimal modulo polynomial factors of $\log n$ and $d$.
  
  \item The proof of the regret bound is facilitated by exploiting a number of distinct structures of primal and dual variables in KKT conditions of the optimization problem solved by the offline oracle. We believe that this style of analysis can be useful in bounding the regret of online algorithms in a broader context.
  
\end{itemize}


\section{Related Work} \label{sec:lit}
We begin by recalling works that are most relevant to our setting. We reserve the term $\emph{OCO setting}$ when $\cW = \cD$ and loss functions are convex in $\cD$. 

For an arbitrary comparator sequence in $\cW$ denoted by $\bs w_{1:n} := (\bs w_1,\ldots,\bs w_n), $\citep{zinkevich2003online} introduces a path variational defined as
\begin{align}
    P_n(\bs w_1, \ldots, \bs w_n) = \sum_{t=1}^{n} \|\bs w_t - \bs w_{t-1} \|_2. \label{eq:pn}
\end{align}
They show that in the OCO setting, the Online Gradient Descent (OGD) algorithm can attain a dynamic regret (Eq.\eqref{eq:d-regret}) of $O(\sqrt{n}(1+P_n))$,
but if $P_n$ is known\footnote{In a sense that we are to only compete with sequences with path length $\leq P_n$, rather than \emph{simultaneously} competing with all sequences $P_n > 0$.}, $O(\sqrt{n(1+P_n)})$ can be achieved by simply increasing the learning rate appropriately.  
By hedging over a collection of OGD algorithms defined by exponential grid of step sizes, \citep{zhang2018adaptive} proposes an algorithm that achieves a faster rate of $O(\sqrt{n(1+P_n)})$ which is shown to be minimax optimal when the loss functions are convex. \citep{yuan2019dynamic} proposes strategies that can attain regret rates of $O(\sqrt{nP_n} \vee \log n)$ and $O(\sqrt{dnP_n} \vee d\log n)$ for strongly convex and exp-concave losses respectively. However, this regret rate is only optimal when $P_n$ approaches $n$ or $P_n = O(1/n)$.

\citep{besbes2015non} introduces the functional variation defined as
\begin{align}
    D_n
    &:= \sum_{t=2}^{n} \max_{\bs w \in \cW} |f_t(\bs w) - f_{t-1}(\bs w)|. \label{eq:f-var}
\end{align}
They show that by using a restarted variant of OGD, one can attain the dynamic regret rate of $O(n^{2/3}D_n^{1/3})$ and $\tilde O(\sqrt{nD_n})$ for convex and strongly convex losses respectively using noisy-gradient feedback.
This setting is incompatible to ours as it exploits \emph{smoothness} in $f_1,...,f_n$ while we allow $f_1,...,f_n$ to be arbitrary. Moreover, they need to know $D_n$.

There is a parallel line of work \citep{hazan2007adaptive,daniely2015strongly,koolen2016specialist} that focuses on controlling the static regret in any sub-interval of $[n]$ . In particular, \citep{daniely2015strongly} proposes the notion of Strongly Adaptive algorithms. An algorithm is said to be \emph{Strongly Adaptive} (SA) if for every continuous interval $I \subseteq [n]$, the static regret incurred by the algorithm is $O(\text{poly}(\log n) R^*(|I|))$ where $R^*(|I|)$ is the value of minimax static regret incurred in an interval of length $|I|$. In this viewpoint, the algorithms proposed by \citep{hazan2007adaptive} for strongly convex / exp-concave losses are in fact Strongly Adaptive.

\citep{zhang2018dynamic} shows that SA methods enjoys a dynamic regret of $\tilde O(n^{2/3}D_n^{1/3})$ for convex functions and $\tilde O(\sqrt{nD_n})$ and $\tilde O(\sqrt{dnD_n})$ for strongly convex and exp-concave losses respectively without prior knowledge of $D_n$. We refer the reader to Appendix \ref{app:lit} for a discussion on various other dynamic regret minimization strategies such as \citep{jadbabaie2015online,yang2016tracking,Mokhtari2016OnlineOI,  chen2018non,Zhao2020DynamicRO}.

The setting of learning with squared error losses we consider in Section \ref{sec:tv} can be regarded as an online version of the batch Total Variation denoising problem. The corresponding offline problem has been studied extensively in the non-parametric regression literature. Many algorithms such as Wavelet Smoothing \citep{donoho1998minimax}, Locally Adaptive Regression Splines \citep{vandegeer1990} and  Trend Filtering \citep{l1tf,tibshirani2014adaptive,wang2014falling, graphtf,  guntuboyina2018constrainedTF} have been shown to achieve the optimal minimax rates of $\tilde O(n^{1/3}C_n^{2/3})$ under squared error loss where $n$ is the number of samples and $C_n$ is the TV of the ground truth. All of these estimators have a key property of \emph{local adaptivity} where the estimators are able to detect abrupt local fluctuations in the ground truth signal and adjust the amount of smoothing to be applied which is essential for optimally estimating TV bounded sequences that can exhibit spatially in-homogeneous degree of smoothness.

\citep{arrows2019,baby2021TVDenoise} studies the problem of estimating TV bounded sequences in an online stochastic optimization framework. They assume that the labels revealed by the adversary is the noisy realization of a ground truth sequence that belongs to a $TV(C_n)$ ball. However, the absence of such statistical assumptions on revealed labels in our setting makes the problem significantly more challenging. Interestingly, a lower bound from \citep{arrows2019} implies that the meta-hedge algorithm of \citep{zhang2018adaptive} requires $\Omega(\sqrt{nP_n})$ dynamic regret even if the loss functions are strongly convex, despite the fact that OGD achieves $O(\log n)$ static regret. Extension to higher order TV classes are considered in \citep{baby2020higherTV}.

We refer the reader to Appendix \ref{app:lit} for an elaborate description on how our TV-denoising framework fits under the umbrella of online non-parametric regression framework developed by \citep{rakhlin2014online} and others
\citep{gaillard2015chaining,koolen2015minimax,kotlowski2016}.

\section{Performance guarantees for squared error losses} \label{sec:tv} \label{sec:sq}
In this section, we focus on the online TV-denoising problem which is a special case of our online learning framework with squared error losses as discussed in Section \ref{sec:intro}. This will help to build the intuitions behind the analysis for general exp-concave losses as well. All unspecified proofs of this section are deferred to Appendix \ref{app:sq}.  We consider the following interaction protocol.

\begin{itemize}
    \item At time $t \in [n]$ learner predicts $x_t \in \cD = [-B,B]$.
    \item Adversary reveals a label $y_t \in [-B,B]$.
    \item Learner suffers loss $(y_t - x_t)^2$.
\end{itemize}

We define the comparator class as the set of TV bounded sequences that takes values in $\cW = [-B,B]$ as in Eq.\eqref{eq:tvb}. The performance of the learner is measured using dynamic regret against the sequences that belongs to $\mathcal{TV}^B(C_n)$, for all $C_n > 0$ simultaneously.

The main SA method that we will be relying on throughout this paper is the \ALG{} algorithm from \citep{hazan2007adaptive}. We provide a description of this algorithm in Appendix \ref{app:prelim} for completeness. We have the following regret guarantee for FLH with Follow-the-Leader (FTL) as base learners (in this case, FTL is equivalent to simple online averaging).

\begin{theorem} \label{thm:main-sq}
Let $x_t$ be the prediction at time $t$ of \ALG{} with learning rate $\zeta = 1/(8B^2)$ and base learners as FTL. Then for any compararator $(w_1,\ldots,w_n) \in \mathcal{TV}^B(C_n)$
\begin{align}
    \sum_{t=1}^{n} (y_t - x_t)^2 - (y_t - w_t)^2
    &= \tilde O \left( n^{1/3}C_n^{2/3}B^{4/3} \vee B^2 \right),
\end{align}
where the labels obey $|y_t| \le B$,\ $\tilde O(\cdot)$ hides dependence on logarithmic factors of horizon $n$ and $a \vee b := \max \{a, b \}$.
\end{theorem}
\begin{remark}[Adaptivity to $C_n$ and (non-stochastic) oracle inequality]\label{rmk:oracle_inequality}
 We remark that FLH-FTL does not require $C_n$ as an input 
 thus Theorem~\ref{thm:main-sq} implies the following oracle inequality
 $$
 \sum_{t=1}^{n} (y_t - x_t)^2 \leq \min_{w_1,...,w_n} \sum_{t=1}^n(y_t - w_t)^2 +  \tilde O \left( n^{1/3}\mathrm{TV}(w_{1:n})^{2/3} B^{4/3} \vee B^2 \right).
 $$
 Such result is not known for any algorithm even in the offline case when $y_1,...,y_n$ is known. Notice that $w_t$ does not need to be constrained because $-B\leq y_t\leq B$. 
 
 The strongest oracle inequality for TV-denoising to our knowledge is that of \citep{guntuboyina2018constrainedTF,ortelli2019prediction}, which shows that the fused-lasso estimator with tuning parameter $\lambda$ obeys 
  $
 \sum_{t=1}^{n} (y_t - x_t)^2 \leq \min_{w_1,...,w_n} \sum_{t=1}^n(y_t - w_t)^2 +   O \left( \lambda \mathrm{TV}(w_{1:n})\right),
 $
 under additional stochastic assumptions of $y_t$. Our results eliminate the need to choose hyperparameter $\lambda$ all together and achieve the same rate achievable by the optimal choice of $\lambda$.
\end{remark}
For the sake of clarity we next present the strategy we adopt for proving Theorem \ref{thm:main-sq}. We also highlight the main technical challenges that are needed to be overcome along the way. This is followed by some useful lemmas and proof of the main theorem in Section \ref{sec:proof-main-sq}.
\subsection{Proof strategy for Theorem \ref{thm:main-sq}}
Let $u_1,\ldots,u_n$ be the \emph{offline optimal} sequence (see Lemma \ref{lem:kkt-sq}) in $\mathcal{TV}^B(C_n)$ which attains the minimum cumulative squared error loss. Note that this offline optimal can depend on the entire sequence of labels $y_1,\ldots,y_n$ chosen by the adversary. 

Consider a partitioning of $[n]$ into $M$ sub-intervals $\{[i_s,i_t]\}_{i=1}^M$. We will also use the number $i$ to refer to the interval $[i_s,i_t]$. For the interval $i$, define the quantities: $n_i = i_t-i_s+1$, $\bar{y}_i = \frac{1}{n_i} \sum_{j=i_s}^{i_t} y_j$, $\bar{u}_i = \frac{1}{n_i} \sum_{j=i_s}^{i_t} u_j$.

We start by the following regret decomposition.
{\small
\begin{align}
    R_n
    &= \sum_{i=1}^{M}\underbrace{\sum_{j=i_s}^{i_t} (x_j - y_j)^2 - (y_j - \bar{y}_i)^2}_{T_{1,i}} + \sum_{i=1}^{M}\underbrace{\sum_{j=i_s}^{i_t} (y_j - \bar{y}_i)^2 - (y_j - \bar{u}_i)^2}_{T_{2,i}} + \sum_{i=1}^{M}\underbrace{\sum_{j=i_s}^{i_t} (y_j - \bar{u}_i)^2 - (y_j - u_j)^2}_{T_{3,i}} \label{eq:reg-sq}
\end{align}
}

Now the task of bounding $R_n$ reduces to bounding $T_{1,i},T_{2,i},T_{3,i}$ for each bin and adding them up across all $M$ bins. Let $C_i$ be the TV within bin $i$ incurred by the offline optimal. In Lemma \ref{lem:part-sq}, we exhibit a partitioning $\cP$ of $[n]$ into $M = O(n^{1/3}C_n^{2/3}B^{-2/3})$ bins such that $C_i  \le B/\sqrt{n_i}$ for each bin. 

Due to strong adaptivity of FLH, the term $T_{1,i} = O(B^2\log n)$ since it is the static regret against the fixed comparator $\bar{y}_i$. Hence adding them across all bins in the partition $\cP$ yields $\sum_{i=1}^{M} T_{1,i} = \tilde O(n^{1/3}C_n^{2/3}B^{4/3})$.

By exploiting the KKT conditions satisfied by the offline optimal and using strong smoothness, we show in Lemma \ref{lem:t3-sq} that $T_{3,i}$ can be at-most $O(n_iC_i^2 + \lambda C_i)$ in general. Here $\lambda \ge 0$ is the optimal dual variable arising from the KKT conditions (Lemma \ref{lem:kkt-sq}). 
Since $C_i = O(B/\sqrt{n_i})$ for bins in the partition $\cP$, we have $n_iC_i^2 = O(B^2)$. However, it is not possible to bound $\lambda C_i = O(1)$ since $\lambda$ can be even $\Theta(n)$ in some cases (See Example \ref{ex} in Appendix \ref{app:sq}). 

This is where the term $T_{2,i}$ plays a crucial role. Note that since $\bar{y}_i$ is the minimizer of $g(x) = \sum_{j=i_s}^{i_t} (y_j - x)^2$, we conclude that $T_{2,i} \le 0$. For simplicity of exposition, let's assume that $T_{2,i} < 0$, deferring formal arguments for the general case to Section \ref{sec:proof-main-sq}. We show that this negative term diminishes the $\lambda C_i$ arising from the bound on $T_{3,i}$ to a quantity that is $O(1)$. Specifically, $T_{2,i} + T_{3,i} = O(B^2)$ even though individually $|T_{2,i}|,|T_{3,i}|$ can be \emph{very large}. The desired regret bound now follows by summing it across all $M = O(n^{1/3}C_n^{2/3}B^{-2/3})$ bins in $\cP$.

\subsection{Regret Analysis} \label{sec:proof-main-sq}

Define the sign function as $\sign{(x)} = 1 \text{ if } x > 0$; $ -1  \text{ if } x < 0$; and some $u \in [-1,1]  \text{ if } x = 0$.
For a vector $\bs x \in \mathbb{R}^d$, $\sign(\bs x) \in \mathbb{R}^d$ is defined by the coordinate-wise application of this rule. We start by presenting a sequence of useful lemmas.

\begin{restatable}{lemma}{lemkktsq} \label{lem:kkt-sq}
(\textbf{characterization of offline optimal}) 
Consider the following convex optimization problem (where $\tilde{z}_1,...,\tilde{z}_{n-1}$ are introduced as dummy variables)
\begin{mini!}|s|[2]                   
    {\tilde u_1,\ldots,\tilde u_n,
    \tilde{z_1},\ldots,\tilde z_{n-1}}                               
    {\frac{1}{2}\sum_{t=1}^{n} (y_t - \tilde u_t)^2}   
    {\label{eq:Example1}}             
    {}                                
    \addConstraint{\tilde z_t}{=\tilde u_{t+1} - \tilde u_{t} \: \forall t \in [n-1],}    
    \addConstraint{\sum_{t=1}^{n-1} |\tilde z_t|}{\le C_n \label{eq:constr-sq-2}}  
\end{mini!}

Let $u_1,\ldots,u_n, z_1,\ldots,z_{n-1}$ be the optimal primal variables and let $\lambda \ge 0$ be the optimal dual variable corresponding to the last constraint \eqref{eq:constr-sq-2}. By the KKT conditions, we have
\begin{itemize}
    \item \textbf{stationarity: } $y_t = u_t - \lambda (s_{t} - s_{t-1})$, where $s_t \in \partial|z_t|$ (a subgradient). Specifically, $s_t=\sign(u_{t+1}-u_t)$ if $|u_{t+1}-u_t|>0$ and $s_t$ is some value in $[-1,1]$ otherwise. For convenience of notations later, we also define 
    $s_n = s_0 = 0$.
    \item \textbf{complementary slackness: } $\lambda \left(\sum_{t=2}^n |u_t - u_{t-1}| - C_n \right) = 0$.
\end{itemize}
\end{restatable}

\begin{remark} \label{rem:prop}
We enumerate some elementary observations about the optimal primal variables in Lemma \ref{lem:kkt-sq} that will be used throughout.
\begin{enumerate}
    \item[P1] For any time point $t$, if the optimal solution $u_{t+1} > u_t$, then $s_t = 1$. Similarly $s_t = -1$ whenever $u_{t+1} < u_t$. If $u_t = u_{t+1}$, the $s_t$ can be any number in $[-1,1]$.
    
    \item[P2] Consider a sub-interval $[a,b]$ with $2 \le a \le n-1$ such that the optimal solution jumps at both the end points. i.e $u_k \neq u_{k-1}$ for $k \in \{b+1,a\}$. Define $\Delta s_{a \rightarrow b} := s_b - s_{a-1}$. Then either $|\Delta s_{a \rightarrow b}| = 0$ or $|\Delta s_{a \rightarrow b}| = 2$ since $s_{a-1} \in \{-1,1 \}$ and  $s_{b} \in \{-1,1 \}$.
    
    \item[P3] Consider a sub-interval $[1,b]$ with $b < n$ such that $u_{b+1} \neq u_b$. Then $|\Delta s_{1 \rightarrow b}| = 1$ since $s_{0} = 0$ by convention (Lemma \ref{lem:kkt-sq}). Similarly for a sub-interval $[a,n]$ with $a > 1$, such that $u_{a-1} \neq u_a$, we have $|\Delta s_{a \rightarrow n}| = 1$.
\end{enumerate}
\end{remark}

\noindent\textbf{Terminology.} We will refer to the optimal primal variables $u_1,\ldots,u_n$ in Lemma \ref{lem:kkt-sq} as the \emph{offline optimal sequence} in this section.

Next, we exhibit a useful partitioning scheme of the interval $[n]$.
\begin{restatable}{lemma}{lempartsq}\label{lem:part-sq}(\textbf{key partition}) 
Initialize $\mathcal Q \leftarrow \Phi$. Starting from time 1, spawn a new bin $[i_s,i_t]$ whenever $\sum_{j=i_s+1}^{i_t+1} | u_j -  u_{j-1} | > B/\sqrt{n_i}$, where $n_i = i_t - i_s + 2$. Add the spawned bin $[i_s,i_t]$ to $\mathcal Q$. Consider the following post processing routine.

\begin{enumerate}
    \item Initialize $\mathcal P \leftarrow \Phi$.
    \item For $i \in [|\mathcal Q|]$:
    \begin{itemize}
        \item if $u_{i_t} = u_{i_{t}+1}$:
        \begin{enumerate}
            \item Let $p$ be the largest time point with $u_{p:i_t}$ being constant and let $q$ be the smallest time point with $u_{i_t+1:q}$ being constant.
            \item Add bin $[i_s,p-1]$ to $\mathcal P$.
            
            \item If $(i+1)_t > q$ then add $[p,q]$ to $\cP$ and set $(i+1)_s \leftarrow q+1$.
            
            \item Goto Step 2.
            
        \end{enumerate}

        \item Add $[i_s,i_t]$ to $\cP$. Goto Step 2.
    \end{itemize}

\end{enumerate}

Let $M:=|\mathcal P|$. We have $M = O\left (1 \vee n^{1/3}C_n^{2/3} B^{-2/3} \right)$. Further for any bin $[i_s,i_t] \in \cP$, it holds that $\sum_{j=i_s+1}^{i_t} | u_j -  u_{j-1} | \le B/\sqrt{n_i}$ where $n_i = i_t - i_s + 1$.
\end{restatable}

\begin{remark} \label{rmk:deltas-lb}
Consider a bin $[i_s,i_t] \in \cP$. Let $\Delta s_i := s_{i_t} - s_{i_s-1}$. By virtue of the post processing routine of Lemma \ref{lem:part-sq}, the bin $[i_s,i_t]$ will conform to either of the cases P2 or P3 in Remark \ref{rem:prop}. So we have $|\Delta s_i| > 0$ implies $|\Delta s_i| \ge 1$.
\end{remark}

We emphasize that the bins $[i_s,i_t]$ we consider in Eq. \eqref{eq:reg-sq} belong to the partition $\cP$ of Lemma \ref{lem:part-sq}.  We proceed to bound $T_{1,i},T_{2,i}$ and $T_{3,i}$ in the regret decomposition of Eq.\eqref{eq:reg-sq}.

\begin{restatable}{lemma} {lemonesq} \label{lem:t1-sq} (\textbf{bounding } $T_{1,i}$)
Assume that we run \ALG{} with the settings described in Theorem \ref{thm:main-sq}. For any bin $i$ we have $T_{1,i} = O \left(B^2\log n \right)$
\end{restatable}

\begin{restatable}{lemma}{lemtwosq} \label{lem:t2-sq} (\textbf{bounding } $T_{2,i}$)
Define $C_i := \sum_{j=i_s+1}^{i_t} |u_j - u_{j-1}|$, the TV within bin $i$ incurred by the offline optimal solution. Let $\Delta s_i := s_{i_t} - s_{i_s - 1}$ and $n_i := i_t - i_s + 1$. We have $T_{2,i} \le \frac{-\lambda^2 (\Delta s_i)^2}{n_i}.$
\end{restatable}

\begin{restatable}{lemma}{lemthreesq} \label{lem:t3-sq} (\textbf{bounding }$T_{3,i}$) Let $C_i$ and $\Delta s_i$ be as in Lemma \ref{lem:t2-sq}.
\begin{description}
\itemsep0em
\item[Case(a)]If $|\Delta s_i| > 0$ then $T_{3,i} \le B^2 + 6\lambda C_i$.
\item[Case(b)] If $\Delta s_i = 0$ with $s_{i_s-1} = s_{i_t} = 1$ and the offline optimal $\bs u$ is non-decreasing within bin $i$, then $T_{3,i} \le B^2$.
\item[Case(c)] If $\Delta s_i = 0$ with $s_{i_s-1} = s_{i_t} = -1$ and the offline optimal $\bs u$ is non-increasing within bin $i$, then $T_{3,i} \le B^2$.
\end{description}
\end{restatable}

\begin{figure}[h!]
\begin{minipage}[t]{0.48\textwidth}
\includegraphics[width=\linewidth,keepaspectratio=true]{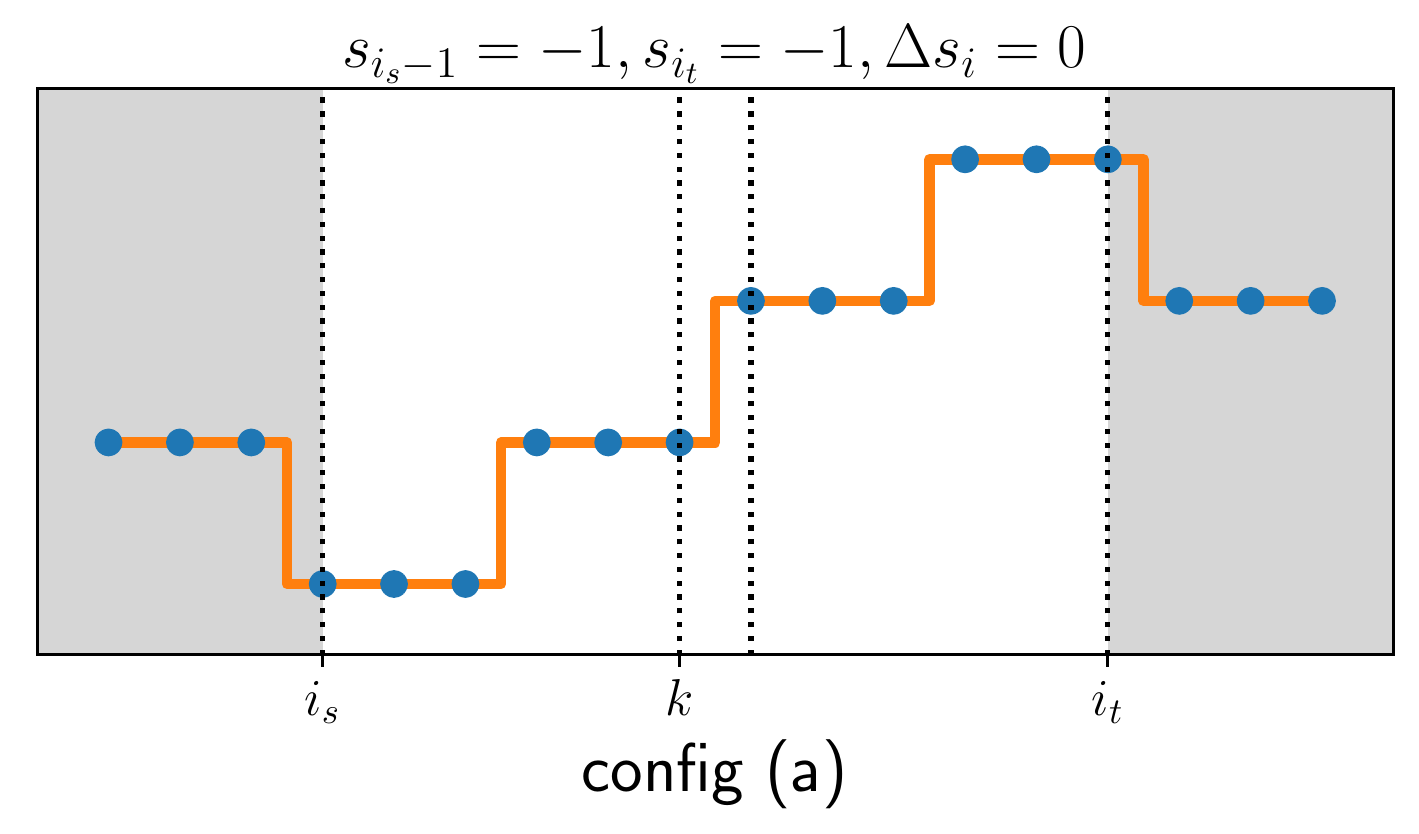}
\end{minipage}
\hspace*{\fill} 
\begin{minipage}[t]{0.48\textwidth}
\includegraphics[width=\linewidth,keepaspectratio=true]{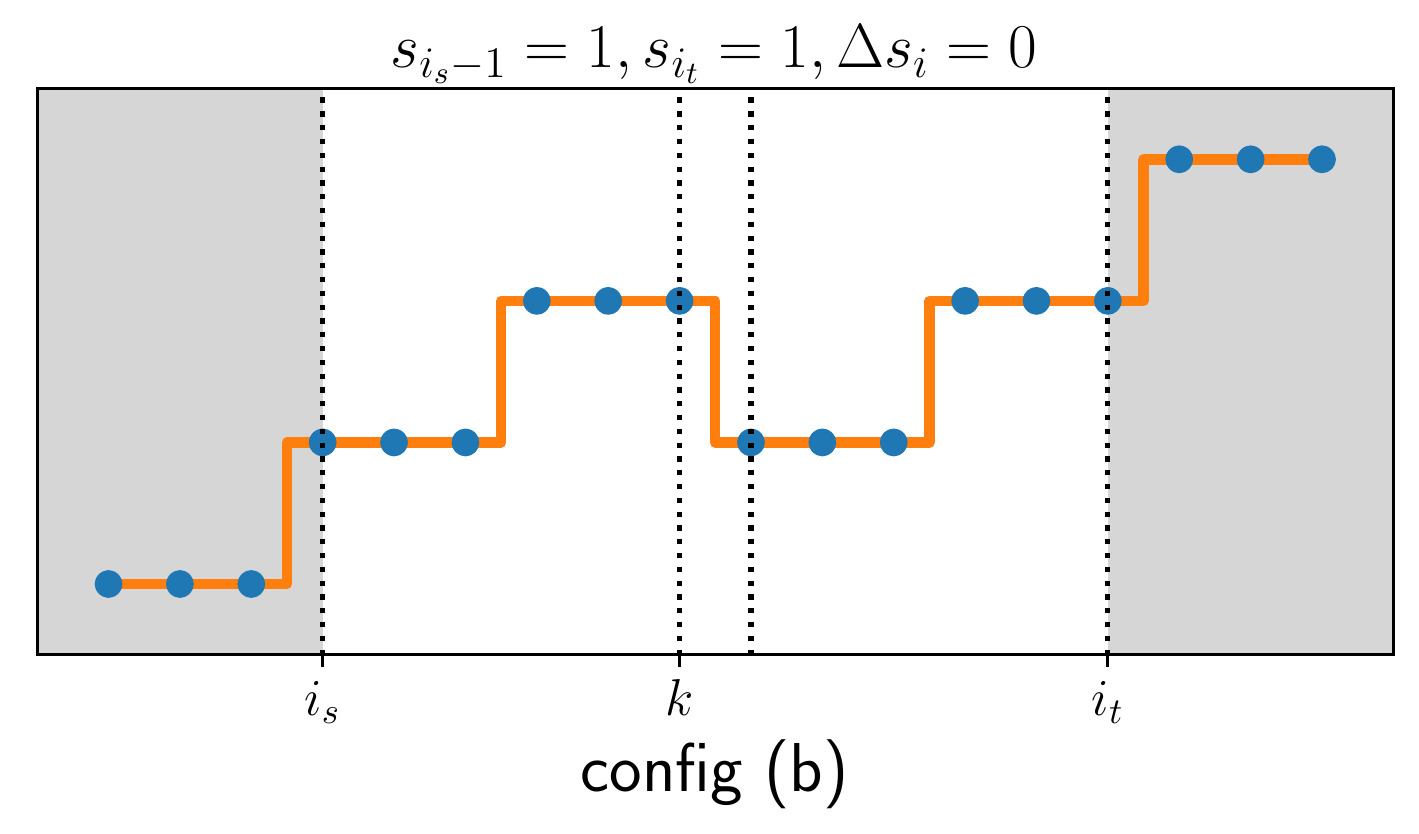}
\end{minipage}
\caption{\emph{Examples of configurations referred in the proof of Theorem \ref{thm:main-sq}. The blue dots corresponds to the offline optimal sequence.}}
\label{fig:sq}
\end{figure}

\begin{proof} \textbf{ of Theorem \ref{thm:main-sq}}
Tree diagrams that represent the flow of arguments in the proof is displayed in Fig.\ref{fig:sq-f1} and \ref{fig:sq-f2} in Appendix \ref{app:sq}. We start from the regret decomposition in Eq. \eqref{eq:reg-sq}. 

\noindent\textbf{Case (a) in Lemma \ref{lem:t3-sq}.}
First we handle case(a) in Lemma \ref{lem:t3-sq} where $|\Delta s_i| > 0$.  Define $T_i:= T_{1,i} + T_{2,i} + T_{3,i}$. From Lemmas \ref{lem:t1-sq}, \ref{lem:t2-sq} and \ref{lem:t3-sq} we have
\begin{align}
    T_i
    &\le  O\left(B^2\log n \right) - \frac{\lambda^2 (\Delta s_i)^2}{n_i}  + B^2 + 6\lambda C_i\\
    &\le O\left(B^2\log n \right) - \frac{\lambda^2 (\Delta s_i)^2}{n_i} + 6\lambda C_i\\
    &\le_{(a)} O\left( B^2 \log n \right) + \frac{9n_iC_i^2}{(\Delta s_i)^2} - \left(\frac{\lamda \Delta s_i}{\sqrt{n_i}} - \frac{3C_i \sqrt{n_i}}{\Delta s_i} \right)^2\\
    &\le_{(b)} O(B^2\log n) + 9B^2\\
    &\le O(B^2\log n),
\end{align}
where line (a) is obtained by completing the square. For line (b) we dropped the negative term used Remark \ref{rmk:deltas-lb} to conclude $|\Delta s_i| \ge 1$. Further $n_i C_i^2 \le B^2$ for bins in the partition $\cP$ of Lemma \ref{lem:part-sq}.

\noindent\textbf{Case (b) and (c) in Lemma \ref{lem:t3-sq}.} To handle case (b) and case (c) in Lemma \ref{lem:t3-sq} where $\Delta s_i = 0$ and monotonic, we have $T_{1,i} = O(B^2\log n)$ due to Lemma \ref{lem:t1-sq}, $T_{2,i}  \le 0$ due to Lemma \ref{lem:t2-sq} and $T_{3,i} \le B^2$ due to Lemma \ref{lem:t3-sq}. So $T_i \le  O\left(B^2 \log n \right) + B^2 \le O(B^2\log n)$.

\noindent\textbf{Other cases:} 

\textbf{(A1)} Consider the case when $\Delta s_i = 0$ with $s_{i_s-1} = s_{i_t} = -1$ and the offline optimal $\bs u$ is non-decreasing within bin $i$. If the sequence is constant within the bin, then trivially we have $T_i = O\left(B^2\log n \right)$ due to Strongly Adaptivity of \ALG{}. Otherwise, we the split the original bin into two sub-bins $[i_s,k]$ and $[k+1,i_t]$ such that $s_k = 1$ with $u_{k+1} > u_k$. See config (a) in Fig.\ref{fig:sq} for an illustration. Then the two sub-bins falls into the category of case (a) in Lemma \ref{lem:t3-sq}. By bounding the regret within each sub-bin separately by following the previous arguments for case (a) and adding them up, we can get $T_i \le  O\left( B^2 \log n \right)$ regret for the original bin. The arguments for the case when $\Delta s_i = 0$ with $s_{i_s-1} = s_{i_t} = 1$ and the offline optimal $\bs u$ is non-increasing within bin $i$ are similar.

\textbf{(A2)} To handle the case when  $\Delta s_i = 0$ and the optimal sequence is \emph{not} monotonic, we split the bin into two parts. Consider the case $s_{i_t} = s_{i_s-1} = 1$. We can split $\bs u_{i_s:i_t}$ as $\bs u_{i_s:k}$ and $\bs u_{k+1:i_t}$ such that the sequence $\bs u_{i_s:k}$ is non-decreasing and $s_k = -1$ with $u_k > u_{k+1}$. See config (b) in Fig.\ref{fig:sq} for an illustration. Notice that both the sub-bins  $\bs u_{i_s:k}$ and $\bs u_{k+1:i_t}$ now falls into the category of case(a) in Lemma \ref{lem:t3-sq}. Adding the bounds within these sub-bins by following the treatment for case (a) above yields $T_i \le O\left(B^2\log n \right)$. The arguments for the scenario $s_{i_t} = s_{i_s-1} = -1$ are similar.

Now the theorem follows by summing $\sum_{i=1}^{M} T_i$ for the $M =  O\left (1 \vee n^{1/3}C_n^{2/3}B^{-2/3} \right)$ bins in the partition $\cP$ of Lemma \ref{lem:part-sq}.
\end{proof}

The previous results generalize to online TV-denoising framework in higher dimensions.

\begin{restatable}{proposition} {prophigher} \label{prop:higher-sq}(\textbf{Extension to higher dimensions}) 
Consider a protocol where at each time the learner predicts a vector $\bs x_t \in \mathbb{R}^d$ after which the adversary reveals $\bs y_t$ such that $\|\bs y_t \|_{\infty} \le B$. Consider a comparator sequence of vectors $\bs w_1,\ldots,\bs w_n$ such that $ TV(\bs w_{1:n}) := \sum_{t=2}^{n} \|\bs w_t - \bs w_{t-1} \|_1 \le C_n$. By running $d$ instances of \ALG{} with learning rate $\zeta = 1/(8B^2)$ and FTL as base learners, where instance $i$, $i \in [d]$, predicts $\bs x_t[i]$ at time $t$, we have 
\begin{equation*}
    R_n(\bs w_{1:n})
    := \sum_{j=1}^{n} \|\bs y_t - \bs x_t \|_2^2 - \|\bs y_t - \bs w_t \|_2^2 = \tilde O\left (d B^2\log n \vee d^{1/3}n^{1/3} C_n^{2/3}B^{4/3} \right).
\end{equation*}
\end{restatable}

\begin{restatable}{proposition}{proplb}(\textbf{Lower bound}) \label{prop:lb-sq}
Assume the protocol and notations of Proposition \ref{prop:higher-sq}. For any algorithm, we have
\begin{align}
    \sup_{\bs w_{1:n} : TV(\bs w_{1:n}) \le C_n} R_n(\bs w_{1:n})
    &= \Omega \left (dB^2 \log n \vee d^{1/3}n^{1/3} C_n^{2/3} B^{4/3} \right).
\end{align}
\end{restatable}

By comparing the upper and lower bounds, we conclude that the FLH-FTL strategy in Proposition \ref{prop:higher-sq} is minimax optimal (modulo log factors) wrt \emph{all} parameters $d,n,B$ and $C_n$.
\begin{remark} \label{rem:other}
Several other non-parametric sequence classes such as the Holder ball
$\cH^{B}(B_n') = \{w_{1:n} : \|D w_{1:n}\|_\infty \le B_n', \| w_{1:n}\|_\infty \le B \}$
and Sobolev ball $\cS^B(C_n') = \{w_{1:n} : \|D w_{1:n}\|_2 \le C_n', \| w_{1:n}\|_\infty \le B \}$ can be shown to embedded inside a $\mathcal{TV}^B(C_n)$ ball for appropriate choices of $C_n, B_n$ and $B_n'$ (see \citep{arrows2019}) with all classes having the same minimax rates of estimation in the iid setting. So the minimax optimality on TV ball for FLH with FTL as base learners implies minimax optimality on the embedded Holder and Sobolev balls as well.
\end{remark}

\section{Performance guarantees for exp-concave losses} \label{sec:ec}

We begin by listing all the assumptions we make about the loss functions.

\begin{enumerate}
\itemsep0em
    \item[EC-1] Without loss of generality, we assume $\bs 0 \in \cW$. Let $B := \sup_{\bs x \in \cW} \|\bs x \|_\infty$. Define $\cD^- := \{\bs x \in \mathbb{R}^d : \| \bs x\|_\infty \le B \}$. The loss functions  $f_t(x):\mathbb{R}^d \rightarrow \mathbb{R}$ are $G$ Lipschitz in  $\cD^-$. 
    \item[EC-2] The loss functions are $\beta$ strongly smooth in $\cD = \{\bs x \in \mathbb{R}^d : \| \bs x \|_\infty \le B + G \}$. i.e $f_t(\bs y) \le f_t(\bs x) + (\bs y - \bs x)^T \nabla f_t(\bs x) + \frac{\beta}{2} \|\bs x - \bs y \|_2^2$,  for all $\bs x, \bs y \in \cD$.  We assume without loss of generality that $\beta \ge 1$.
    \item[EC-3] The loss functions are $\alpha$ exp-concave in  $\cD$. i.e $f_t(\bs y) \ge f_t(\bs x) + (\bs y - \bs x)^T \nabla f_t(\bs x) + \frac{\alpha}{2} \left( (\bs y - \bs x)^T \nabla f_t(\bs x) \right)^2$ for all $\bs x,\bs y\in\mathcal{D}$.
    \item[EC-4] The loss functions $f_t(x):\mathbb{R}^d \rightarrow \mathbb{R}$ are $G^\dagger$ Lipschitz in $\cD$.
\end{enumerate}

Below, we give an example of a family of loss functions that satisfy the above assumptions.
\begin{example}[Generalized linear models]\label{ex:glm}
Let $f_t(\bs x) = g(\bs v_t^T \bs x)$, where $g:\mathbb{R} \rightarrow \mathbb{R}$ is a convex function and $\bs v_t$ is a feature vector. Let $\| \bs v_t\|_2 \le R$. Assume that for all $\bs x \in \cD^-$ we have $|g_t'(\bs v_t^T \bs x)| \le a$. Further for all $\bs x \in \cD$, let $|g_t'(\bs v_t^T \bs x)| \le a^+$, $g_t''(\bs v_t^T \bs x) \le b$, $g_t''(\bs v_t^T \bs x) \ge c > 0$. Then Assumptions EC 1-5 are satisfied by by the losses $f_t$ with $G = aR$, $\beta = bR^2$, $\alpha = c/((a^+)^2)$ and $G^\dagger = Ra^+$.
\end{example}

We are interested in characterizing the maximum dynamic regret
\begin{align}
    R_n^+(C_n) := \sup_{\substack{\bs w_1,\ldots,\bs w_n \in \cD^- \\ \sum_{t=2}^n \|\bs w_t - \bs w_{t-1} \|_1 \le C_n}}
    \sum_{t=1}^n f_t(\bs x_t) - f_t(\bs w_t),
\end{align}
where $\bs x_t$ are the predictions of the learner. Since $\cW \subseteq \cD^-$, the dynamic regret against comparators in $\cD^-$ trivially upperbounds the dynamic regret against $\cW$. The algorithms that we study throughout this section are improper in the sense that the predictions of the algorithms belong to $\cD \supset \cW$. 

Before diving into the details, we remark that our main focus is to get optimal dependence on $n$ and $C_n$. The dimension $d$ is considered as a constant problem parameter and we do not try to optimize its polynomial dependence. All unspecified proofs of this section are given in Appendix \ref{app:ecd}. 

We have the following regret guarantee for exp-concave losses.

\begin{restatable}{theorem}{main}\label{thm:ec-d}
By using the base learner as ONS with parameter $\zeta = \min \left \{\frac{1}{4G^\dagger(2B\sqrt{d} + 2G/\beta)}, \alpha \right \}$, decision set $\cD$ and  choosing learning rate $\eta = \alpha$, \ALG{} obeys $R_n^+(C_n) = \tilde O \left(d^{3.5} (n^{1/3}C_n^{2/3} \vee 1) \right)$ if $C_n > 1/n$  and  $O(d^{1.5}\log n)$ otherwise. Here $a \vee b := \max \{a, b \}$ and $\tilde O(\cdot)$ hides dependence on the constants $B,G,G^\dagger,\alpha$ and factors of $\log n$.
\end{restatable}
\begin{proof}[proof sketch]
Let $\bs u_1,\ldots,\bs u_n$ be the \emph{offline optimal} sequence such that $\sum_{t=1}^{n} f_t(\bs u_t)$ is minimum across all sequences that obeys: (a) $\sum_{t=2}^{n} \|\bs u_{t} - \bs u_{t-1}\|_1 \le C_n $; (b) $\bs u_t \in \cD^-$ for all $t \in [n]$ (see Lemma \ref{lem:kkt-ec-d} in Appendix \ref{app:ecd} for more details).

Let $\cP$ be a partition of $[n]$ into $M =O^*(n^{1/3}C_n^{2/3})$ bins obtained by a similar scheme in Lemma \ref{lem:part-sq} where within each bin, we have $\sum_{j=i_s+1}^{i_t} \| \bs u_j -  \bs u_{j-1} \|_1 \le B/\sqrt{n_i}$. Let $[i_s, i_t]$ denote the $i^{th}$ bin in $\cP$ and let $n_i$ be its length.  Define $\bar {\bs u}_i = \frac{1}{n_i} \sum_{j=i_s}^{i_t} \bs u_j$ and $\dot {\bs u}_i = \bar {\bs u}_i - \frac{1}{n_i \beta} \sum_{j=i_s}^{i_t} \grad f_j(\bar {\bs u}_i)$ where $\beta$ is as in Assumption EC-2. Let $\bs x_j$ be the prediction made by \ALG{} at time $j$. We start with following regret decomposition.
\begin{align}
    R_n^+(C_n)
    &\le \sum_{i=1}^{M} \underbrace{\sum_{j=i_s}^{i_t} f_j(\bs x_j) - f_j(\dot {\bs u}_i)}_{T_{1,i}} + \sum_{i=1}^{M} \underbrace{\sum_{j=i_s}^{i_t} f_j(\dot {\bs u}_i) - f_j(\bar {\bs u}_i)}_{T_{2,i}} + \sum_{i=1}^{M} \underbrace{\sum_{j=i_s}^{i_t} f_j(\bar {\bs u}_i) - f_j(\bs u_j)}_{T_{3,i}} \label{eq:reg-ecm}
\end{align}

Unlike the squared error case, for the term $T_{1,i}$, we \emph{do not} compete with the minimizer of $g(\bs x) := \sum_{j=i_s}^{i_t} f_j(\bs x)$. Instead we compete with $\dot {\bs u}_i$ which is obtained by a one-step gradient descent of $g(\bs x)$ from the point $\bar {\bs u}_i$ where the step size is set as $1/(n_i \beta)$. 

Recall that the purpose of $\bar y_i$ in Eq. \eqref{eq:reg-sq} was to make $T_{2,i}$ non-positive thereby facilitating potential cancellation of terms arising from the bound on $T_{3,i}$. Since $g(\bs x)$ is $n_i \beta$ strongly smooth, by the well known \emph{descent lemma} in first order optimization (eg. see Eq. 3.5 in \citep{bubeck2015ConvexOA}), 
we can bound $T_{2,i}$ in Eq. \eqref{eq:reg-ecm} with a ``sufficiently negative'' term $ - \frac{1}{2n_i\beta}\|\nabla g(\bar{\bs u}_i)\|^2$ as well.
Also, observe that
\begin{align}
    \| \dot{\bs u}\|_\infty
    &\le \| \bar{\bs u}_i\|_\infty + \frac{\sum_{j=i_s}^{i_t}\|\grad f_j(\bar u) \|_\infty}{n_i \beta} \le B + G,
\end{align}
where in the last line we used the fact $\bar{\bs u}_i \in \cD^-$ and the Lipschitzness assumption in EC-1 along with $\beta > 1$ by assumption EC-2. So in $T_{1,i}$ the comparator term $\dot{\bs u}_i \in \cD$. The base learners of the FLH produce predictions in $\cD$ to compete with such a comparator hence making the overall algorithm improper. We do not project $\dot{\bs u}$ to the set $\cW$, because doing so appears to make $T_{2,i}$ not negative enough to adequately diminish the terms arising from $T_{3,i}$.

Rest of the proof proceeds by introducing lemmas analogous to the squared error case, carefully bounding $T_{1,i}+T_{2,i}+T_{3,i}$ for each bin in $\cP$ and summing them up across all bins. However, we remark that the analysis is significantly more involved in comparison to that of squared error case due to dual variables introduced by the additional constraint that $\bs u_t \in \cD^-$. 

We first present the proof for the 1D-exp-concave case in Appendix~\ref{app:ec_1d}, which illustrates how boundedness constraints are handled by the structures in the KKT-conditions (Lemma~\ref{lem:kkt-ec}) and by discussing various combinations (see Fig. \ref{fig:ec1-f1}-\ref{fig:ec1-f3}).
Then we present the full proof for the higher-dimensional exp-concave losses in Appendix~\ref{app:ec_highdim}, where the structure becomes too complex for us to enumerate all combinations. We address this by constructing an iterative algorithm that generates bins and prove that the algorithm is guaranteed to find a partition with cardinality $O^*(n^{1/3}C_n^{2/3})$ that satisfies a number of additional properties that give rise to the regret bound we claim.
\end{proof}

\begin{restatable}{proposition}{propsc}
For strongly convex losses, the regret bound can be improved to $\tilde O \left (d^2 (n^{1/3}C_n^{2/3} \vee 1) \right)$ if $C_n > 1/n$  and  $O(\log n)$ otherwise by using OGD as base learners in the FLH procedure. See Appendix \ref{app:ec_highdim} for a proof.
\end{restatable}

By comparing with the lower bound in Proposition \ref{prop:lb-sq} we conclude that the dynamic regret bound of Theorem \ref{thm:ec-d} is minimax optimal (up to $\log n$ factors) in $n$ and $C_n$.

\begin{remark}[Implications in statistical methodology.] \label{rem:meth}
Example~\ref{ex:glm} and Theorem~\ref{thm:ec-d} extends the locally-adaptive nonparametric regression theory that are typically studied for square loss to an arbitrary strongly convex / exp-concave loss while allowing covariates (exogenous variables) to be modeled. Moreover, the method enjoys strong oracle inequalities (e.g. Remark~\ref{rmk:oracle_inequality}) that certifies the predictive performance in a fully agnostic / model-misspecified setting with no stochastic assumptions. In addition, the method does not introduce additional tuning parameters at all.
\end{remark}

\section{Conclusion and further discussions}
In this paper, we considered the problem of dynamic regret minimization with exp-concave losses and showed that SA methods are minimax optimal (modulo factors of $\log n$ and $d$) in a setting where improper learning is allowed. To the best of our knowledge this is the first work that attains optimal dynamic regret rates under this setting. The resulting algorithms are adaptive to the path variation of the comparator sequence. Further, our results have far reaching consequences in locally adaptive non-parametric regression as mentioned in Remark \ref{rem:meth}. 

An open problem to investigate is if SA methods can still perform optimally in a proper learning setting. If we consider a very restrictive setup where the loss functions are exp-concave and for each function,  at-least one of the global optimal points lie in the comparator set $\cW$, it is indeed the case. An example of this scenario is the squared error loss $f_t(x) = (y_t - x)^2$ with $|y_t| \le B$ and $\cW = [-B,B]$ as in the TV-denoising setup.  On the other hand, if there exists an SA learner that can guarantee $O(\log n)$ static regret against any point in $\mathbb{R}^d$ in any time interval, then our results provides optimal proper learning when $\cD = \cW = \mathbb{R}^d$.





\section*{Acknowledgments}
The research was partially supported by NSF Award \#2029626, \#2007117
and a start-up grant from UCSB CS department.

\bibliography{tf,yx}
\bibliographystyle{plainnat}

\newpage

\appendix
\section{More on Related Work} \label{app:lit} 

Throughout this section, we refer to the variationals $P_n$ in Eq.\eqref{eq:pn} and $D_n$ in Eq.\eqref{eq:f-var} where the arguments are dropped for brevity. In the OCO setting, when the environment is benign, \citep{Zhao2020DynamicRO} replaces the $\sqrt{n}$ dependence in the regret of $O(\sqrt{n(1+P_n)})$ attained by \citep{zhang2018dynamic} with problem dependent quantities that could be much smaller than $\sqrt{n}$. Although the linear smoother lower bound in Proposition 2 of \citep{arrows2019} would imply that an OEGD (Online Extra Gradient Descent) \citep{Zhao2020DynamicRO} expert with any learning rate sequences require $\Omega(\sqrt{n P_n})$ dynamic regret for the 1D-TV-denoising problem.

Interestingly in \citep{yuan2019dynamic} the authors mention that even in the one-dimensional setting, a lower bound on dynamic regret for strongly convex / exp-concave losses that holds uniformly for the entire range $0 \le P_n \le n$ is unknown. However, we find that one can combine the existing lower bounds on univariate TV-denoising in a stochastic setting \citep{donoho1998minimax} with the lower bound construction of \citep{vovk2001} (or see Theorem 11.9 in \citep{BianchiBook2006}) for online learning with squared error losses to obtain an $\Omega(\log n \vee n^{1/3}C_n^{2/3})$ in one dimensions (see Appendix \ref{app:sq} for details). In this work, we show that SA methods can achieve a regret that matches this lower bound (modulo polynomial factors of dimension and $\log n$) when losses are strongly convex / exp-concave.

When the loss functions are strongly convex, \citep{Mokhtari2016OnlineOI} studies the dynamic regret against the comparator points that are the unique minimizers of the revealed losses in the set $\cW$ ($=\cD$). Specifically when $\bs w_t^* = \argmin_{\bs x \in \cW} f_t(\bs x)$, and $C^*_n := \sum_{t=2}^{n} \| \bs w^*_t - \bs w^*_{t-1}\|_2$, they show that OGD can be used to get the rate of $O(1+C_n^*)$ for the dynamic regret against the sequence $\bs w^*_{1:n}$. However, as noted in \citep{zhang2018adaptive}, that even though this implies an $O(1+C_n^*)$ bound on the dynamic regret against arbitrary comparator sequences in Eq.\eqref{eq:d-regret}, the resulting bound can be overly pessimistic. As an example, in TV-denoising, $f_t(x) =  (y_t-x)^2$ where $y_t = w_t + \text{Noise}$. Even if $w_t$ obeys that $\mathrm{TV}(w_{1:n}) = O(1)$, we would still have $E[C_n^*] = \sum_{t=2}^nE[|y_t - y_{t-1}|] \geq  \Omega(n)$ , thus the $O(1+C_n^*)$ regret bound does not imply any non-trivial bounds in our setting, e.g., if we take the comparator sequence to be $w_1,...,w_n$.

Different variational measures capture different aspects of the online learning problem and are not comparable in general. \citep{jadbabaie2015online} introduces a policy that attains dynamic regret in terms of $D_n$ and $P_n$ simultaneously. Various other interesting variational measures and strategies to control dynamic regret can can be found in the works of \citep{yang2016tracking,chen2018non}.

The seminal work of \citep{hazan2007adaptive} introduces the notion of weakly adaptive regret which is defined as the maximum static regret incurred by the learning algorithm in any continuous interval. They propose algorithms that obtain static regret guarantees of $\tilde O(\sqrt{n})$ for convex losses and $\tilde O(1)$ and $\tilde O(d)$ for strongly convex and exp-concave losses respectively. This has been further developed in \citep{koolen2016specialist}. However, one drawback of weakly adaptive regret stems from its trivial regret guarantees on short intervals. For example, an $\tilde O(\sqrt{n})$ static regret guarantee on an interval of length $\sqrt{n}$ is meaningless. This drawback is overcame by the notion of Strongly Adaptive regret as discussed in Section \ref{sec:lit} by taking into account the length of the interval where the static regret is computed.

\citep{zhang2018dynamic} shows that SA methods enjoys a dynamic regret of $\tilde O(n^{2/3}D_n^{1/3})$ for convex functions and $\tilde O(\sqrt{nD_n})$ and $\tilde O(\sqrt{dnD_n})$ for strongly convex and exp-concave losses respectively in an OCO setting. Thus when combined with our results, we can conclude that SA methods are simultaneously optimal wrt to the dynamic regret based on $D_n$ and $P_n$ in an application that allows improper learning.


Our online TV-denoising setting studied in Section \ref{sec:sq} can be cast into the framework of \citep{rakhlin2014online}. They study the regret against non-parametric function classes under squared error loss as follows
\begin{align}
    R_n' = \sum_{t=1}^{n}(\hat{y}_t - y_t)^2 - \inf_{f \in \cF} \sum_{t=1}^{n}(f( x_t) - y_t)^2.
\end{align}
Our TV-denoising setting setting becomes identical to \citep{rakhlin2014online} if one takes the comparator class $\cF$ to be space of TV bounded functions and when the features $x_t$ are revealed in an isotonic order: $ x_1 \le \ldots \le  x_n$. All the results can be trivially extended to the case of arbitrary covariates that may be non-isotonically revealed by maintaining online averages across all intervals in a Geometric Cover on $[n]$ and using the specialist aggregation scheme in \citep{koolen2016specialist} (see \citep{baby2021TVDenoise} for an illustration of this idea in a stochastic setting). We do not follow this path for the sake of simplicity of exposition.

The results of \citep{rakhlin2014online} establish the minimax rates for the quantity $R_n'$ when $\cF$ is taken to be a Besov ball. It is known that a TV ball is sandwiched between two Besov spaces (see for eg. \citep{donoho1998minimax}) that have the same minimax rate for $R_n'$. Hence results of \citep{rakhlin2014online} establishes that minimax regret of our problem is $\tilde{O}(n^{1/3})$. However their bounds don't capture the correct dependence on $C_n$ and are obtained by non-constructive arguments. In contrary we obtain upper bounds with optimal dependence on both $n$ and $C_n$ by an efficient algorithm.

\citep{kotlowski2016} proposes a policy that achieves a rate of $\tilde{O}(n^{1/3})$ for $R_n'$ when $\cF$ is the family of isotonic functions that take values in $[0,1]$. This class is indeed a subset of $\mathcal{TV}^B(1)$. They exploit the property that the optimal isotonic function is piecewise constant and within a constant section, it takes the value equal to mean of labels $y_t$ within that section. However for our case the offline problem solved by the oracle is an instance of a constrained fused LASSO which doesn't yield such nice closed form expression for value of optimal function within a constant section. 

\citep{gaillard2015chaining} proposes a novel chaining algorithm that achieves optimal rate for $R_n'$ when $\cF$ is the family of Holder smooth functions. The functions residing in this class are spatially homogeneous and more regular than the TV class. We show that our policy is also optimal for regret against Holder ball embedded within a $\mathcal{TV}^B$ space (see Remark \ref{rem:other}). Interestingly the generic forecaster they proposed can be shown to yield the optimal $\tilde{O}(n^{1/3})$ rate for our problem. However, the run-time of that policy is exponential.

Our setting is closely related to the setup studied in \citep{koolen2015minimax}. Their setting can be viewed as competing against Sobolev sequences which are more regular than TV bounded sequences. We show that our policy is also optimal for regret against Sobolev ball embedded within a TV bounded space (see Remark \ref{rem:other}).

We now proceed to explain how the model agnostic regret guarantees presented in this paper imply minimax statistical estimation rate in a stochastic setting. When applied to squared error losses, the FLH-FTL procedure can yield
\begin{align}
    \sum_{t=1}^n (y_t - x_t)^2 - (y_t - w_t)^2
    &= \tilde O(n^{1/3}C_n^{2/3}), \label{eq:regret}
\end{align}
where $x_t$ are the predictions of FLH-FTL procedure and $w_{1:n} \in \mathcal{TV}^B(C_n)$ (see Eq.\eqref{eq:tvb}). We demonstrate how this implies minimax estimation in an  iid setting which is the usual subject of study in non-parametric regression among the statistics community. In the stochastic setting we have the following observation model:
\begin{align}
    y_t = w_t + \epsilon_t, \: t \in [n]
\end{align}
for some fixed $w_{1:n} \in \mathcal{TV}^B(C_n)$ and $\epsilon_t$ are iid zero mean subgaussian noise with magnitude at-most $B$ (We could relax the boundedness to be obeyed with high probability). We have,
\begin{align}
    \sum_{t=1}^n E[(y_t - x_t)^2] - E[(y_t - w_t)^2]
    &= \sum_{t=1}^n E[(x_t-w_t)^2] - 2 E[\epsilon_t (x_t-w_t)] + E[\epsilon_t^2] - E[\epsilon_t^2]\\
    &=_{(a)} \sum_{t=1}^n E[(x_t-w_t)^2] - 2 E[\epsilon_t] E[(x_t-w_t)]\\
    &= \sum_{t=1}^n E[(x_t-w_t)^2]\\
    &=_{(b)} \tilde  O(n^{1/3}C_n^{2/3}),
\end{align}
where line (a) is due to the fact that $x_t$ and $\epsilon_t$ are mutually independent and line (b) is due to Eq.\eqref{eq:regret}. From \citep{arrows2019}, this is indeed the minimax rate of estimating $w_{1:n}$ under the stochastic setting.

Thus we conclude that the model agnostic regret guarantees presented in this paper implies minimax estimation rate in a stochastic setting and hence the former is strictly stronger.

\section{Preliminaries} \label{app:prelim}
In this section, we recall the Follow-the-Leading-History (FLH) algorithm from \citep{hazan2007adaptive} along with some basic definitions.

\begin{definition} (\textbf{Strong convexity})
 \label{as:sc}
 Loss functions $f_t$ are said to be $H$ strongly convex in the domain $\mathcal{D}$ if it satisfies $$f_t(\bs y) \ge f_t(\bs x) + (\bs y - \bs x)^T \nabla f_t(\bs x) + \frac{H}{2} \|\bs x - \bs y \|^2,$$  for all $\bs x, \bs y \in \cD$.
\end{definition}

\begin{figure}[h!]
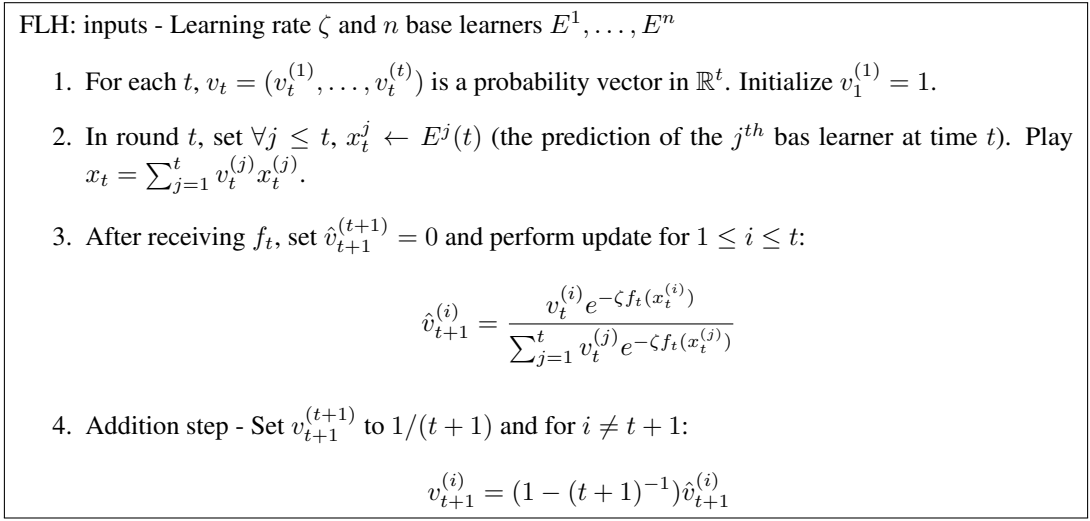

	\centering
	\fbox{
		\begin{minipage}{14 cm}
		FLH: inputs - Learning rate $\zeta$ and $n$ base learners $E^1,\ldots,E^n$
            \begin{enumerate}
                \item For each $t$, $v_t = (v_t^{(1)},\ldots,v_t^{(t)})$ is a probability vector in $\mathbb{R}^t$. Initialize $v_1^{(1)} = 1$.
                \item In round $t$, set $\forall j \le t$, $x_t^j \leftarrow E^j(t)$ (the prediction of the $j^{th}$ bas learner at time $t$). Play $x_t =  \sum_{j=1}^t v_t^{(j)}x_t^{(j)}$.
                \item After receiving $f_t$, set $\hat v_{t+1}^{(t+1)} = 0$ and perform update for $1 \le i \le t$:
                \begin{align}
                    \hat v_{t+1}^{(i)}
                    &= \frac{v_t^{(i)}e^{-\zeta f_t(x_t^{(i)})}}{\sum_{j=1}^t v_t^{(j)}e^{-\zeta f_t(x_t^{(j)})}}
                \end{align}
                \item Addition step - Set $v_{t+1}^{(t+1)}$ to $1/(t+1)$ and for $i \neq t+1$:
                \begin{align}
                    v_{t+1}^{(i)}
                    &= (1-(t+1)^{-1}) \hat v_{t+1}^{(i)}
                \end{align}
            \end{enumerate}
		\end{minipage}
	}
	\caption{FLH algorithm}
	\label{fig:flh}
\end{figure}

FLH enjoys the following guarantee against any base learner.
\begin{proposition}\label{prop:flh} \citep{hazan2007adaptive}
Suppose the loss functions are exp-concave with parameter $\alpha$. For any interval $I = [r,s]$ in time, the algorithm FLH Fig.\ref{fig:flh} with learning rate $\zeta = \alpha$ gives $O(\alpha^{-1}( \log r + \log |I|))$ regret against the base learner in hindsight.
\end{proposition}

\begin{definition} (\citep{daniely2015strongly}) \label{def:sa}
An algorithm is said to be Strongly Adaptive (SA) if for every contiguous interval $I \subseteq [n]$, the static regret incurred by the algorithm is $O(\text{poly}(\log n) \Gamma^*(|I|))$ where $\Gamma^*(|I|)$ is the value of minimax static regret incurred in an interval of length $|I|$.
\end{definition}

It is known from \citep{hazan2007logregret} that OGD and ONS achieves static regret of $O(\log n)$ and $O(d \log n)$ for strongly convex and exp-concave losses respectively. Hence in view of Proposition \ref{prop:flh} and Definition \ref{def:sa}, we can conclude that:
\begin{itemize}
    \item FLH with OGD as base learners is an SA algorithm for strongly convex losses.
    \item FLH with ONS as base learners is an SA algorithm for exp-concave losses. (We treat dimension $d$ as a constant problem parameter and consider minimaxity only wrt $n$.)
\end{itemize}

We have the following guarantee on runtime.
\begin{proposition} \citep{hazan2007adaptive}
Let $\rho$ be the per round run time of base learners and $r_n$ be the static regret suffered by the base learners over $n$ rounds. Then FLH procedure has a runtime of $O(\rho n)$ per round. To improve the runtime one can use AFLH procedure from \citep{hazan2007adaptive} that incurs $O(\rho \log n)$ runtime overhead per round and suffers $O(r_n \log n)$ static regret in any interval.
\end{proposition}

\section{Proofs for Section \ref{sec:sq}} \label{app:sq}

We start by providing an example of a scenario where $\lambda$ in Lemma \ref{lem:kkt-sq} can scale linearly with $n$.

\begin{example} \label{ex}
Consider the $\mathcal{TV}(C_n)$ class with $C_n = 1$ and $n \ge 6$. Let the offline optimal be given by the step sequence $u_1=\ldots =u_{(n/2)-1} = 0$ and $u_{n/2}=\ldots = u_n = 1$. Our aim is to generate a sequence of labels $y_t$ such that this sequence $\bs u$ is indeed the offline optimal in the class $\mathcal{TV}(1)$ along with the property that the optimal dual variable $\lambda$ scales linearly with the horizon $n$.

Clearly we must have $s_{(n/2)-1} = 1$. For some appropriate parameter $\epsilon$, consider the following sign assignment:
\begin{itemize}
    \item $s_{(n/2)-2} = 1-\epsilon, s_{(n/2)-3} = 1-2\epsilon,\ldots,s_{1} = 1-((n/2)-2)\epsilon$,
    \item $s_{n-1} = \epsilon, s_{n-2} = 2\epsilon,\ldots,s_{n/2} = (n/2)\epsilon$.
\end{itemize}

By setting $\epsilon = 2/n$ for $ n \ge 6$, we get a consistent sign assignment because $s_{t} \in [-1,1]$ for all $1 \le t \le (n/2)-2$ which corresponds to the portion where $u_t = 0$; $s_{(n/2)-1} = 1$; and $s_{t} \in [-1,1]$ for all $n/2 \le t \le n-1$ which corresponds to the portion where $u_t = 1$.

By taking $\lambda = n/2$ the adversary can generate labels $y_t$ according to the stationarity condition in Lemma \ref{lem:kkt-sq} as follows:

\begin{itemize}
    \item $y_1 = -2$,
    \item $y_t = -1, \text{ for }2\le t \le (n/2)-1$,
    \item $y_{n/2} = 1$,
    \item $y_t = 2, \text{ for } (n/2)+1 \le t \le n$.
\end{itemize}

Since the TV of the sequence $\bs u$ is 1, the complementary slackness is also satisfied. Thus we conclude that if the labels $y_t \in [-2,2]$ are generated as above, the offline optimal sequence in $\mathcal{TV}(1)$ class is given by the step sequence $\bs u$. Furthermore, the optimal dual variable $\lambda = n/2$ scales linearly with the horizon.

\end{example}

\lemkktsq*
\begin{proof}
We can form the Lagrangian of the optimization problem as:
\begin{align}
    \mathcal{L}(\tilde{\bs u} , \tilde{\bs z}, \tilde{\bs v}, \tilde{\lambda})
    &= \frac{1}{2}\sum_{t=1}^{n} (y_t - \tilde u_t)^2 + \tilde \lambda \left( \sum_{t=1}^{n-1} |\tilde z_t| - C_n\right) + \sum_{t=1}^{n-1} \tilde v_t( \tilde u_{t+1} - \tilde u_t - \tilde z_t),
\end{align}
for dual variables $\tilde \lambda > 0$ and $\tilde {\bs v} \in \mathbb{R}^{n-1}$ unconstrained. Let the $(\bs u, \bs z, \bs v, \lambda)$ be the optimal primal and dual variables. By stationarity conditions, we have 
\begin{align}
    u_t - y_t = v_t - v_{t-1},
\end{align}
where we take $v_0 = v_n = 0$ and
\begin{align}
    v_t = \lambda s_t
\end{align}
Combining the above two equations and the complementary slackness rule yields the lemma.
\end{proof}

\lempartsq*
\begin{proof}
Let's use the notation $TV[a,b]$ to denote the TV incurred by the optimal solution sequence in the interval $[a,b]$. Let $\mathcal Q = \{[\ubar t_1, \bar t_1],\ldots,[\ubar t_N, \bar t_N] \}$ with $\ubar t_1 := 1$ and $\bar t_N := n$. Let $n_j:= \bar t_j - \ubar t_j + 1$ We have,

\begin{align}
    \sum_{j=1}^{N-1} TV[\ubar t_j, \bar t_j+1] \le C_n.
\end{align}

By construction we have $TV[\ubar t_j, \bar t_j+1] > \nu/\sqrt{n_j}$. So,

\begin{align}
    C_n
    &\ge \sum_{j=1}^{N-1} \nu/\sqrt{n_j}\\
    &\ge (N-1)^{3/2} \nu/\sqrt{n},
\end{align}
where the last line follows by Jensen's inequality. Rearranging gives the bound on $N = O\left (1 \vee n^{1/3}C_n^{2/3} B^{-2/3} \right)$. Now the post processing step only increases the number of bins by $O(N)$. Thus we get $M = O\left (1 \vee n^{1/3}C_n^{2/3} B^{-2/3} \right)$.
\end{proof}

\lemonesq*
\begin{proof}
Note that FTL with squared error losses outputs predictions which are online averages of the past labels that the algorithm has seen so far. Hence the predictions of all base learners as well as \ALG{} belong to the interval $[-B,B]$. It is known that (see for eg. \citep{BianchiBook2006}, Chapter 3) squared error losses are $1/(8B^2)$ exp-concave in the interval $[-B,B]$ . Further FTL with squared error losses suffers only logarithmic regret of $O(B^2 \log n)$ (\citep{BianchiBook2006}, Chapter 3). 

Hence due to the adaptive regret bound of \ALG{} (Theorem 3.2 in \citep{hazan2007adaptive}) by setting the learning rate $\zeta = 1/(8B^2)$, we have that the static regret of \ALG{} in any interval $[i_s,i_t]$ is also $O(\log n)$. This proves the lemma.
\end{proof}

\lemtwosq*
\begin{proof}
From the stationarity conditions in Lemma \ref{lem:kkt-sq}, we can write
\begin{align}
    \bar u_i - \bar y_i = \frac{\lamda \Delta s_i}{n_i}. \label{eq:t1-sq}
\end{align}

Further,
\begin{align}
    \sum_{j=i_s}^{i_t} (y_j - \bar y_i)^2 - (y_j - \bar u_i)^2
    &= n_i (\bar u_i - \bar y_i)^2 + 2\sum_{j=i_s}^{i_t} (y_j - \bar u_i) (\bar u_i - \bar y_i)\\
    &= -n_i (\bar u_i - \bar y_i)^2
\end{align}

Now plugging in Eq. \eqref{eq:t1-sq} yields the lemma.

\end{proof}

\lemthreesq*
\begin{proof}
Applying stationarity conditions, we have
\begin{align}
    T_{3,i}
    &= \sum_{j=i_s}^{i_t} (y_j - \bar u_i)^2 - (y_j - u_j)^2\\
    &= \sum_{j=i_s}^{i_t} (u_j - \bar u_i) (2y_j - \bar u_i - u_j)\\
    &= \sum_{j=i_s}^{i_t} (u_j - \bar u_i) (2y_j - 2 u_j + u_j - \bar u_i)\\
    &= \sum_{j=i_s}^{i_t} (u_j - \bar u_i)^2 + 2 \lamda (u_j - \bar u_i) (s_{j-1} - s_j)\\
    &\le n_i C_i^2 + \sum_{j=i_s}^{i_t} 2 \lamda (u_j - \bar u_i) (s_{j-1} - s_j),\label{eq:t3-sq-proof}
\end{align}
where in the last line we used $|u_j - \bar u_i| \le C_i$. Also observe that $n_i C_i^2 \le B^2$ for bins in the partition $\cP$ by Lemma \ref{lem:part-sq}. Now by expanding the second term followed by a regrouping of the terms in the summation, we can write
\begin{align}
   \sum_{j=i_s}^{i_t} 2 \lamda (u_j - \bar u_i) (s_{j-1} - s_j) 
   &= 2\lambda \left( s_{i_s-1} (u_{i_s} - \bar u_i) - s_{i_t} (u_{i_t} - \bar u_i)\right) + 2\lambda \sum_{j=i_s+1}^{i_t} |u_j - u_{j-1}|\\
   &= 2 \lamda C_i + 2\lambda \left( s_{i_s-1} (u_{i_s} - \bar u_i) - s_{i_t} (u_{i_t} - \bar u_i)\right) \label{eq:inter-sq}
\end{align}
Now we discuss the three cases.
\begin{description}
\item[Case (a)] When $|\Delta s_i| > 0$, then by triangle inequality we have \\$2\lambda \left( s_{i_s-1} (u_{i_s} - \bar u_i) - s_{i_t} (u_{i_t} - \bar u_i)\right) \le 4\lambda C_i$.
\item[Case (b)] In this case we have 
$2\lambda \left( s_{i_s-1} (u_{i_s} - \bar u_i) - s_{i_t} (u_{i_t} - \bar u_i)\right) = \lambda (u_{i_s} - u_{i_t}) =  -2\lambda C_i$ since the sequence is non-decreasing within the bin. Hence this term cancels with the corresponding additive term  of  $2\lamda C_i$ in Eq. \eqref{eq:inter-sq}. 
\item[Case (c)]By similar logic as in case (b) we can once again write\\
$2\lambda \left( s_{i_s-1} (u_{i_s} - \bar u_i) - s_{i_t} (u_{i_t} - \bar u_i)\right) =  -2\lambda C_i$.
\end{description}
Substituting the bound of each case into \eqref{eq:t3-sq-proof}. we obtain the expression as stated.
\end{proof}

\prophigher*
\begin{proof}
Let $\bs u_1, \ldots, \bs u_n$ be the offline optimal sequence. Let $C_n[k] = \sum_{t=2}^{n} |\bs u_t [k] - \bs u_{t-1}[k]|$ be its TV allocated to coordinate $k$. WLOG, let's assume the \ALG{} for coordinates $k \in [k']$ for $k' \le d$ incurs $\tilde O\left(n^{1/3} (C_n[k])^{2/3} B^{4/3} \right)$ regret and the regret incurred by \ALG{} for coordinates $k > k'$ is $O(\log n)$. Since squared error losses decomposes coordinate-wise, we have
\begin{align}
    R_n(\bs w_{1:n})
    &\le \sup_{\bs w_{1:n} : TV(\bs w_{1:n}) \le C_n} R_n(\bs w_{1:n})\\
    &=  R_n(\bs u_{1:n})\\
    &= (d-k') B^2 \log n + \sum_{k=1}^{k'} \tilde O\left(n^{1/3} (C_n[k])^{2/3} B^{4/3} \right)\\
    &\le (d-k') B^2 \log n + \tilde O\left(n^{1/3}  (k')^{1/3} B^{4/3}\left( \sum_{k=1}^{k'} C_n[k] \right) ^{2/3} \right),
\end{align}
where the last line follows by Holder's inequality $\bs x^T \bs y \le \| x\|_3 \| y\|_{3/2}$, where we treat $\bs x$ as just a vector of ones in $\mathbb{R}^{k'}$. The above expression can be further upper bounded by\\ $\tilde O\left (2d B^2 \log n \vee 2d^{1/3}n^{1/3} C_n^{2/3} B^{4/3}\right)$.

\end{proof}

\proplb*
\begin{proof}
Consider a fixed (but unknown) sequence $\bs u_1,\ldots,\bs u_n$ such that $TV(\bs u_{1:n}) \le C_n$ with $\| \bs u_t\|\infty \le B/2$ and TV along the coordinate $k \in [d]$, $TV(\bs u_{1:n}[k]) \le C_n/d$ for all $k$. Let the labels be $\bs y_t = \bs u_t + \bs \epsilon_t$  where each coordinate of $\bs \epsilon_t$ is generated by iid $U[-B/2,B/2]$. Further $\bs \epsilon_1,\ldots,\bs \epsilon_n$ are also iid. Then by the results of \citep{donoho1998minimax}, for any prediction strategy that produces outputs $\bs x_t$, we have
\begin{align}
    \sup_{\bs w_{1:n} : TV(\bs w_{1:n}) \le C_n} R_n(\bs w_{1:n})
    &\ge \sum_{k=1}^d \sum_{t=1}^{n} E \left[ (\bs y_t[k] - \bs x_t[k])^2 - (\bs y_t[k] - \bs u_t[k])^2 \right]\\
    &=_{(a)}\sum_{k=1}^d \sum_{t=1}^{n} E \left[ (\bs u_t[k] - \bs x_t[k])^2 \right]\\
    &= \sum_{k=1}^d \Omega(n^{1/3} (C_n/d)^{2/3} B^{4/3})\\
    &= \Omega(d^{1/3} n^{1/3} C_n^{2/3} B^{4/3}),
\end{align}
where in line (a) we used the fact that $\bs x_t[k]$ is independent of $\bs y_t[k]$ and $\bs y_t[k] - \bs u_t[k] \sim U[-B/2, B/2]$.

The $d B^2 \log n$ part of the lower bound is implied by the lower bound construction of Vovk \citep{vovk2001} (or cf. proof of Theorem 11.9 in \citep{BianchiBook2006}). 

\end{proof}

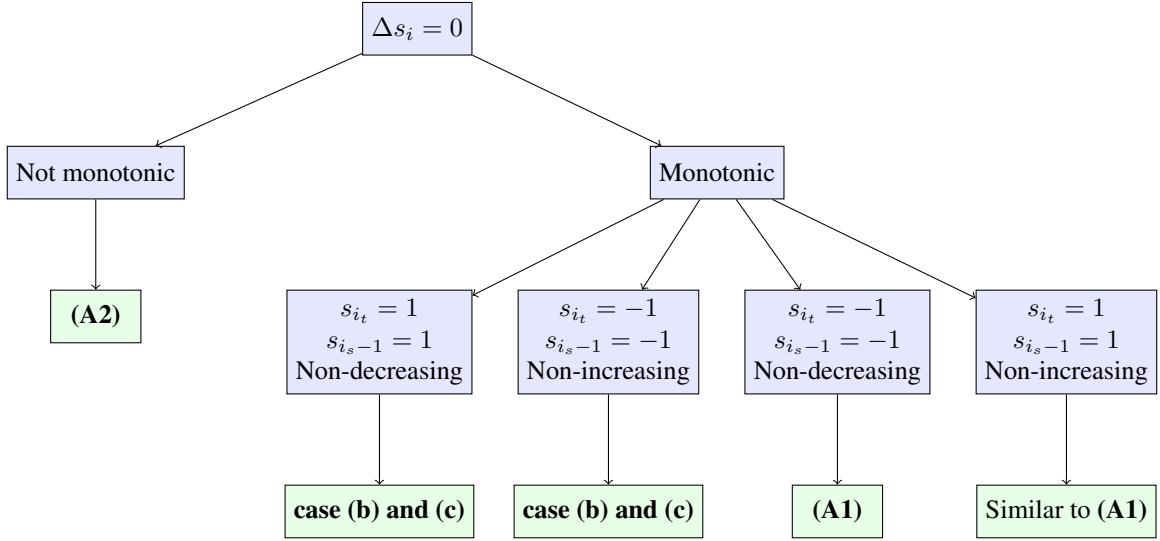
\begin{figure}
\centering
\begin{tikzpicture}[
    node distance = 12mm and 6mm,
       box/.style = {rectangle, draw, fill=#1, 
                     minimum width=12mm, minimum height=7mm}
                        ]
\node (not-mono) [box=blue!10] {Not monotonic};
\node (delta) [box=blue!10, above right = of not-mono, xshift = 50] {$\Delta s_i = 0$};
\node (mono) [box=blue!10,below right=of delta, xshift=50] {Monotonic};
\node (a1) [align=center, box=blue!10,below  left=of mono, xshift= -50] {$s_{i_t} = 1$ \\ $s_{i_s-1} = 1$\\ Non-decreasing};
\node (a2) [align=center, box=blue!10, right=of a1] {$s_{i_t} = -1$ \\ $s_{i_s-1} = -1$\\ Non-increasing};
\node (a3) [align=center, box=blue!10, right=of a2] {$s_{i_t} = -1$ \\ $s_{i_s-1} = -1$\\ Non-decreasing};
\node (a4) [align=center, box=blue!10, right=of a3] {$s_{i_t} = 1$ \\ $s_{i_s-1} = 1$\\ Non-increasing};
\node (L1) [align=center, box=green!10, below=of not-mono] {\textbf{(A2)}};
\node (L2) [align=center, box=green!10, below=of a1] {\textbf{case (b) and (c)}};
\node (L3) [align=center, box=green!10, below=of a2] {\textbf{case (b) and (c)}};
\node (L4) [align=center, box=green!10, below=of a3] {\textbf{(A1)}};
\node (L5) [align=center, box=green!10, below=of a4] {Similar to \textbf{(A1)}};
\draw[->] (delta) to  (not-mono);
\draw[->] (delta) to  (mono);
\draw[->] (mono) to  (a1);
\draw[->] (mono) to  (a2);
\draw[->] (mono) to  (a3);
\draw[->] (mono) to  (a4);
\draw[->] (not-mono) to  (L1);
\draw[->] (a1) to  (L2);
\draw[->] (a2) to  (L3);
\draw[->] (a3) to  (L4);
\draw[->] (a4) to  (L5);

\end{tikzpicture}
\caption{\emph{Various configurations of the optimal sequence within a bin $[i_s,i_t]$ with $\Delta s_i = 0$. The leaf nodes indicate the labels of the paragraphs in the Proof of Theorem \ref{thm:main-sq} to handle each scenario.}}
\label{fig:sq-f1}
\end{figure}

\begin{figure}[H]
\centering
\begin{tikzpicture}[
    node distance = 12mm and 6mm,
       box/.style = {rectangle, draw, fill=#1, 
                     minimum width=12mm, minimum height=7mm}
                        ]
\node (delta) [box=blue!10] {$\Delta s_i \neq 0$};
\node (L1) [box=green!10, below = of delta] {\textbf{case (a)}};
\draw[->] (delta) to  (L1);
\end{tikzpicture}
\caption{\emph{A configuration of optimal sequence within a bin $[i_s,i_t]$ with $|\Delta s_i| \neq 0$. The leaf node indicate the label of the paragraph in the Proof of Theorem \ref{thm:main-sq} to handle this scenario.}}
\label{fig:sq-f2}
\end{figure}
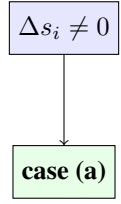

\textbf{Close comparison to lower bound in \citep{arrows2019}.} For the case of 1D forecasting of TV bounded sequences, \citep{arrows2019} consider a stochastic setting where the labels obey $y_t = w_t + \epsilon_t$ for some iid $\sigma$ subgaussian noise $\epsilon_t$ and $w_t \in \mathcal{TV}^B(C_n)$. They provide a lower bound of $\tilde \Omega \left((nB^2 \wedge n\sigma^2 \wedge n^{1/3}C_n^{2/3}\sigma^{4/3})  + (nB^2 \wedge BC_n) + B^2\right)$ where $(a \wedge b) = min \{a , b \}$. In accordance with the proof of Proposition \ref{prop:lb-sq}, we can take $\sigma = B/2$ and $\bs w_{1:n} \in \mathcal{TV}^{B/2}(C_n)$ to translate this lower bound into our setting for 1D case to get a lower bound of:
\begin{align}
    R_n(C_n)
    &= \tilde \Omega \left((nB^2 \wedge n^{1/3}C_n^{2/3}B^{4/3})  + (nB^2 \wedge BC_n) + B^2\right). \label{eq:lb1}
\end{align}

Any learner must have to incur $O(B^2)$ loss in the first round. Combining this with the upper bound in Theorem \ref{thm:main-sq} along with the trvial regret bound of $O(nB^2)$ we can get a refined regret upper bound of:
\begin{align}
    R_n(C_n)
    &= \tilde O\left(( nB^2 \wedge n^{1/3}C_n^{2/3}B^{4/3} \right) + B^2 ). \label{eq:ub1}
\end{align}

Comparing Eq.\eqref{eq:lb1} and \eqref{eq:ub1} seems to falsely suggest that during the regime where $n^{1/3}C_n^{2/3}B^{4/3} < BC_n < nB^2$ upper bound in Eq.\eqref{eq:ub1} is smaller than the lower bound in Eq.\eqref{eq:lb1}. But $n^{1/3}C_n^{2/3}B^{4/3} < BC_n$ happens when $C_n > nB$, in which case $BC_n < nB^2$ is not satisfied. Hence we conclude that this regime is not realisable implying no contradictions.

\textbf{Close comparison to lower bound in \citep{yuan2019dynamic}.} Proposition 1 of \citep{yuan2019dynamic} considers squared error losses in 1D and show that when $C_n = n^{\frac{2+\gamma}{4-\gamma}}$ for all $\gamma \in (0,1)$, the dynamic regret obeys
\begin{align}
    R_n(C_n) 
    &= \Omega \left( \log n \vee (nC_n)^{\gamma/2} \right).
\end{align}

We proceed to show that our lower bound of $\Omega(\log n \vee n^{1/3}C_n^{2/3})$ is tighter than this. Whenever $C_n = n^{\frac{2+\gamma}{4-\gamma}}$, we have
\begin{align}
    (nC_n)^{\gamma/2}
    &= n^{\frac{3\gamma}{4-\gamma}},
\end{align}

and,
\begin{align}
    n^{1/3} C_n^{2/3}
    &= n^{\frac{8+\gamma}{12-3\gamma}}.
\end{align}

It can be verified that for all $\gamma \in (0,1)$, $n^{\frac{3\gamma}{4-\gamma}} \le n^{\frac{8+\gamma}{12-3\gamma}}$ making our lower bound tighter.

\section{Proofs for Section \ref{sec:ec}} \label{app:ecd}
\subsection{One dimensional setting}\label{app:ec_1d}
In the section, we adopt all the notations used in Section \ref{sec:sq}. For the sake of simplicity of exposition, we first present the results in one dimensional setting and extend it later to higher dimensions.  We have the following guarantee in one dimension.
\begin{theorem}\label{thm:ec-1d} ($d=1$) 
By using the base learner as ONS with parameter $\zeta = \min \left \{\frac{1}{4G^\dagger(2B+ 2G/\beta)}, \alpha \right \}$ and decision set $\cD$ and  choosing learning rate $\eta = \alpha$, \ALG{} guarantees a dynamic regret $R_n(C_n) = \tilde O \left(n^{1/3}C_n^{2/3}  \vee \log n \right)$.
\end{theorem}

We start the analysis by inspecting the KKT conditions.
\begin{restatable}{lemma}{lemkktec} \label{lem:kkt-ec} (\textbf{characterization of offline optimal}) 
Consider the following convex optimization problem.
\begin{mini!}|s|[2]                   
    {\tilde u_1,\ldots,\tilde u_n, \tilde z_1,\ldots,\tilde z_{n-1}}                               
    {\sum_{t=1}^n f_t(\tilde { u}_t)}   
    {\label{eq:Example1}}             
    {}                                
    \addConstraint{\tilde z_t}{=\tilde u_{t+1} - \tilde u_{t} \: \forall t \in [n-1],}    
    \addConstraint{\sum_{t=1}^{n-1} |\tilde z_t|}{\le C_n, \label{eq:constr-ec-1}}  
    \addConstraint{-B}{\le \tilde u_t \: \forall t \in [n],\label{eq:constr-ec-2}}
    \addConstraint{\tilde u_t}{\le B \: \forall t \in [n],\label{eq:constr-ec-3}}
\end{mini!}
Let $u_1,\ldots,u_n,z_1,\ldots,z_{n-1}$ be the optimal primal variables and let $\lambda \ge 0$ be the optimal dual variable corresponding to the constraint \eqref{eq:constr-ec-1}. Further, let $\gamma_t^- \ge 0, \gamma_t^+ \ge 0$ be the optimal dual variables that correspond to constraints \eqref{eq:constr-ec-2} and \eqref{eq:constr-ec-3} respectively for all $t \in [n]$. By the KKT conditions, we have

\begin{itemize}
    \item \textbf{stationarity: } $\grad f_t({ u}_t) = \lambda \left ( s_t - s_{t-1} \right) +  \gamma^-_t -  \gamma^+_t$, where $s_t \in \partial|z_t|$ (a subgradient). Specifically, $s_t=\sign(u_{t+1}-u_t)$ if $|u_{t+1}-u_t|>0$ and $s_t$ is some value in $[-1,1]$ otherwise. For convenience of notations later, we also define 
    $s_n = s_0 = 0$.
    \item \textbf{complementary slackness: } (a) $\lamda \left(\sum_{t=2}^n |u_t - u_{t-1}| - C_n \right) = 0$; (b)  $ \gamma^-_t ( u_t + B) = 0$ and $ \gamma^+_t ( u_t - B) = 0$ for all $t \in [n]$
\end{itemize}
\end{restatable}

\textbf{Terminology.} We will refer to the optimal primal variables $u_1,\ldots,u_n$ in Lemma \ref{lem:kkt-ec} as the \emph{offline optimal sequence} in this section.

Next, we record an easy corollary of Lemma \ref{lem:part-sq}.

\begin{corollary}(\textbf{key partition}) \label{cor:part-1d}
Assume the notations of Lemma \ref{lem:part-sq}. Create a partition of $\cP$ of $[n]$ with the procedure mentioned in Lemma \ref{lem:part-sq} . Then for any $[i_s,i_t] \in \cP$, we have
\begin{itemize}
    \item (TV constraint) $\sum_{j=i_s+1}^{i_t} | u_j -  u_{j-1} | \le B/\sqrt{n_i}$,
    \item (Bins bound) $M:=|\cP| = O(n^{1/3}C_n^{2/3})$.
    \item (Structural property) If $i_s > 1$ then $u_{i_s} \ne u_{i_s-1}$. Similarly if $i_t < n$ then $u_{i_t} \neq u_{i_t+1}$.
\end{itemize}
\end{corollary}

Now we make an important observation regrading the dual variables $\gamma_j^-$ and $\gamma_j^+$. The following property will be used several times in the proofs to follow.

\begin{lemma} \label{lem:ppt-1d}
Define $\Gamma^+_i := \sum_{j=i_s}^{i_t} \gamma^+_j$ and $\Gamma^-_i := \sum_{j=i_s}^{i_t} \gamma^-_j$. Consider a bin $[i_s,i_t] \in \cP$, where $\cP$ is the partition of $[n]$ constructed in Corollary \ref{cor:part-1d}. Then at-least one of the following is always satisfied.
\begin{itemize}
    \item $\gamma_j^- = 0$ for all $j \in [i_s,i_t]$.
    \item $\gamma_j^+ = 0$ for all $j \in [i_s,i_t]$.
\end{itemize}
Consequently we have $\sum_{j=i_s}^{i_t} |\gamma_j^-| + |\gamma_j^+|  = \left|\Gamma_i^- - \Gamma_i^+ \right|$, for any bin $[i_s,i_t] \in \cP$.
\end{lemma}
\begin{proof}
From the properties of the partition $\cP$ in Corollary \ref{cor:part-1d}, we have that the TV of the offline optimal incurred within each bin is at-most $B/\sqrt{n_i} \le B$. Hence within bin $[i_s,i_t] \in \cP$, if the optimal sequence attains the value $-B$ at some time point, it can never attain the value $B$ and vice-versa. So due to complementary slackness rule in Lemma \ref{lem:kkt-ec}, either $\gamma_j^+ = 0$ or $\gamma_j^- = 0$ uniformly for all $j \in [i_s,i_t]$. The last line in the statement of lemma follows by recalling that $\gamma_j^- \ge 0$ and $\gamma_j^+ \ge 0$ from Lemma \ref{lem:kkt-ec}.

\end{proof}

For convenience, we recall here the regret decomposition of Eq.\eqref{eq:reg-ecm} specified to one dimensional setting. Let $\cP$ be a partition of $[n]$ into $M$ bins as specified in Corollary \ref{cor:part-1d}. Let $[i_s, i_t]$ denote the $i^{th}$ bin in $\cP$ and let $n_i$ be its length.  Define $\bar u_i = \frac{1}{n_i} \sum_{j=i_s}^{i_t} u_j$ and $\dot u_i = \bar u_i - \frac{1}{n_i \beta} \sum_{j=i_s}^{i_t} \grad f_j(\bar u_i)$ where $\beta$ is as in Assumption EC-2. Let $x_j$ be the prediction made by \ALG{} at time $j$. We start with following regret decomposition.
\begin{align}
    R_n(C_n)
    &\le \sum_{i=1}^{M} \underbrace{\sum_{j=i_s}^{i_t} f_j(x_j) - f_j(\dot u_i)}_{T_{1,i}} + \sum_{i=1}^{M} \underbrace{\sum_{j=i_s}^{i_t} f_j(\dot u_i) - f_j(\bar u_i)}_{T_{2,i}} + \sum_{i=1}^{M} \underbrace{\sum_{j=i_s}^{i_t} f_j(\bar u_i) - f_j(u_j)}_{T_{3,i}}. \label{eq:reg-ec}
\end{align}

We proceed to bound the terms $T_{1,i},T_{2,i},T_{3,i}$ for the bins that belong to the partition $\cP$.

\begin{lemma} \label{lem:t1-ec} (\textbf{bounding } $T_{1,i}$)
Let the experts in \ALG{} be the ONS algorithms with parameter $\zeta = \min \left \{\frac{1}{4G^\dagger(2B + 2G)}, \alpha \right \}$ and decision set $\cD$. Also choose learning rate $\eta = \alpha$, for \ALG{}. Then for any bin $[i_s,i_t]$ we have,
\begin{align}
    \sum_{j=i_s}^{i_t} f_j(x_j) - f_j(\dot { u}_i) 
    &= O\left( BG^\dagger \log n + GG^\dagger \log n + \frac{\log n}{\alpha}\right)\\
    &= O(\log n).
\end{align}
\end{lemma}
\begin{proof}
First we proceed to bound $| \dot { u}_i|$.  Since $|\grad f_j(u_j) | \le G$ by Assumption EC-1, we have
\begin{align}
    | \dot {u}_i|
    &\le |\bar {u}_i | + \frac{G}{\beta}\\
    &\le B + G,
\end{align}
since $\beta \ge 1$ by Assumption EC-2.
For any $x \in \mathcal D$, we have $|x -   \dot {u}_i| \le 2B + 2G$ by triangle inequality.

By Assumption EC-4 we have $|\grad f_j(x) | \le G^\dagger$ for any $x \in \mathcal D$. Also, recall that by Assumption EC-3, the loss functions $f_j$ are $\alpha$ exp-concave in the domain $\cD$. Let $p_j$ be the predictions of ONS in the interval $[i_s,i_t]$. If we choose $\zeta = \min \left \{\frac{1}{4G^\dagger(2B + 2G)}, \alpha \right \}$ as the parameter of the ONS, Theorem 2 of \citep{hazan2007logregret} implies that 
\begin{align}
    \sum_{j=i_s}^{i_t} f_j( p_j) - f_j(\dot {u}_i)
    &= O\left( BG^\dagger \log n + GG^\dagger \log n \right)\\
    &= O\left(\log n\right).
\end{align}

Now the Lemma is implied by the SA regret bound of FLH (Theorem 3.2 of \citep{hazan2007adaptive}).

\end{proof}

\begin{lemma} \label{lem:t2-ec} (\textbf{bounding } $T_{2,i}$).
For a bin $[i_s,i_t] \in \cP$, let $C_i, n_i$ and $\Delta s_i$ be as in Lemma \ref{lem:t2-sq} and $\Gamma^+_i, \Gamma^-_i$ be as in Lemma \ref{lem:ppt-1d}. We have
\begin{align}
    \sum_{j=i_s}^{i_t} f_j(\dot u_i) - f_j(\bar u_i)
    &\le \frac{-\left (\lambda\Delta s_i + \Gamma^-_i - \Gamma^+_i \right)^2}{2n_i \beta} + \lambda |\Delta s_i| C_i + |\Gamma^-_i - \Gamma^+_i | C_i.
\end{align}
\end{lemma}
\begin{proof}
We start with the short proof the descent lemma. Let $g(x)$ be a $L$ strongly smooth function. Let $x^+ = x - \mu \grad f(x)$ for some $\mu > 0$. Then we have
\begin{align}
    g(x^+) - g(x)
    &\le (\grad g(x))^2 \left(\frac{L}{2} \mu^2  - \mu \right)\\
    &= \frac{-(\grad g(x))^2}{2L},
\end{align}
by choosing $\mu = 1/L$. By taking $g(x) = \sum_{j=i_s}^{i_t} f_j(x)$ and noting that $g$ is $n_i \beta$ gradient Lipschitz due to Assumption EC-2, we get
\begin{align}
    T_{2,i}
    &:=\sum_{j=i_s}^{i_t} f_j(\dot u_i) - f_j(\bar u_i)\\
    &\le \frac{-\left(\sum_{j=i_s}^{i_t} \grad f_j(\bar u_i) \right)^2}{2 n_i \beta}\\
    &= \frac{-1}{2n_i \beta} \left(\sum_{j=i_s}^{i_t} \grad f_j( u_j) + \grad f_j(\bar u_i) - \grad f_j( u_j)\right)^2\\
    &\le \frac{-1}{2n_i \beta} \left(\sum_{j=i_s}^{i_t} \grad f_j(u_j) \right)^2 + \frac{1}{n_i \beta} \left|\sum_{j=i_s}^{i_t} \grad f_j(u_j) \right| \left| \sum_{j=i_s}^{i_t} \grad f_j(\bar u_i) - \grad f_j(u_j) \right|.
\end{align}
From the KKT conditions in Lemma \ref{lem:kkt-ec} we have $\sum_{j=i_s}^{i_t} \grad f_j(u_j) = \lambda \Delta s_i + \Gamma_i^- - \Gamma_i^+$. Since $f_j$ are $\beta$-gradient Lipschitz and $|\bar u_i - u_j | \le C_i$, we also have 
\begin{align}
  \left| \sum_{j=i_s}^{i_t} \grad f_j(\bar u_i) - \grad f_j(u_j) \right| &\le  n_i \beta C_i.
\end{align}
Substituting these we get,
\begin{align}
    T_{2,i}
    &\le \frac{-\left (\lambda\Delta s_i + \Gamma^-_i - \Gamma^+_i \right)^2}{2n_i \beta} + \lambda |\Delta s_i| C_i + |\Gamma^-_i - \Gamma^+_i | C_i.
\end{align}
\end{proof}

\begin{figure}[h!]
\begin{minipage}[t]{0.48\textwidth}
\includegraphics[width=\linewidth,keepaspectratio=true]{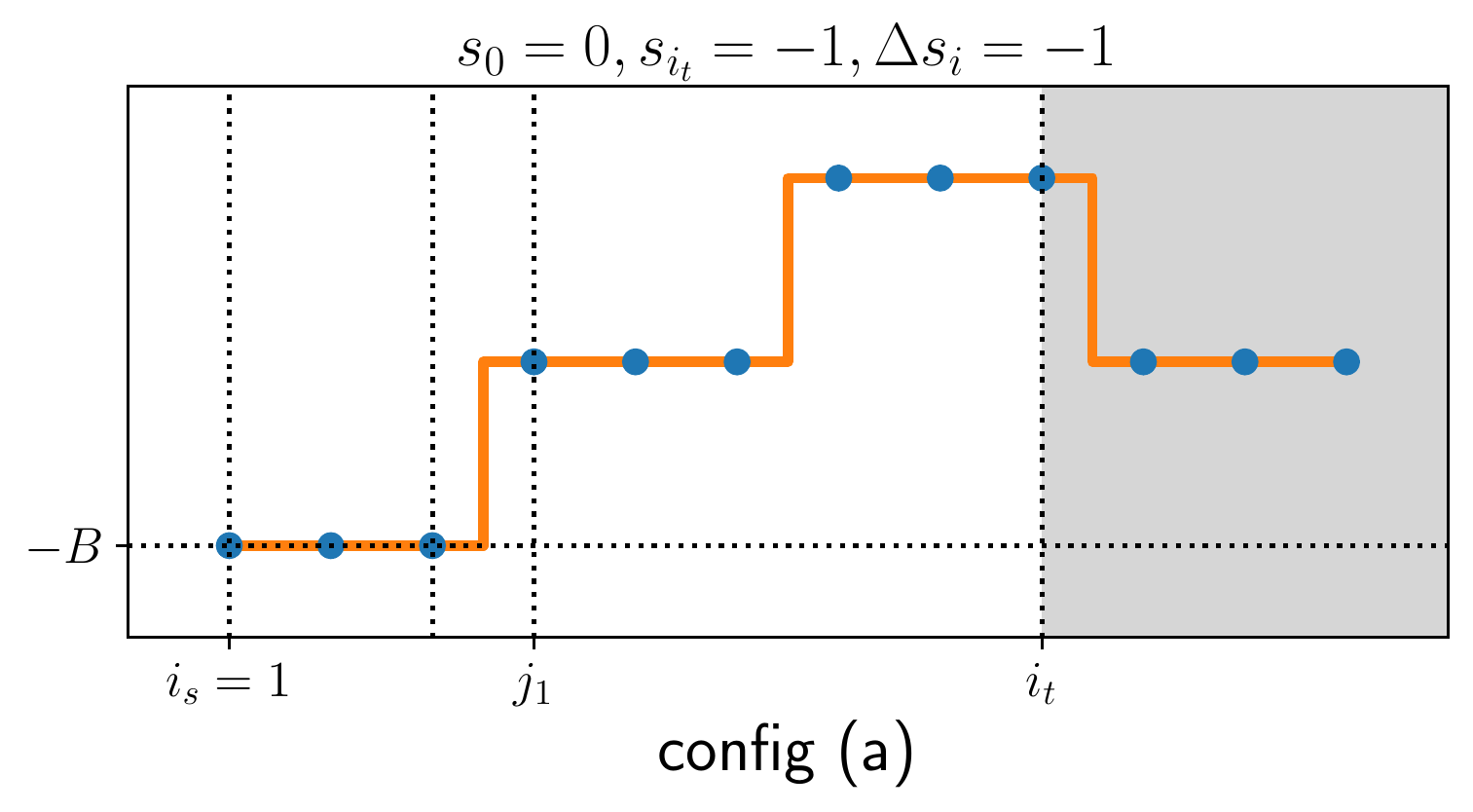}
\end{minipage}
\hspace*{\fill} 
\begin{minipage}[t]{0.48\textwidth}
\includegraphics[width=\linewidth,keepaspectratio=true]{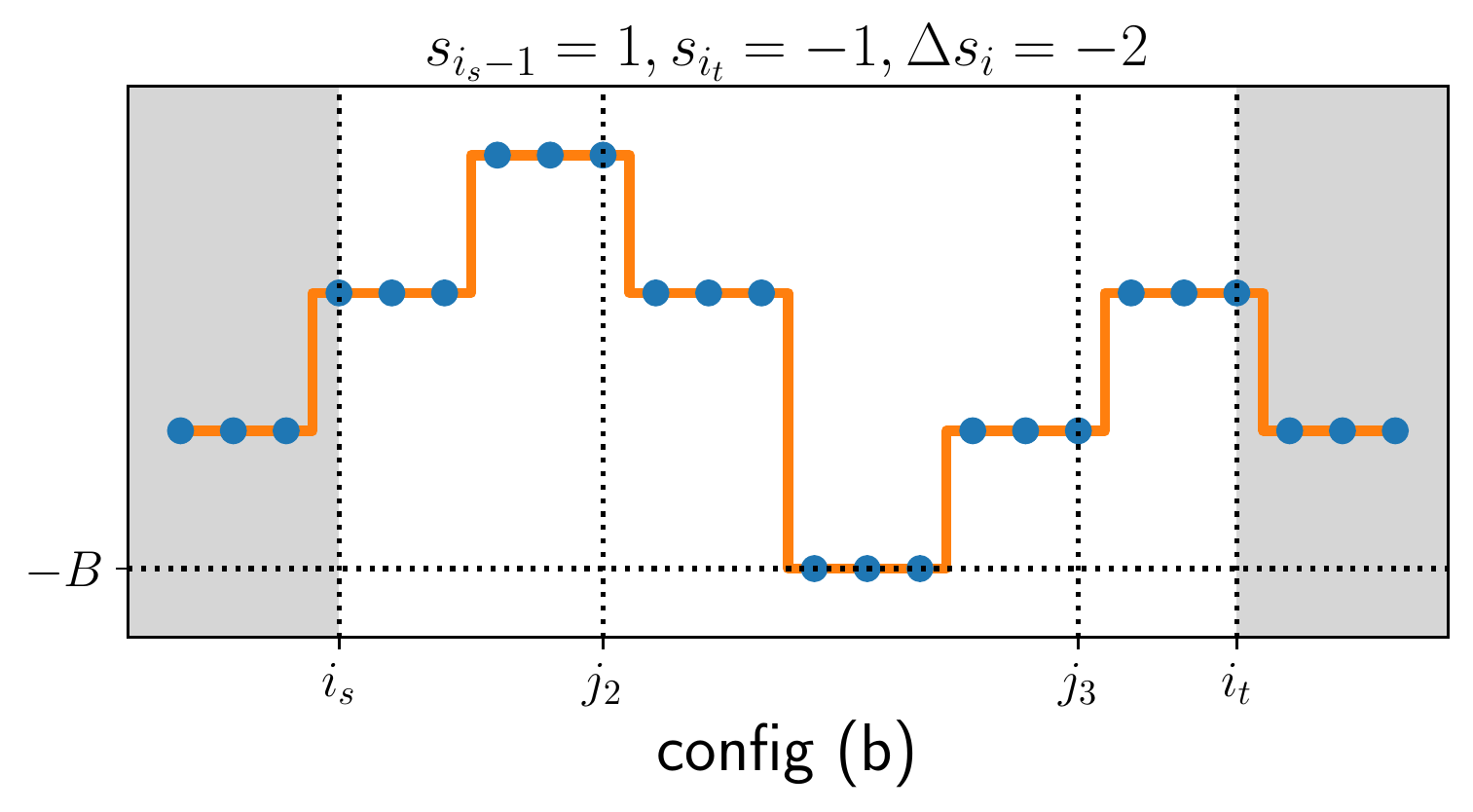}
\end{minipage}
\caption{\emph{Examples of configurations referred in the proof of Theorem \ref{thm:ec-1d}. The blue dots corresponds to the offline optimal sequence.}}
\label{fig:ec}
\end{figure}

\begin{lemma} \label{lem:t3-ec} (\textbf{bounding }$T_{3,i}$) For a bin $[i_s,i_t] \in \cP$, let $C_i, n_i$ and $\Delta s_i$ be as in Lemma \ref{lem:t2-sq} and $\Gamma^+_i, \Gamma^-_i$ be as in Lemma \ref{lem:ppt-1d}.\\
\textbf{case(a)} If $|\Delta s_i| > 0$ then we have,
\begin{align}
    \sum_{j=i_s}^{i_t} f_j(\bar u_i) - f_j(u_j)
    &\le  \frac{\beta n_i C_i^2}{2} + 3\lambda C_i + |\Gamma^-_i - \Gamma^+_i | C_i.
\end{align}
\textbf{case(b)} If $\Delta s_i = 0$ with $s_{i_s-1} = s_{i_t} = 1$ and the offline optimal $\bs u$ is non-decreasing within bin $i$ with $-B < u_i < B$ for all $i \in [i_s,i_t]$, then
\begin{align}
    \sum_{j=i_s}^{i_t} f_j(\bar u_i) - f_j(u_j)
    &\le  \frac{\beta n_i C_i^2}{2}.
\end{align}
\textbf{case(c)} If $\Delta s_i = 0$ with $s_{i_s-1} = s_{i_t} = -1$ and the offline optimal $\bs u$ is non-increasing within bin $i$ with $-B < u_i < B$ for all $i \in [i_s,i_t]$, then
\begin{align}
    \sum_{j=i_s}^{i_t} f_j(\bar u_i) - f_j(u_j)
    &\le  \frac{\beta n_i C_i^2}{2}.
\end{align}

\end{lemma}
\begin{proof}
Due to strong smoothness, we have
\begin{align}
    T_{3,i}
    &:=\sum_{j=i_s}^{i_t} f_j(\bar u_i) - f_j(u_j)\\
    &\le \sum_{j=i_s}^{i_t} \grad f_j(u_j) (\bar u_i - u_j) + \frac{\beta}{2} (\bar u_i - u_j)^2\\
    &\le \frac{\beta n_i C_i^2}{2} + \sum_{j=i_s}^{i_t} \grad f_j(u_j) (\bar u_i - u_j).
\end{align}
Now by expanding the second term and using the structure of gradients as in Lemma \ref{lem:kkt-ec} followed by a regrouping of the terms in the summation we can write,
\begin{align}
    \sum_{j=i_s}^{i_t} \grad f_j(u_j) (\bar u_i - u_j)
    &= \lambda \left( s_{i_s-1} (u_{i_s} - \bar u_i) - s_{i_t} (u_{i_t} - \bar u_i)\right) + \lambda \sum_{j=i_s+1}^{i_t} |u_j - u_{j-1}|\\
    &\quad + \sum_{j=i_s}^{i_t} (\gamma_j^- - \gamma_j^+)(\bar u_i - u_j)\\
    &\le \lambda \left( s_{i_s-1} (u_{i_s} - \bar u_i) - s_{i_t} (u_{i_t} - \bar u_i)\right) + \lambda C_i + |\Gamma^-_i - \Gamma^+_i | C_i, \label{eqn:inter}
\end{align}
where the last line follows due to Lemma \ref{lem:ppt-1d} and $|\bar u_i - u_j| \le C_i$ for all $j \in [i_s,i_t]$.

Now we consider three cases in the statement of the lemma.

\textbf{case (a) }When $|\Delta s_i| > 0$, then by triangle inequality we have \\
$\lambda \left( s_{i_s-1} (u_{i_s} - \bar u_i) - s_{i_t} (u_{i_t} - \bar u_i)\right) \le 2\lambda C_i$.

\textbf{case (b) }In this case we have\\
$\lambda \left( s_{i_s-1} (u_{i_s} - \bar u_i) - s_{i_t} (u_{i_t} - \bar u_i)\right) = \lambda (u_{i_s} - u_{i_t}) =  -\lambda C_i$ since the sequence is non-decreasing within the bin. Hence this term cancels with the corresponding additive term  of  $\lamda C_i$ in Eq. \eqref{eqn:inter}. Further $\gamma_j^- = \gamma_j^+ = 0$ since $-B < u_j < B$ for all $j \in [i_s,i_t]$.

\textbf{case (c)} By similar logic as in case (b) we can once again write\\
$\lambda \left( s_{i_s-1} (u_{i_s} - \bar u_i) - s_{i_t} (u_{i_t} - \bar u_i)\right) =  -\lambda C_i$.

Putting everything together now yields the lemma.

\end{proof}
\begin{proof}\textbf{of Theorem} \ref{thm:ec-1d}.
The strategy of the proof is to bound the regret incurred within each time interval $[i_s,i_t] \in \cP$ where $\cP$ is as in Corollary \ref{cor:part-1d} and add them up towards the end. We annotate several key paragraphs for the purposes of referring the arguments contained in them at later points.

If the the partition $\cP$ contains only one bin, then we split it into at-most two bins $[1,a]$ and $[a+1,n]$ such that the optimal sequence is constant within $[1,a]$ and hence regret incurred within this bin is $\tilde O(1)$ by Strong Adaptivity of FLH. The regret incurred in the bin $[a+1,i_t]$ can be bounded by using the arguments below. So in what follows we assume for a bin $[i_s,i_t]$ either $i_s >1$ or $i_t < n$.

By virtue of Lemma \ref{lem:ppt-1d}, any bin $[i_s,i_t] \in \cP$ will have either $\gamma^-_j = 0$ for all $j \in [i_s,i_t]$ or $\gamma^+_j = 0$ for all $j \in [i_s,i_t]$. Below we bound the regret for bins with $\gamma^+_j = 0$ uniformly for all $j \in [i_s,i_t]$. The arguments for the alternate case where $\gamma^-_j = 0$ follows similarly. Figures \ref{fig:ec1-f1}, \ref{fig:ec1-f2} and \ref{fig:ec1-f3} sketch the floor plan of the proof pictorially. Throughout the proof, we will use the properties in Corollary \ref{cor:part-1d} in conjunction with the observations in Remark \ref{rmk:deltas-lb}. 

\textbf{\setword{(S1)}{Word:S1}:} Consider a bin with $\Delta s_i = 0$ with $s_{i_t} = s_{i_s-1} = 1$ and the optimal sequence is non-decreasing within the bin. By the structural property of Corollary \ref{cor:part-1d}, this happens when $u_{i_s} > u_{i_s-1}$, $u_{i_t+1} > u_{i_t}$ where $1 < i_s < i_t < n$. Since the sequence is non-decreasing, it never attains $-B$ within this bin. Hence this is the same situation as in case (b) of Lemma \ref{lem:t3-ec}. We have $T_{1,i} = \tilde O(1)$ due to Lemma \ref{lem:t1-ec}. $T_{2,i} = 0$ due to Lemma \ref{lem:t2-ec} as $\Gamma^+_i = \Gamma^-_j = 0$ since the sequence never attains $\pm B$ within the current bin combined with the fact that $\Delta s_i = 0$. $T_{3,i} = O(1)$ due to Lemma \ref{lem:t3-ec} combined with the fact that $C_i \le B/\sqrt{n_i}$ due to Corollary \ref{cor:part-1d}. So the total regret within the current bin is bounded by $T_{1,i}+T_{2,i}+T_{3,i} = \tilde O(1)$.

The total regret for a bin satisfying case (c) of Lemma \ref{lem:t3-ec} can be bound using similar arguments as above.

The three cases where (i) $\Delta s_i = 0$ with $s_{i_s-1} = s_{i_t} = -1$ and the offline optimal $\bs u$ is non-decreasing within bin $i$; (ii) $\Delta s_i = 0$ with $s_{i_s-1} = s_{i_t} = 1$ and the offline optimal $\bs u$ is non-increasing within bin $i$ and (iii) $\Delta s_i = 0$ and $\bs u$ is not monotonic will be covered shortly in the arguments to follow.

Consider a bin with $|\Delta s_i| > 0$ and $\gamma^+_j = 0$ uniformly. From Lemmas \ref{lem:t2-ec} and \ref{lem:t3-ec} and using the fact that $|\Delta s_i| \le 2$ we have,
\begin{align}
    T_{2,i} + T_{3,i}
    &\le \frac{\beta n_i C_i^2}{2} + \underbrace{\frac{-\lamda^2 (\Delta s_i)^2}{2n_i \beta} + 7 \lamda C_i}_{(1)} + \underbrace{ \frac{\left(- \Gamma^-_i\right)^2}{2n_i\beta} + 2 \Gamma^-_i C_i}_{(2)} - \frac{\lamda \Delta s_i \Gamma^-_i}{n_i \beta}.
\end{align}
By completing the squares with the terms (1) and (2) in the above display and dropping the negative terms, we get
\begin{align}
    T_{2,i} + T_{3,i}
    &\le \frac{\beta n_i C_i^2}{2} + \frac{49C_i^2 n_i \beta}{2(\Delta s_i)^2} + 2\beta n_i C_i^2 - \frac{\lamda \Delta s_i \Gamma^-_i}{n_i \beta}\\
    &\le 27 B^2 \beta - \frac{\lamda \Delta s_i \Gamma^-_i}{n_i \beta}, \label{eq:drop-squares}
\end{align}
where in last line we used the facts that $C_i \le B/\sqrt{n_i}$ by Corollary \ref{cor:part-1d} and $|\Delta s_i| > 1$ whenever $|\Delta s_i| \neq 0$ by Remark \ref{rmk:deltas-lb}. 

Define $T_i:= \sum_{j=i_s}^{i_t} f_j(x_j) - f_j(u_j)$. Notice that:
\begin{itemize}
    \item \textbf{\setword{(A1)}{Word:A1}:} When $\Gamma^-_i = 0$ and $|\Delta s_i| > 0$, combining Lemma \ref{lem:t1-ec} we have $T_i = \tilde O(1)$;
    \item \textbf{\setword{(A2)}{Word:A2}:} Similarly when $\Delta s_i > 0$, we get $T_i = \tilde O(1)$ as $\Gamma^-_i \ge 0$ by Lemma \ref{lem:kkt-ec}.
\end{itemize}
   In what follows, we try to split an original bin $[i_s,i_t]$ with $\Delta s_i < 0$ into sub-bins that satisfy the above conditions \ref{Word:A1} or \ref{Word:A2}.

If optimal sequence is uniformly constant, we can appeal to the static regret guarantee of FLH to get logarithmic regret over $n$ rounds. So we assume that the optimal sequence is not constant uniformly in the analysis below.

Next, we consider the case when $\Delta s_i < 0$. We start with the following observation.

\textbf{\setword{(B1)}{Word:B1}:} Consider a bin $[i_s,i_t]$ that satisfies the structural property in Corollary \ref{cor:part-1d}. When either $i_s >1$ or $i_t < n$ and $\Delta s_i < 0$, then $s_{i_t} \in \{-1,0\}$ and $s_{i_s-1} \in \{0,1 \}$ with at-least one of them being non-zero.

Since by our assumption $|\cP| >1$, $i_s$ and $i_t$ can't be 1 and $n$ simultaneously. So for any bin $[i_s,i_t]$ with $\Delta s_i < 0$, observation \ref{Word:B1} has to be satisfied. 

When $\Delta s_i < 0$, we can have three cases as follows. 

\textbf{\setword{Case (1)}{Word:case-1}:} If the optimal solution is constant (i.e $C_i = 0$) within the bin $i$. Then we trivially get $T_i = \tilde O(1)$.

\textbf{\setword{Case (2)}{Word:case-2}:} If the optimal solution is monotonic within bin $i$ (see config (a) in Fig.\ref{fig:ec} for an example of this configuration). Then we split the original bin $[i_s,i_t]$ into at-most 2 bins. Let $j_1,j_2$ be such that  $u_m = -B \forall m \in [i_s,j_1-1] \cup [j_2+1,i_t]$ and $u_{j_1} > -B, u_{j_2} > -B$. If $u_{i_s} > -B$, then $j_1 = i_s$ and $[i_s,j_1-1]$ is viewed as an empty interval. Similar logic applies for the right interval $[j_2+1,i_t]$. Since the optimal sequence is monotonic within $[i_s,i_t]$, either $j_1 = i_s$ or $j_2 = i_t$. Without loss of generality let's assume that $j_2 = i_t$. We proceed to bound the regret incurred within each of the two sub-bins separately.

Let's annotate bin $[i_s,j_1-1]$ by $i^{(1)}$ and bin $[j_1,i_t]$ by $i^{(2)}$. For the bin $i^{(1)}$, the optimal solution is constant and hence the regret $T_{i^{(1)}} = \tilde O(1)$. For the bin $i^{(2)}$, notice that $\gamma^-_j = 0 \: \forall j \in [j_1,i_t]$ since the sequence is monotonic with $u_{j_1} > -B$ and since our assumption $j_2 = i_t$ implies $u_{j_2} > -B$. Hence we have $\Gamma^-_{i^{(2)}} = 0$. Since $s_{j_1-1} \in \{0,1\}$ and  by observation \ref{Word:B1}, $s_{i_t} \in \{-1, 0\}$  with at-least one of them being non-zero, we have $|\Delta s_i| \neq 0$. Hence the bin $i^{(2)}$ falls into the category \ref{Word:A1}. So $T_{i^{(2)}} = \tilde O(1)$. Adding the regret incurred in each sub-bin separately yields $T_i = \tilde O(1)$. 

\textbf{\setword{Case (3)}{Word:case-3}:} Consider the alternate case where we have $\Delta s_i < 0$ and the sequence is not monotonic (see config (b) in Fig. \ref{fig:ec} for an example of this configuration). We split the original bin $[i_s,i_t]$ into at-most three sub-bins $[i_s,j_1-1], [j_1,j_2], [j_2+1,i_t]$ such that (i) If $u_{i_s} = -B$, then $u_m = -B \: \forall m \in [i_s,j_1-1]$ and $u_{j_1} > -B$. If $u_{i_s} > -B$, then we take $j_1 = i_s$ and view $[i_s,j_1-1]$ as empty interval. (ii) $j_2$ is the smallest point in $[j_1,i_t]$ such that $s_{j_2} = -1$ and $u_{j_2} > u_{j_2+1}$.

Let's annotate bins $[i_s,j_1-1], [j_1,j_2], [j_2+1,i_t]$ by $i^{(1)},i^{(2)},i^{(3)}$ respectively. If bin $i^{(1)}$ is not empty, then we have $T_{i^{(1)}} = \tilde O(1)$ since $\bs u$ is constant within that bin.

Since $\Delta s_i < 0$, we must have $s_{j_1-1} \in \{ 0, 1\}$ even if $j_1 = i_s$. By construction the sequence $\bs u$ never attains the value $-B$ in the bin $i^{(2)}$ since $u_{j_1} > -B$ and $j_2$ is the first time point since $j_1$ after which the optimal sequence jumps downwards. So we have $\Gamma_{i^{(2)}} = 0$. Further we also have $|\Delta s_{i^{(2)}}| > 0$ within bin $i^{(2)}$. So we get $T_{i^{(2)}} = \tilde O(1)$ since $i^{(2)}$ falls into category \ref{Word:A1}

For simplicity let's assume that $u_{i_t} > -B$, otherwise we can create another bin that ends at time $i_t$ where optimal solution assumes a constant value of $-B$ and proceed with similar arguments as before to bound the regret in the constant interval.

\textbf{\setword{(S2)}{Word:S2}:} If the sequence $\bs u$ is not monotonic in $i^{(3)}$, we split the bin $i^{(3)}$ into two parts $[j_2+1, j_3], [j_3+1,i_t]$ such that $j_3$ is the largest point in $[j_2+1,i_t]$ with $s_{j_3} = 1$ and $u_{j_3} < u_{j_3+1}$. Let's annotate the bins $[j_2+1, j_3], [j_3+1,i_t]$ by $q^{(1)},q^{(2)}$ respectively. We have $\Delta s_{q^{(1)}} > 0$ since $s_{j_3} = 1$ and $s_{j_2} = -1$. Hence the bins $q^{(1)}$ falls into the category\ref{Word:A2} mentioned before and we get $T_{q^{(1)}} = \tilde O(1)$. Notice that $s_{i_t} \in \{ -1,0\}$ as $\Delta s_i < 0$. Since $j_3$ is the largest point in $[j_2+1,i_t]$ with $s_{j_3} = 1$ and it is assumed before that $u_{i_t} > -B$, we conclude that the sequence in the interval $q^{(2)}$ is a non-increasing sequence that never attains the value $-B$. So $\Gamma^-_{q^{(2)}} = 0$. Further we have  $|\Delta s_{q^{(2)}}| > 1$ . So $T_{q^{(2)}} = \tilde O(1)$ since $q^{(2)}$ falls into the category \ref{Word:A1}. We pause to remark that the arguments we used to bound the regret in the bin $i^{(3)}$ can be used to bound the regret of any bin $[r_s,r_t] \in \cP$ with $\Delta s_r = 0$ and the sequence $\bs u$ being not monotonic within bin $r$.

Note that since $u_{j_2+1} < u_{j_2}$, bin $i^{(3)}$ satisfies the structural property of Corollary \ref{cor:part-1d}. So if the sequence $\bs u$ is non-increasing in bin $i^{(3)}$ and $s_{i_t} = -1$, it fits into case (c) of Lemma \ref{lem:t3-ec}. So we can bound $T_{i^{(3)}} = \tilde O(1)$ using arguments presented in \ref{Word:S1}.

If the sequence $\bs u$ is monotonic in bin $i^{(3)}$ and $s_{i_t} = 0$ (which happens when $i_t = n$), then we have $\Delta s_{i^{(3)}} = 0 - (-1) = 1 > 0$. So bin $i^{(3)}$ falls into the category\ref{Word:A2} mentioned before. Hence the regret $T_{i^{(3)}} = \tilde O(1)$.

\textbf{\setword{(S3)}{Word:S3}:} If the sequence $\bs u$ is non-decreasing in bin $i^{(3)}$, we split the bin into two intervals $[j_2+1,k], [k+1,i_t]$ such that $k$ is any point in $[j_2+1,i_t]$ with $s_{k} = 1$ and $u_{k+1} > u_k$. (This configuration is similar to that of config (a) in Fig.\ref{fig:sq}). Annotate $[j_2+1,k], [k+1,i_t]$ by $q^{(1)},q^{(2)}$ respectively. In bin $q^{(1)}$ we have $\Delta s_{q^{(1)}} = 2$ and hence $T_{q^{(1)}} = \tilde O(1)$ since $q^{(1)}$ falls into the category \ref{Word:A2}. Within bin $q^{(2)}$ due to the assumption that $u_{i_t} > -B$, we have $\Gamma^-_{q^{(2)}} = 0$. We also have $|\Delta s_{q^{(2)}}| > 0$ and consequently $q^{(2)}$ falls into category \ref{Word:A1}. So we have $T_{q^{(2)}} = \tilde O(1)$. We pause to remark that the arguments we used to bound the regret in bin $i^{(3)}$ for the case where $\bs u$ is non-decreasing, can also be used to bound the regret of any bin  $[r_s,r_t]$ with $\Delta s_r = 0$ and $s_{r_t} = s_{r_s-1} = -1$ and the sequence $\bs u$ is non-decreasing. The regret for the alternate case where $\Delta s_r = 0$ and $s_{r_t} = s_{r_s-1} = 1$ and the sequence $\bs u$ is non-increasing can be bounded similarly using a mirrored argument.

So summarizing, in case (3) we get $T_i = \tilde O(1)$. Since the intermediate splitting operations can only increase the number of bins to at-most $6M$, adding the regret across all $O(M)$ bins in Corollary \ref{cor:part-1d} yields the Theorem.

\end{proof}

\begin{figure}
\centering
\begin{tikzpicture}[
    node distance = 12mm and 6mm,
       box/.style = {rectangle, draw, fill=#1, 
                     minimum width=12mm, minimum height=7mm}
                        ]
\node (not-mono) [box=blue!10] {Not monotonic};
\node (delta) [box=blue!10, above right = of not-mono, xshift = 50] {$\Delta s_i = 0$};
\node (mono) [box=blue!10,below right=of delta, xshift=50] {Monotonic};
\node (a1) [align=center, box=blue!10,below  left=of mono, xshift= -50] {$s_{i_t} = 1$ \\ $s_{i_s-1} = 1$\\ Non-decreasing};
\node (a2) [align=center, box=blue!10, right=of a1] {$s_{i_t} = -1$ \\ $s_{i_s-1} = -1$\\ Non-increasing};
\node (a3) [align=center, box=blue!10, right=of a2] {$s_{i_t} = -1$ \\ $s_{i_s-1} = -1$\\ Non-decreasing};
\node (a4) [align=center, box=blue!10, right=of a3] {$s_{i_t} = 1$ \\ $s_{i_s-1} = 1$\\ Non-increasing};
\node (L1) [align=center, box=green!10, below=of not-mono] {Similar to \ref{Word:S2}};
\node (L2) [align=center, box=green!10, below=of a1] {\ref{Word:S1}};
\node (L3) [align=center, box=green!10, below=of a2] {Similar to \ref{Word:S1}};
\node (L4) [align=center, box=green!10, below=of a3] {Similar to \ref{Word:S3}};
\node (L5) [align=center, box=green!10, below=of a4] {Mirrored \ref{Word:S3}};
\draw[->] (delta) to  (not-mono);
\draw[->] (delta) to  (mono);
\draw[->] (mono) to  (a1);
\draw[->] (mono) to  (a2);
\draw[->] (mono) to  (a3);
\draw[->] (mono) to  (a4);
\draw[->] (not-mono) to  (L1);
\draw[->] (a1) to  (L2);
\draw[->] (a2) to  (L3);
\draw[->] (a3) to  (L4);
\draw[->] (a4) to  (L5);

\end{tikzpicture}
\caption{\emph{Various configurations of the optimal sequence within a bin $[i_s,i_t]$ with $\Delta s_i = 0$. The leaf nodes indicate the arguments used in the proof of Theorem \ref{thm:ec-1d} to handle each scenario.}}
\label{fig:ec1-f1}
\end{figure}

\begin{figure}[H]
\centering
\begin{tikzpicture}[
    node distance = 12mm and 6mm,
       box/.style = {rectangle, draw, fill=#1, 
                     minimum width=12mm, minimum height=7mm}
                        ]
\node (zero) [box=blue!10] {$\Gamma_i^- = 0$};
\node (delta) [box=blue!10, above right = of zero, xshift = 50] {$\Delta s_i < 0$};
\node (non-zero) [box=blue!10,below right=of delta, xshift=50] {$\Gamma_i^- > 0$};
\node (a1) [align=center, box=blue!10,below  left=of non-zero] {constant};
\node (a2) [align=center, box=blue!10,  right=of a1] {Monotonic};
\node (a3) [align=center, box=blue!10,  right=of a2] {Not monotonic};
\node (L1) [align=center, box=green!10, below=of zero] {\ref{Word:A1}};
\node (L2) [align=center, box=green!10, below=of a1] {\ref{Word:case-1}};
\node (L3) [align=center, box=green!10, below=of a2] {\ref{Word:case-2}};
\node (L4) [align=center, box=green!10, below=of a3] {\ref{Word:case-3}};
\draw[->] (delta) to  (zero);
\draw[->] (delta) to  (non-zero);
\draw[->] (non-zero) to  (a1);
\draw[->] (non-zero) to  (a2);
\draw[->] (non-zero) to  (a3);
\draw[->] (zero) to  (L1);
\draw[->] (a1) to  (L2);
\draw[->] (a2) to  (L3);
\draw[->] (a3) to  (L4);
\end{tikzpicture}
\caption{\emph{Various configurations of optimal sequence within a bin $[i_s,i_t]$ with $\Delta s_i < 0$. The leaf nodes indicate the arguments used in the proof of Theorem \ref{thm:ec-1d} to handle each scenario.}}
\label{fig:ec1-f2}
\end{figure}

\begin{figure}[H]
\centering
\begin{tikzpicture}[
    node distance = 12mm and 6mm,
       box/.style = {rectangle, draw, fill=#1, 
                     minimum width=12mm, minimum height=7mm}
                        ]
\node (delta) [box=blue!10] {$\Delta s_i > 0$};
\node (L1) [box=green!10, below = of delta] {\ref{Word:A2}};
\draw[->] (delta) to  (L1);
\end{tikzpicture}
\caption{\emph{A configuration of optimal sequence within a bin $[i_s,i_t]$ with $\Delta s_i > 0$. The leaf node indicate the arguments used in the proof of Theorem \ref{thm:ec-1d} to handle each scenario.}}
\label{fig:ec1-f3}
\end{figure}

\subsection{Multi dimensional setting}\label{app:ec_highdim}
We start by inspecting the KKT conditions.
\begin{lemma} \label{lem:kkt-ec-d} (\textbf{characterization of offline optimal}) 
Consider the following convex optimization problem.
\begin{mini!}|s|[2]                   
    {\tilde{ \bs u}_1,\ldots,\tilde{ \bs u}_n,\tilde{ \bs z}_1,\ldots,\tilde{ \bs z}_{n-1}}                               
    {\sum_{t=1}^n f_t(\tilde {\bs u}_t)}   
    {\label{eq:Example1}}             
    {}                                
    \addConstraint{\tilde{ \bs z}_t}{=\tilde{ \bs u}_{t+1} - \tilde{ \bs u}_{t} \: \forall t \in [n-1],}    
    \addConstraint{\sum_{t=1}^{n-1} \|\tilde{ \bs z}_t\|_1}{\le C_n, \label{eq:constr-ec-1d}}  
    \addConstraint{\| \tilde {\bs u}_t\|_\infty}{\le B \: \forall t \in [n],\label{eq:constr-ec-2d}}
\end{mini!}

Let $\bs u_1,\ldots,\bs u_n,\bs z_1,\ldots,\bs z_{n-1} \in \mathbb{R}^d$ be the optimal primal variables and let $\lambda \ge 0$ be the optimal dual variable corresponding to the constraint \eqref{eq:constr-ec-1d}. Further, let $\bs \gamma_t^+, \bs \gamma_t^- \in \mathbb{R}^d$ with $\bs \gamma_t^+ \ge 0$ and $\bs \gamma_t^- \ge 0$ be the optimal dual variables that correspond to constraint \eqref{eq:constr-ec-2d}. Specifically for $k \in [d]$, $\bs \gamma_t^+[k]$ corresponds to the dual variable for the constraint $\bs u_t[k] \le B$ induced by the relation \eqref{eq:constr-ec-2d}. Similarly $\bs \gamma_t^-[k]$ corresponds to the constraint $-B \le \bs u_t[k]$. By the KKT conditions, we have

\begin{itemize}
    \item \textbf{stationarity: } $\grad f_t({\bs u}_t) = \lambda \left ( \bs s_t - \bs s_{t-1} \right) +  \bs \gamma^-_t -  \bs \gamma^+_t$, where $\bs s_t \in \partial|\bs z_t|$ (a subgradient). Specifically, $\bs s_t[k]=\sign(\bs u_{t+1}[k]-\bs u_t[k])$ if $|\bs u_{t+1}[k]-\bs u_t[k]|>0$ and $\bs s_t[k]$ is some value in $[-1,1]$ otherwise. For convenience of notations later, we also define 
    $\bs s_n = \bs s_0 = \bs 0$.
    \item \textbf{complementary slackness: } (a) $\lamda \left(\sum_{t=2}^n \|\bs u_t - \bs u_{t-1}\|_1 - C_n \right) = 0$; (b)  $ \bs \gamma^-_t[k] ( \bs u_t[k] + B) = 0$ and $ \bs \gamma^+_t[k] ( \bs u_t[k] - B) = 0$ for all $t \in [n]$ and all $k \in [d]$.
\end{itemize}
\end{lemma}

The proof of the above lemma is similar to the 1D case and hence omitted.

\noindent\textbf{Terminology.} We will refer to the optimal primal variables $\bs u_1,\ldots,\bs u_n$ in Lemma \ref{lem:kkt-ec-d} as the \emph{offline optimal sequence} in this section.

Next, we claim the existence of a partitioning of $[n]$ with some useful properties.
\begin{lemma} \label{lem:part-ec-d} (\textbf{key partition})
There exist a partitioning $\cP$ of $[n]$ into $M = O(d n^{1/3}C_n^{2/3})$ intervals viz $\{[i_s,i_t]\}_{i=1}^M$ such that for any interval $[i_s,i_t] \in \cP$, $C_i \le B/\sqrt{n_i}$ where $C_i := \sum_{j=i_s+1}^{i_t} \|\bs u_j - \bs u_{j-1}\|_1$ and $n_i$ is the length of the interval. 

Define $\bs \Gamma^+_i := \sum_{j=i_s}^{i_t} \bs \gamma^+_j$ and $\bs \Gamma^-_i := \sum_{j=i_s}^{i_t} \bs \gamma^-_j$. Let $\Delta \bs s_i = \bs s_{i_t} - \bs s_{i_s - 1}$, where $\bs s$ is as defined in Section \ref{sec:proof-main-sq}. We also have that each bin $[i_s,i_t] \in \cP$ satisfies at-least one of the following properties.
\begin{enumerate}
    \item[Property 1] Across each coordinate $k \in [d]$, the sequence $\bs u_j[k], j \in [i_s,i_t]$ is either non-decreasing or non-increasing.
    \item[Property 2] $ \|\lambda \Delta \bs s_i + \bs \Gamma^-_i - \bs \Gamma^+_i\|_2 \ge \lamda/4$.
\end{enumerate}
\end{lemma}
The proof of the above lemma is deferred to Section \ref{sec:tech}.

We recall Eq.\eqref{eq:reg-ecm} here for convenience. Let $\cP$ be a partition of $[n]$ into $M$ bins obtained in Lemma \ref{lem:part-ec-d} Let $[i_s, i_t]$ denote the $i^{th}$ bin in $\cP$ and let $n_i$ be its length.  Define $\bar {\bs u}_i = \frac{1}{n_i} \sum_{j=i_s}^{i_t} \bs u_j$ and $\dot {\bs u}_i = \bar {\bs u}_i - \frac{1}{n_i \beta} \sum_{j=i_s}^{i_t} \grad f_j(\bar {\bs u}_i)$ where $\beta$ is as in Assumption EC-2. Let $\bs x_j$ be the prediction made by \ALG{} at time $j$. We start with following regret decomposition.
\begin{align}
    R_n(C_n)
    &\le \sum_{i=1}^{M} \underbrace{\sum_{j=i_s}^{i_t} f_j(\bs x_j) - f_j(\dot {\bs u}_i)}_{T_{1,i}} + \sum_{i=1}^{M} \underbrace{\sum_{j=i_s}^{i_t} f_j(\dot {\bs u}_i) - f_j(\bar {\bs u}_i)}_{T_{2,i}} + \sum_{i=1}^{M} \underbrace{\sum_{j=i_s}^{i_t} f_j(\bar {\bs u}_i) - f_j(\bs u_j)}_{T_{3,i}}.
\end{align}

\begin{lemma} \label{lem:t1-ec-d} (\textbf{bounding } $T_{1,i}$) Let the experts in \ALG{} be the ONS algorithms with parameter $\zeta = \min \left \{\frac{1}{4G^\dagger(2B\sqrt{d} + 2G\sqrt{d})}, \alpha \right \}$ and decision set $\cD$. Also choose learning rate $\eta = \alpha$, for \ALG{}. Then for any bin $[i_s,i_t]$ we have,
\begin{align}
    \sum_{j=i_s}^{i_t} f_j( \bs x_j) - f_j(\dot {\bs u}_i) 
    &= O\left(d^{3/2} BG^\dagger \log n + d^{3/2}GG^\dagger \log n + \frac{\log n}{\alpha}\right)\\
    &= O(d^{3/2} \log n),
\end{align}
where $\bs x_j \in \mathbb{R}^d$ are the outputs of \ALG{}.
\end{lemma}
\begin{proof}
First we proceed to bound $\| \dot {\bs u}_i\|_\infty$.  Since $\|\grad f_j(\bs u_j) \|_2 \le G$ by Assumption EC-1, we have
\begin{align}
    \| \dot {\bs u}_i\|_\infty
    &\le \|\bar {\bs u}_i \|_\infty + \frac{G}{\beta}\\
    &\le B + \frac{G}{\beta}\\
    &\le B + G,
\end{align}
where we used $\beta > 1$ from Assumption EC-2.

For any $\bs x \in \mathcal D$, we have $\|\bs x -   \dot {\bs u}_i\|_2 \le 2B\sqrt{d} + 2G \sqrt{d}$ by triangle inequality and the fact $\|\bs y\|_2 \le \sqrt{d} \| \bs y \|_\infty$.

By Assumption EC-4 we have $\|\grad f_j(\bs x) \|_2 \le G^\dagger$ for any $\bs x \in \mathcal D$. Also, recall that by Assumption EC-3, the loss functions $f_j$ are $\alpha$ exp-concave in the domain $\cD$. Let ${ \bs p}_j, \: j \in [i_s,i_t]$ be the predictions of an ONS algorithm when run in the interval $[i_s,i_t]$. If we choose $\zeta = \min \left \{\frac{1}{4G^\dagger(2B\sqrt{d} + 2G \sqrt{d})}, \alpha \right \}$ as the parameter of the ONS, Theorem 2 of \citep{hazan2007logregret} implies that
\begin{align}
    \sum_{j=i_s}^{i_t} f_j({ \bs p}_j) - f_j(\dot {\bs u}_i)
    &= O\left(d^{3/2} BG^\dagger \log n + d^{3/2}GG^\dagger \log n + \frac{\log n}{\alpha} \right)\\
    &= O\left(d^{3/2} \log n\right).
\end{align}

Now the Lemma is implied by the SA regret bound of FLH (Theorem 3.2 of \citep{hazan2007adaptive}).
\end{proof}

For strongly convex, losses the tern $T_{1,i}$ can enjoy a better bound.
\begin{lemma} \label{lem:t1-sc-d} (\textbf{bounding } $T_{1,i}$ \textbf{ for strongly convex losses}) Suppose that the losses are $H$ strongly convex. Take experts in  \ALG{} as OGD with step size $1/(Hn)$ and decision set $\cD$. Also choose learning rate $\eta = H/(G ^\dagger)^2$, for \ALG{}. Then for any bin $[i_s,i_t]$ we have,
\begin{align}
    \sum_{j=i_s}^{i_t} f_j( \bs x_j) - f_j(\dot {\bs u}_i) 
    &= O\left(\frac{(G ^\dagger)^2 \log n}{H}\right),
\end{align}
where $\bs x_j \in \mathbb{R}^d$ are the outputs of \ALG{}.
\end{lemma}
\begin{proof}[Proof Sketch]
The lemma follows by using the regret bound of OGD with strongly convex losses from \citep{hazan2007logregret} and following similar lines of arguments as in Lemma \ref{lem:t1-ec-d}.
\end{proof}
We state the next lemma to be generically valid for any bin which is not necessarily a member of $\cP$.

\noindent\textbf{Some notations.} For a bin $[a,b]$, introduce the notations $\Delta \bs s_{a \rightarrow b} := s(\bs u_{b+1} - \bs u_b) - s(\bs u_{a} - \bs u_{a-1})$, $\bs \Gamma^+_{a \rightarrow b} = \sum_{j=a}^{b} \bs \gamma^+_j$ and $\bs \Gamma^-_{a \rightarrow b} := \sum_{j=a}^{b} \bs \gamma^-_j$. $n_{a \rightarrow b} := b - a + 1$. $\bar{\bs u}_{a \rightarrow b} = \frac{1}{n_{a \rightarrow b}}\sum_{j=a}^b \bs u_j$ and $\dot{\bs u}_{a \rightarrow b} = \bar{\bs u}_{a \rightarrow b} - \frac{1}{\beta n_{{a \rightarrow b}}} \sum_{j=a}^b \grad f_j(\bar{\bs u}_{a \rightarrow b})$.

\begin{lemma} \label{lem:t2-ec-d}
For any bin $[a,b]$, we have
\begin{align}
    T_{2,[a,b]}
    &:=\sum_{j=a}^{b} f_j(\dot {\bs u}_{a \rightarrow b}) - f_j(\bar {\bs u}_i)\\
    &\le \frac{- \|\lambda \Delta \bs s_{a \rightarrow b} + \bs \Gamma^-_{a \rightarrow b} - \bs \Gamma^+_{a \rightarrow b}\|_2^2}{2n_{a \rightarrow b} \beta} + \|\lambda \Delta \bs s_{a \rightarrow b} + \bs \Gamma^-_{a \rightarrow b} - \bs \Gamma^+_{a \rightarrow b}\|_1 C_{a \rightarrow b}.
\end{align}
\end{lemma}

\begin{proof}
Let $g(\bs x)$ be a $\alpha$-strongly smooth function. Let $\bs x^+ = \bs x - \mu \grad f(\bs x)$ for some $\mu > 0$. Then we have
\begin{align}
    g(\bs x^+) - g(\bs x)
    &\le \|\grad g(\bs x)\|_2^2 \left(\frac{\alpha}{2} \mu^2  - \mu \right)\\
    &= \frac{-\|\grad g(\bs x)\|_2^2}{2\alpha},
\end{align}
by choosing $\mu = 1/\alpha$. By taking $g(x) = \sum_{j=a}^{b} f_j(\bs x)$ and noting that $g$ is $n_i \beta$ gradient Lipschitz due to Assumption SC-2, we get
\begin{align}
    T_{2,[a,b]}
    &:=\sum_{j=a}^{b} f_j(\dot {\bs u}_{a \rightarrow b}) - f_j(\bar {\bs u}_{a \rightarrow b})\\ &\le \frac{-\left\|\sum_{j=a}^{b} \grad f_j(\bar {\bs u}_{a \rightarrow b}) \right\|_2^2}{2 n_{a \rightarrow b} \beta}\\
    &= \frac{-1}{2n_{a \rightarrow b} \beta} \left\|\sum_{j=a}^{b} \grad f_j( {\bs u}_j) + \grad f_j(\bar {\bs u}_{a \rightarrow b}) - \grad f_j( {\bs u}_j)\right\|_2^2\\
    &\le \frac{-1}{2n_{a \rightarrow b} \beta} \left\|\sum_{j=a}^{b} \grad f_j({\bs u}_j) \right\|_2^2 + \frac{1}{n_{a \rightarrow b} \beta} \left \|\sum_{j=a}^{b} \grad f_j({\bs u}_j) \right\|_1 \left\| \sum_{j=a}^{b} \grad f_j(\bar {\bs u}_{a \rightarrow b}) - \grad f_j({\bs u}_j) \right\|_2, \label{eqn:t2-last}
\end{align}
where we used $\langle \bs x, \bs y \rangle \le \|\bs x\|_2 \| \bs y \|_2 \le \|\bs x\|_1 \| \bs y \|_2$ and dropped a negative term from expanding the squared norm. From the KKT conditions in Lemma \ref{lem:kkt-ec-d} we have $\sum_{j=a}^{b} \grad f_j({\bs u}_j) = \lambda \Delta \bs s_{a \rightarrow b} + \bs \Gamma^-_{a \rightarrow b} - \bs \Gamma^+_{a \rightarrow b}$. Since $f_j$ are $\beta$-gradient Lipschitz and $\|\bar {\bs u}_{a \rightarrow b} - {\bs u}_j \|_2 \le \|\bar {\bs u}_{a \rightarrow b} - {\bs u}_j \|_1 \le C_{a \rightarrow b}$, we also have 
\begin{align}
  \left\| \sum_{j=a}^{b} \grad f_j(\bar {\bs u}_{a \rightarrow b}) - \grad f_j({\bs u}_j) \right\|_2 &\le  n_{a \rightarrow b} \beta C_{a \rightarrow b}.
\end{align}
Substituting these we get the statement of the lemma.
\end{proof}

\begin{lemma} \label{lem:t3-ec-d} For any bin $[i_s,i_t] \in \cP$, we have
\begin{align}
    \sum_{j=i_s}^{i_t} f_j(\bar {\bs u}_i) - f_j({\bs u}_j)
    &\le  \frac{\beta n_i C_i^2}{2} + 5\lamda C_i + \left \| \lamda \Delta \bs s_i + \bs \Gamma^-_i - \bs \Gamma^+_i \right \|_2 C_i.
\end{align}
\end{lemma}
\begin{proof}
Due to strong smoothness, we have
\begin{align}
    T_{3,i}
    &:=\sum_{j=i_s}^{i_t} f_j(\bar {\bs u}_i) - f_j({\bs u}_j)\\
    &\le_{(a)} \sum_{j=i_s}^{i_t} \langle \grad f_j({\bs u}_j), \bar {\bs u}_i - {\bs u}_j \rangle + \frac{\beta}{2} \|\bar {\bs u}_i - {\bs u}_j\|_1^2\\
    &\le \frac{\beta n_i C_i^2}{2} + \sum_{j=i_s}^{i_t} \langle \grad f_j({\bs u}_j), \bar {\bs u}_i - {\bs u}_j \rangle, \label{eq:t3-ec}
\end{align}
where in line (a) we used $\|\bar {\bs u}_i - {\bs u}_j\|_2 \le \|\bar {\bs u}_i - {\bs u}_j\|_1$.

Further,
\begin{align}
    \sum_{j=i_s}^{i_t} \langle \grad f_j({\bs u}_j), \bar {\bs u}_i - {\bs u}_j \rangle
    &= \lambda \left( \langle \bs s_{i_s-1}, {\bs u}_{i_s} - \bar {\bs u}_i \rangle - \langle \bs s_{i_t} , {\bs u}_{i_t} - \bar {\bs u}_i \rangle \right) \\ &\qquad +\lambda \sum_{j=i_s+1}^{i_t} \|{\bs u}_j - {\bs u}_{j-1}\|_1 + \sum_{j=i_s}^{i_t} \langle \bs \gamma^-_j - \bs \gamma^+_j, \bar{\bs u}_i - \bs u_j \rangle
\end{align}

By triangle and Holder's inequalities, the first two terms can be bounded by $3 \lamda C_i$ (recall that $\|\bs u_t - \bar {\bs u}_i\|_1 \le C_i$ for all $t \in [i_s,i_t]$ ). Let's proceed to bound the last term in the above display. From Lemma \ref{lem:part-ec-d}, we have $C_i \le B/\sqrt{n_i}$. So the TV incurred across each coordinate of the optimal solution is at-most $B$. Using similar arguments as in Lemma \ref{lem:ppt-1d}, the complementary slackness in Lemma \ref{lem:kkt-ec-d} implies that for each $k \in [d]$, if $\bs \gamma^-_j[k] > 0$ for at-least one $j \in [i_s,i_t]$ then $\bs \gamma^+_j[k] = 0$ for all $j \in [i_s,i_t]$. Similarly for each $k \in [d]$, if $\bs \gamma^+_j[k] > 0$ for at-least one $j \in [i_s,i_t]$ then $\bs \gamma^-_j[k] = 0$ for all $j \in [i_s,i_t]$. This observation allows us to write,

\begin{align}
    \sum_{j=i_s}^{i_t} |\bs \gamma^-_j[k] - \bs \gamma^+_j[k] |
    &= \left|\bs \Gamma^-_i[k] - \bs \Gamma^+_i[k] \right| \label{eqn:obs}
\end{align}

Define $C_i^k := \sum_{j=i_s+1}^{i_t} | \bs u_j[k] - \bs u_{j-1}[k] |$. We have,
\begin{align*}
    &\quad \sum_{j=i_s}^{i_t} \langle \bs \gamma^-_j - \bs \gamma^+_j, \bar{\bs u}_i - \bs u_j \rangle
    = \sum_{k=1}^{d} \sum_{j=i_s}^{i_t} (\bs \gamma^-_j[k] - \bs \gamma^+_j[k])(\bar{\bs u}_i[k] - \bs u_j[k])\\
    &\le_{(a)} \sum_{k=1}^{d} \left|\bs \Gamma^-_i[k] - \bs \Gamma^+_i[k] \right| C_i^k\\
    &= \sum_{k=1}^{d} \left( \lamda \Delta \bs s_i[k] \: \sign {\left(\bs \Gamma^-_i[k] - \bs \Gamma^+_i[k] \right)} + \sign {\left(\bs \Gamma^-_i[k] - \bs \Gamma^+_i[k] \right)}\left(\bs \Gamma^-_i[k] - \bs \Gamma^+_i[k] \right) \right) C_i^k \\
    &\qquad - \sum_{k=1}^{d} \lamda \Delta \bs s_i[k] \: \sign {\left(\bs \Gamma^-_i[k] - \bs \Gamma^+_i[k] \right)} C_i^k\\
    &\le_{(b)} \left \| \lamda \Delta \bs s_i + \bs \Gamma^-_i - \bs \Gamma^+_i \right \|_2 C_i + 2\lamda C_i,
\end{align*}
where in line (a) we applied $\langle \bs x, \bs y \rangle\le \| \bs x \|_1 \| \bs y\|_\infty$ along with the Eq. \eqref{eqn:obs}. In line (b) we applied $\langle \bs x, \bs y \rangle \le \| \bs x \|_2 \| \bs y\|_1$ for the first term and $\langle \bs x, \bs y \rangle \le \| \bs x \|_\infty \| \bs y\|_1$ for the second term. Putting everything together yields the Lemma.

\end{proof}

\begin{figure}[p]
	\centering
	\fbox{
		\begin{minipage}{14 cm}
		    \code{splitMonotonic}: Inputs - (1) an interval $[i_s,i_t]$ such that the offline optimal is monotonic across each coordinate $k \in [d]$; (2) offline optimal sequences $\bs u_{1:n}$ and the sequence of subgradients (dual variables) $\bs s_{1:n-1}$ (recall that $\bs s_0 = \bs s_n = 0$ by convention.).
			\begin{enumerate}
				\setlength\itemsep{0em}
                \item Initialize $\cT \leftarrow \Phi$, $\mathcal S \leftarrow \Phi$.
                \item Add $i_s,i_t$ to $\mathcal T$.
                \item For each coordinate $k \in [d]$:
                \begin{enumerate}
                    \item If $\bs u[k]$ is constant in $[i_s,i_t]$, then skip the current coordinate.
                    \item Initialize $z_1 \leftarrow i_s, z_2 \leftarrow i_t$.
                    \item If $\bs u_{i_s}[k] = \pm B$, let $z_1$ be the first time point in $[i_s,i_t]$ where $\bs u_{z_1}[k] \neq \pm B$. Add $z_1-1,z_1$ to $\mathcal T$.
                    \item If $\bs u_{i_t}[k] = \pm B$, let $z_2$ be the last time point in $[i_s,i_t]$ where $\bs u_{z_2}[k] \neq \pm B$. Add $z_2,z_2+1$ to $\mathcal T$.
                    \item If $\bs u[k]$ is non-decreasing in $[i_s,i_t]$ then let $p \ge z_1$ be the first point with $\bs s_{p-1}[k] = 1$. If $p > z_1$, add $p-1,p$ to $\cT$.
                    \item If $\bs u[k]$ is non-decreasing in $[i_s,i_t]$ then let $q \le z_2$ be the last point with $\bs s_q[k] = 1$. If $q < z_2$, add $q,q+1$ to $\cT$.
                    \item If $\bs u[k]$ is non-increasing in $[i_s,i_t]$ then let $p \ge z_1$ be the first point with $\bs s_{p-1}[k] = -1$. If $p > z_1$, add $p-1,p$ to $\cT$.
                    \item If $\bs u[k]$ is non-increasing in $[i_s,i_t]$ then let $q \le z_2$ be the last point with $\bs s_q[k] = -1$. If $q < z_2$, add $q,q+1$ to $\cT$.            
                \end{enumerate}
                \item For each entry $t$ in $\cT$:
                \begin{enumerate}
                    \item If $t$ appears more than 2 times, delete some occurences of $t$ such that $t$ only appears 2 times in $\cT$. 
                \end{enumerate}
                \item Sort $\cT$ in non-decreasing order. For each consecutive points $s,t \in \mathcal T$, add $[s,t]$ to $\mathcal S$.                
                \item Return the partition $\mathcal S$.
            	\end{enumerate}
		\end{minipage}
	}
	\caption{\code{splitMonotonic} procedure}
	\label{fig:split}
\end{figure}
\begin{figure}
    \centering
    \includegraphics[width =0.7\textwidth]{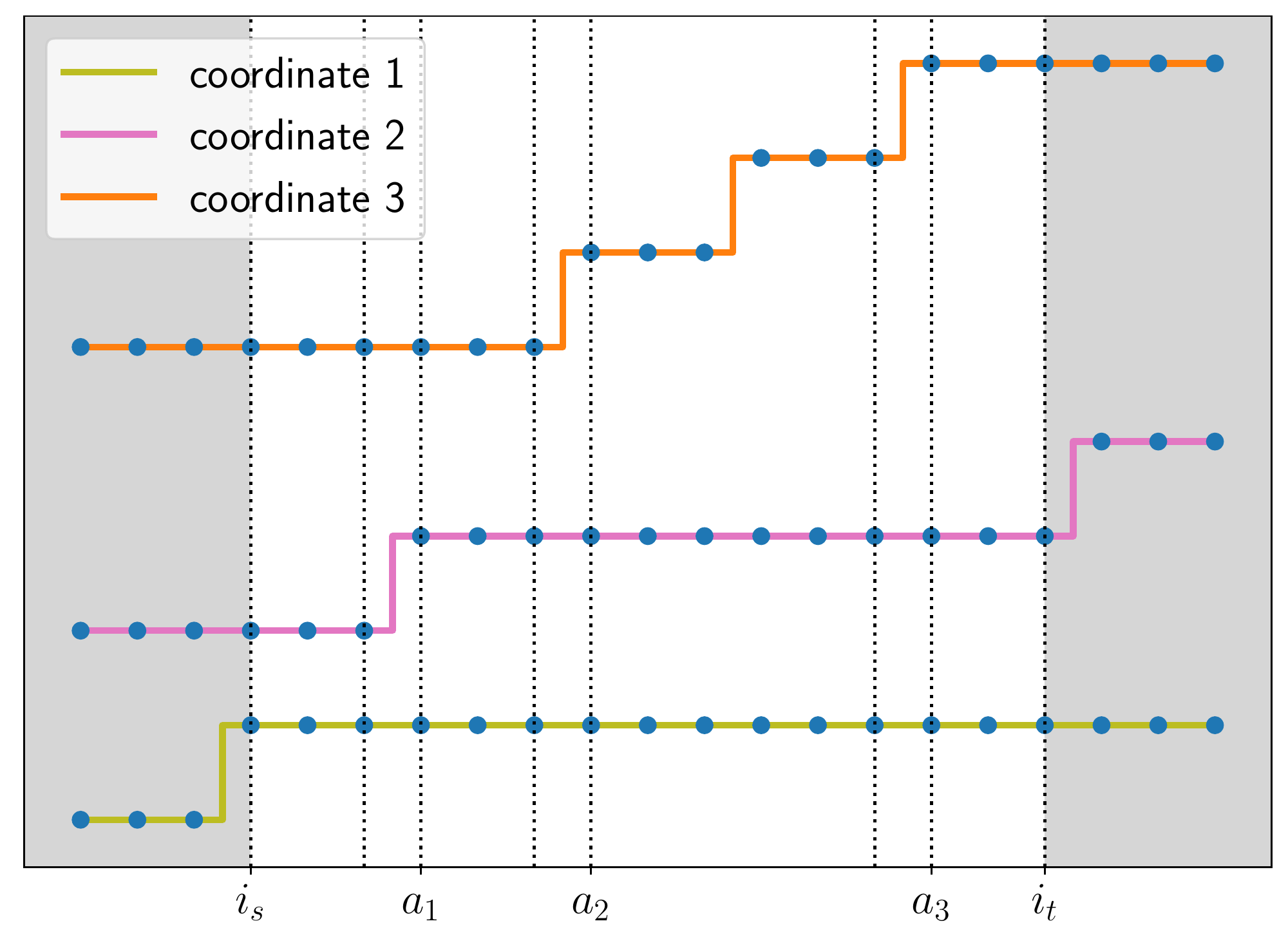}
    \caption{\emph{An example of the partitioning created by \code{splitMonotonic} (See Fig. \ref{fig:split}). The partition $\cS$ returned by \code{splitMonotonic}} is $\{[i_s,a_1-1], [a_1,a_2-1], [a_2,a_3-1], [a_3, i_t] \}$. Blue dots indicate the offline optimal sequence.}
    \label{fig:ec-mono}
\end{figure}

\begin{lemma} \label{lem:split}
Let \code{splitMonotonic} in Fig.\ref{fig:split} be run with an input $[i_s,i_t]$. Then the partition $\mathcal S$ it return obeys $|\mathcal S| = O(d)$.
\end{lemma}
\begin{proof}
From the psuedo-code in Fig. \ref{fig:split} it is obvious that each coordinate can contribute to increasing the bin count by $O(1)$. Hence the overall bin count in $\mathcal S$ is $O(d)$.
\end{proof}
An illustrative example of the input and output of \code{splitMonotonic}  is given in Fig.~\ref{fig:ec-mono}.

\main*
\begin{proof}
Consider a bin $[i_s,i_t] \in \cP$. By Lemma \ref{lem:part-ec-d}, the bin has to satisfy one of the two Properties. Let's first focus on the scenario where $[i_s,i_t]$ satisfies Property 2.

Combining the results of Lemmas \ref{lem:t2-ec-d}, \ref{lem:t3-ec-d} we can write,

\begin{align}
    T_{2,i} + T_{3,i}
    &\le \frac{- \|\lambda \Delta \bs s_i + \bs \Gamma^-_i - \bs \Gamma^+_i\|_2^2}{2n_i \beta} + \|\lambda \Delta \bs s_i + \bs \Gamma^-_i - \bs \Gamma^+_i\|_2 C_i \\
    &\quad +  \frac{\beta n_i C_i^2}{2} + 5\lamda C_i + \left \| \lamda \Delta \bs s_i + \bs \Gamma^-_i - \bs \Gamma^+_i \right \|_2 C_i\\
    &\le_{(a)} \frac{\beta B^2}{2} - \frac{\|\lambda \Delta \bs s_i + \bs \Gamma^-_i - \bs \Gamma^+_i\|_2^2}{2n_i \beta} + 7 \left( \left \| \lamda \Delta \bs s_i + \bs \Gamma^-_i - \bs \Gamma^+_i \right \|_2 \vee \lamda \right)C_i\\
    &= \frac{\beta B^2}{2} - \left(\frac{\|\lambda \Delta \bs s_i + \bs \Gamma^-_i - \bs \Gamma^+_i\|_2}{\sqrt{2n_i \beta}} - \frac{7C_i\sqrt{n_i \beta} \left( \left \| \lamda \Delta \bs s_i + \bs \Gamma^-_i - \bs \Gamma^+_i \right \|_2 \vee \lamda \right)}{\|\lambda \Delta \bs s_i + \bs \Gamma^-_i - \bs \Gamma^+_i\|_2 \sqrt{2}} \right)^2\\
    &\quad + \frac{49 n_i \beta C_i^2}{2} \left( \frac{ \left \| \lamda \Delta \bs s_i + \bs \Gamma^-_i - \bs \Gamma^+_i \right \|_2 \vee \lamda}{\|\lambda \Delta \bs s_i + \bs \Gamma^-_i - \bs \Gamma^+_i\|_2} \right)^2\\
    &\le_{(b)} \frac{\beta B^2}{2} + 392 \beta n_i C_i^2\\
    &\le 393 \beta B^2,
\end{align}
where in line (a) we used $C_i \le B/\sqrt{n_i}$ for partitions in $\cP$ (Lemma \ref{lem:part-ec-d}). In line (b) we used $\frac{ \| \lamda \Delta \bs s_i + \bs \Gamma^-_i - \bs \Gamma^+_i \|_2 \vee \lamda}{\|\lambda \Delta \bs s_i + \bs \Gamma^-_i - \bs \Gamma^+_i\|_2} \ge 4$ since $\|\lambda \Delta \bs s_i + \bs \Gamma^-_i - \bs \Gamma^+_i\|_2 \ge \lamda/4$ by Property 2 of Lemma \ref{lem:part-ec-d}.

Now using Lemma \ref{lem:t1-ec-d}, for the bins $[i_s,i_t]$ that satisfy property 2, we can write

\begin{align}
    T_{1,i} + T_{2,i} + T_{3,i}
    &= \tilde O(d^{1.5}). \label{eq:prop2}
\end{align}

Now suppose that the bin $[\ubar t ,\bar t]$ satisfies Property 1 in Lemma \ref{lem:part-ec-d}. In this case, via a call to \code{splitMonotonic} function with the input interval as $[\ubar t ,\bar t]$, we split the original bin into $O(d)$ sub-bins (see Lemma \ref{lem:split}). Further for a fixed $k$, if $\bs u_j[k],\: j \in [\ubar t ,\bar t]$ is non-decreasing, then we can group those consecutive sub-bins into at-most three categories: (a) a section of time where $\bs u_j[k]$ is constant; (b) a section of time where $\bs u_j[k]$ is non-decreasing; (c) a section of time where $\bs u_j[k]$ is constant.

We proceed to define these sections formally (where $p, m, q$ are indices defined for convenience)
\begin{itemize}
    \item For section (a) let $\mathcal A = \{ [\ubar t,\ubar t_{-p} -1], [\ubar t_{-p}, \bar t_{-p}], \ldots [\ubar t_0, \bar t_0] \}$
    \item For section (b) let $\mathcal B = \{[\ubar t_1,\bar t_1],\ldots, [\ubar t_m,\bar t_m] \}$
    \item For section (c) let $\mathcal C = \{ [\ubar t_{m+1},\bar t_{m+1}],\ldots,[\bar t_q + 1,\bar t] \}$
\end{itemize}

As mentioned before, these sections are constructed so that the oflline optimal satisfy the following properties.

\begin{enumerate}
    \item[(i)] $\bs u_j[k] \: j \in [\ubar t,\bar t_0]$ is constant.
    \item[(ii)] $\bs u_{\ubar t_1}[k] > \bs u_{\ubar t_1 -1 }[k]$ and $\bs u_{\ubar t_{m+1}}[k] > \bs u_{\ubar t_{m+1} -1}[k]$.
    \item[(iii)] $\bs u_j[k] \: j \in [\ubar t_1, \bar t_m]$ is non-decreasing.
    \item[(iv)] $\bs u_j[k] \: j \in [\ubar t_{m+1},\bar t] $ is constant.
\end{enumerate}
We remark that the grouping may be different for different coordinates $k$. Further some of $\mathcal A,\mathcal B$ or $\mathcal C$ can be empty. In the example we gave in Fig. \ref{fig:ec-mono}:
\begin{itemize}
    \item For coordinate 1 $\mathcal A = \phi$, $\mathcal B = \phi$, $\mathcal C = \{[i_s,a_1-1], [a_1,a_2-1], [a_2,a_3-1], [a_3, i_t] \}$.
    \item For coordinate 2 $\mathcal A = [i_s,a_1-1]$, $\mathcal B = \phi$, $\mathcal C = \{[a_1,a_2-1], [a_2,a_3-1], [a_3, i_t] \}$.
    \item For coordinate 3 $\mathcal A = \{ [i_s,a_1-1], [a_1,a_2-1]\}$, $\mathcal B = \{ [a_2,a_3-1]\}$, $\mathcal C = \{ [a_3, i_t]\}$
\end{itemize}

We fill focus on the aforementioned scenario where $\bs u_j[k],\: j \in [\ubar t ,\bar t]$ is non-decreasing. The arguments for the case where  $\bs u_j[k],\: j \in [\ubar t ,\bar t]$ is non-increasing are similar. Further similar to the proof of Theorem \ref{thm:ec-1d}, we give arguments for the case where $\gamma^+_j[k] = 0$ for all $j$ in the interval $[\ubar t ,\bar t]$ stating that arguments for the case $\gamma^-_j[k] = 0$ uniformly in  $[\ubar t ,\bar t]$ are similar.

From Lemma \ref{lem:t2-ec-d}, we have
\begin{align}
    \sum_{j=a}^{b} f_j(\dot {\bs u}_{a \rightarrow b}) - f_j(\bar {\bs u}_{a \rightarrow b})
    &\le \frac{- \|\lambda \Delta \bs s_{a \rightarrow b} + \bs \Gamma^-_{a \rightarrow b} - \bs \Gamma^+_{a \rightarrow b}\|_2^2}{2n_{a \rightarrow b} \beta} + \|\lambda \Delta \bs s_{a \rightarrow b} + \bs \Gamma^-_{a \rightarrow b} - \bs \Gamma^+_{a \rightarrow b}\|_1 C_{a \rightarrow b}.\label{eq:ec1}
\end{align}

Observe that the relation in Eq. \eqref{eq:t3-ec} holds for any generic bin $[a,b]$ that may not be a member of $\cP$  (replacing $C_i,n_i,\bar{\bs u}_i$ with $C_{a\rightarrow b},n_{a \rightarrow b},\bar {\bs u}_{a \rightarrow b}$). So
\begin{align}
    T_{3,[a,b]}
    &:=\sum_{j=a}^{b} f_j(\bar {\bs u}_{a \rightarrow b}) - f_j({\bs u}_j)
    \le \sum_{k=1}^d \frac{\beta n_{a \rightarrow b} C_{a \rightarrow b}^2}{2d} + \sum_{j=a}^{b} \langle \grad f_j({\bs u}_j), \bar {\bs u}_{a \rightarrow b} - {\bs u}_j \rangle.\label{eq:ec2}
\end{align}
Note that Eq. \eqref{eq:ec1} and \eqref{eq:ec2} decompose coordinate-wise. So for the bin $[\ubar t, \bar t] \in \cP$ where the optimal sequence is monotonic across each coordinate, our strategy is to bound

\begin{align}
    \bs S_{a \rightarrow b}[k]
    &:= \frac{- \left(\lambda \Delta \bs s_{a \rightarrow b}[k] + \bs \Gamma^-_{a \rightarrow b}[k] - \bs \Gamma^+_{a \rightarrow b}[k]\right)^2}{2n_{a \rightarrow b} \beta} + |\lambda \Delta \bs s_{a \rightarrow b}[k] + \bs \Gamma^-_{a \rightarrow b}[k] - \bs \Gamma^+_{a \rightarrow b}[k]| C_{a \rightarrow b} \nonumber\\
    &\quad + \frac{\beta n_{a \rightarrow b} C_{a \rightarrow b}^2}{2d} + \sum_{j=a}^{b}  \grad f_j({\bs u}_j)[k]( \bar {\bs u}_{a \rightarrow b}[k] - {\bs u}_j[k]), \label{eq:coord}
\end{align}
for each $k \in [d]$ and $[a,b] \in \mathcal A \cup \mathcal B \cup \mathcal C$ and finally adding them across all coordinates to bound $\sum_{k=1}^d \bs S_{a \rightarrow b}[k]$. Doing so will result in a bound on $T_{2,[a,b]} + T_{3,[a,b]}$. Further, $T_{1,[a,b]}$ can be bound by strongly adaptive regret. This enables us to bound $\sum_{[a,b] \in \mathcal A \cup \mathcal B \cup \mathcal C} T_{1,[a,b]} + T_{2,[a,b]} + T_{3,[a,b]}$ thereby leading to a regret bound in the parent bin $[\ubar t, \bar t] \in \cP$ which was the input interval for the call to \code{splitMonotonic} that we started with.

Let $C_{[a,b]}[k]$ be the TV of offline optimal incurred in the interval any interval $[a, b]$ along coordinate $k$. First we focus on the  bins in $\mathcal B$. If $\mathcal B$ is not empty, then $\bs \gamma_j[k] = 0 \: \forall j \in [\ubar t_1 \rightarrow \bar t_m]$ due to property (ii) and (iii) above. By using the stationarity conditions in Lemma \ref{lem:kkt-ec-d}, we can write
\begin{align}
    \sum_{[a,b] \in \mathcal B} \sum_{j=a}^b \grad f_j({\bs u}_j)[k]( \bar {\bs u}_{a \rightarrow b}[k] - {\bs u}_j[k])
    &=  \lamda C_{\ubar t_1 \rightarrow \bar t_m}[k] + \lamda \left(s_{\ubar t_1 - 1}[k] \bs u_{\ubar t_1}[k]  - s_{\bar t_m}[k] \bs u_{\bar t_m}[k]\right)\\
    &\quad + \sum_{[a,b] \in \mathcal B} \lamda \bar {\bs u}_{a \rightarrow b}[k]  \Delta \bs s_{a \rightarrow b}[k].
\end{align}
So we have,
\begin{align}
    \sum_{[a,b] \in \mathcal B} \bs S_{a \rightarrow b}[k]
    &\le \frac{\beta n_{\ubar t_1 \rightarrow \bar t_m} C_{\ubar t_1 \rightarrow \bar t_m}^2}{2d} + \lamda C_{\ubar t_1 \rightarrow \bar t_m}[k] + \lamda \left(s_{\ubar t_1 - 1}[k] \bs u_{\ubar t_1}[k]  - s_{\bar t_m}[k] \bs u_{\bar t_m}[k]\right) \\
    &\qquad +\sum_{[a,b] \in \mathcal B} \frac{-\lambda^2  (\Delta \bs s_{a \rightarrow b}[k])^2}{2n_{a \rightarrow b} \beta} + \lamda |\Delta \bs s_{a \rightarrow b}[k]| C_{\ubar t_1 \rightarrow \bar t_m} + \lamda \bar {\bs u}_{a \rightarrow b}[k]  \Delta \bs s_{a \rightarrow b}[k]\\
    &\le_{(a)} \frac{\beta n_{\ubar t \rightarrow \bar t} C_{\ubar t \rightarrow \bar t}^2}{2d} + \lamda C_{\ubar t_1 \rightarrow \bar t_m}[k]  + \lamda \left(s_{\ubar t_1 - 1}[k] \bs u_{\ubar t_1}[k]  - s_{\bar t_m}[k] \bs u_{\bar t_m}[k]\right) \\
    &\qquad +\sum_{[a,b] \in \mathcal B} \frac{-\lambda^2  (\Delta \bs s_{a \rightarrow b}[k])^2}{2n_{a \rightarrow b} \beta} + \lamda |\Delta \bs s_{a \rightarrow b}[k]| C_{\ubar t \rightarrow \bar t} + \lamda \bar {\bs u}_{a \rightarrow b}[k]  \Delta \bs s_{a \rightarrow b}[k]\\
    &\le_{(b)} \frac{\beta B^2}{2d} +\sum_{[a,b] \in \mathcal B} \frac{-\lambda^2  (\Delta \bs s_{a \rightarrow b}[k])^2}{2n_{a \rightarrow b} \beta} + \lamda |\Delta \bs s_{a \rightarrow b}[k]| C_{\ubar t \rightarrow \bar t} + \lamda \bar {\bs u}_{a \rightarrow b}[k]  \Delta \bs s_{a \rightarrow b}[k],    
\end{align}
where in line (a) we used the fact that $C_{\ubar t_1 \rightarrow \bar t_m} \le C_{\ubar t \rightarrow \bar t}$ and $n_{\ubar t_1 \rightarrow \bar t_m} \le n_{\ubar t \rightarrow \bar t}$ since $[\ubar t_1,\bar t_m]$ is contained within $[\ubar t, \bar t]$. In line (b) we used $C_{\ubar t \rightarrow \bar t} \le B/\sqrt{n_{\ubar t \rightarrow \bar t}}$ (since $[\ubar t, \bar t] \in \cP$) along with the fact that \\
$\lamda \left(\bs s_{\ubar t_1 - 1}[k] \bs u_{\ubar t_1}[k]  - \bs s_{\bar t_m}[k] \bs u_{\bar t_m}[k]\right) = -\lamda C_{\ubar t_1 \rightarrow \bar t_m}$ since \\
$\bs s_{\ubar t_1 - 1}[k] = \bs s_{\bar t_m} = 1$ due to property (ii) and (iii) above.

Define $\check{ \bs u}_{\mathcal B} := \frac{1}{|\mathcal B|} \sum_{[a,b] \in \mathcal B} \bar {\bs u}_{a \rightarrow b}$. 
Observe that since $\bs s_{\ubar t_1 - 1}[k] = \bs s_{\bar t_m}[k] = 1$, we can write
$\sum_{[a,b] \in \mathcal B} \Delta \bs s_{a\rightarrow b}[k] =  0$ by the telescoping structure.

By noting that we can subtract $0 = \lambda\check{ \bs u}_{\mathcal B}[k] \sum_{(a,b)\in \cB}\Delta \bs s_{a \rightarrow b}[k]$  and that 
$|\bar {\bs u}_{a \rightarrow b}[k] - \check{ \bs u}_{\mathcal B}[k]| \le C_{\ubar t \rightarrow \bar t}$, we have

\begin{align*}
     \sum_{[a,b] \in \mathcal B} \bs S_{a \rightarrow b}[k]
     &\le \frac{\beta B^2}{2d} +\sum_{[a,b] \in \mathcal B} \frac{-\lambda^2  (\Delta \bs s_{a \rightarrow b}[k])^2}{2n_{a \rightarrow b} \beta} + \lamda |\Delta \bs s_{a \rightarrow b}[k]| C_{\ubar t \rightarrow \bar t} + \lamda \left(\bar {\bs u}_{a \rightarrow b}[k] - \check{ \bs u}_{\mathcal B}[k] \right)\Delta \bs s_{a \rightarrow b}[k]\\
     &\le \frac{\beta B^2}{2d} +\sum_{[a,b] \in \mathcal B} \frac{-\lambda^2  (\Delta \bs s_{a \rightarrow b}[k])^2}{2n_{a \rightarrow b} \beta} + 2 \lamda |\Delta \bs s_{a \rightarrow b}[k]| C_{\ubar t \rightarrow \bar t}\\
     &= \frac{\beta B^2}{2d} +\sum_{[a,b] \in \mathcal B} -\left(\frac{\lamda \Delta \bs s_{a \rightarrow b}[k]}{\sqrt{2 n_{a \rightarrow b} \beta}}  - C_{\ubar t \rightarrow \bar t} \sqrt{2 n_{a \rightarrow b} \beta}\right)^2 + 2\beta n_{a \rightarrow b} C_{\ubar t \rightarrow \bar t}^2\\
     &\le \frac{\beta B^2}{2d} + 2\beta n_{\ubar t \rightarrow \bar t} C_{\ubar t \rightarrow \bar t}^2\\
     &\le \frac{\beta B^2}{2d} + 2\beta B^2\\
     &\le 3\beta B^2.
\end{align*}

Next, we address bins present in $\mathcal A$ and $\mathcal C$. We provide the arguments for bounding $\sum_{[a,b] \in \mathcal A} \bs S_{a \rightarrow b}[k]$. Bounding the sum for bins in $\mathcal C$ can be done using similar arguments.

Observe that by property (i) above, the sequence $\bs u_j[k]$ for $j \in [i_s,\bar t_0]$ is a constant. So the last  term in Eq. \eqref{eq:coord} is zero for any $\bs S_{a \rightarrow b}[k]$ where $[a,b] \in \mathcal A$. Now proceeding similar to above by completing the squares and dropping the negative terms, we get

\begin{align}
    \sum_{[a,b] \in \mathcal A} \bs S_{a \rightarrow b}[k]
    &\le \sum_{[a,b] \in \mathcal A} \left( \frac{- \left(\lambda \Delta \bs s_{a \rightarrow b}[k] + \bs \Gamma^-_{a \rightarrow b}[k] - \bs \Gamma^+_{a \rightarrow b}[k]\right)^2}{2n_{a \rightarrow b} \beta} \right. \\
    &\qquad \left. + |\lambda \Delta \bs s_{a \rightarrow b}[k] + \bs \Gamma^-_{a \rightarrow b}[k] - \bs \Gamma^+_{a \rightarrow b}[k]| C_{a \rightarrow b}  + \frac{\beta n_{a \rightarrow b} C_{a \rightarrow b}^2}{2d} \vphantom{\frac{- \left(\lambda \Delta \bs s_{a \rightarrow b}[k] + \bs \Gamma^-_{a \rightarrow b}[k] - \bs \Gamma^+_{a \rightarrow b}[k]\right)^2}{2n_{a \rightarrow b} \beta}}\right)\\
    &= \sum_{[a,b] \in \mathcal A} \left( -\left( \frac{\lambda \Delta \bs s_{a \rightarrow b}[k] + \bs \Gamma^-_{a \rightarrow b}[k] - \bs \Gamma^+_{a \rightarrow b}[k] }{\sqrt{2n_{a \rightarrow b} \beta}}  - C_{a \rightarrow b} \sqrt{\frac{n_{a \rightarrow b} \beta}{2}}\right)^2 \right. \\
    &\qquad \left. + \frac{n_{a \rightarrow b} \beta  C_{a \rightarrow b}^2}{2} + \frac{\beta n_{a \rightarrow b} C_{a \rightarrow b}^2}{2d} \vphantom{\left( \frac{-\lambda \Delta \bs s_{a \rightarrow b}[k] + \bs \Gamma^-_{a \rightarrow b}[k] }{\sqrt{2n_{a \rightarrow b} \beta}}  - C_{a \rightarrow b} \sqrt{\frac{n_{a \rightarrow b} \beta}{2}}\right)^2} \right)\\
    &\le  \sum_{[a,b] \in \mathcal A} n_{a \rightarrow b} \beta  C_{\ubar t \rightarrow \bar t}^2\\
    &\le n_{\ubar t \rightarrow \bar t} \beta  C_{\ubar t \rightarrow \bar t}^2\\
    &\le \beta B^2.
\end{align}
Similarly it can be shown that $\sum_{[a,b] \in \mathcal C} \bs S_{a \rightarrow b}[k] = O(1)$. Recalling that $|\mathcal A| + |\mathcal B|+ |\mathcal C| = O(d)$ we have 
\begin{align}
    T_{2,[\ubar t, \bar t]} + T_{3,[\ubar t, \bar t]}
    &\le \sum_{k=1}^{d}\sum_{[a,b] \in \mathcal A \cup \mathcal B \cup \mathcal C} \bs S_{a \rightarrow b}[k]
    = O(d).
\end{align}

From Lemma \ref{lem:t1-ec-d} we have
\begin{align}
    T_{1,[\ubar t, \bar t]}
    &= \tilde O(d^{2.5}), \label{eq:prop1}
\end{align}
for bins $[\ubar t, \bar t] \in \cP$ that satisfy property 2 in Lemma \ref{lem:part-ec-d}.

Comparing Eq. \eqref{eq:prop2} and \eqref{eq:prop1} we conclude that 
\begin{align}
    T_{1,i}+T_{2,i}+T_{3,i}
    &= \tilde O(d^{2.5}), \label{eq:prop3}
\end{align}
for all bins $[i_s,i_t]$ in the partition $\cP$ of Lemma \ref{lem:part-ec-d}. Since $|\cP| = O(d n^{1/3}C_n^{2/3})$, adding the above bound across all bins leads to the theorem.

If $C_n \le 1/n$, then we have
\begin{align}
    \sum_{t=1}^{n} f_t(\bs x_j) - f_t(\bs u_t)
    &\le \sum_{t=1}^{n} f_t(\bs x_j) - f_t(\bs u_1) + \sum_{t=1}^{n} f_t(\bs u_1) - f_t(\bs u_t)\\
    &\le_{(a)} \tilde O(d^{1.5}) + G^\dagger nC_n\\
    &= \tilde O(d^{1.5})
\end{align}
where line (a) follows from the fact that $f_t$ is $G^\dagger$ Lipschitz in $\cD$.

\end{proof}

\propsc*
\begin{proof}[Proof Sketch]
First we consider the case where the offline optimal in monotonic in each coordinate of a bin in $\cP$.
The static regret in any bin for strongly convex losses is $O(\log n)$ by Lemma \ref{lem:t1-sc-d} (as opposed to $\tilde O(d^{1.5})$ for exp-concave losses). Hence Eq.\eqref{eq:prop1} can be re-written as $T_{1,[\ubar t, \bar t]} = \tilde O(d)$. By following similar arguments as in proof of Theorem \ref{thm:ec-d}, we can re-write Eq.\eqref{eq:prop3} as
\begin{align}
    T_{1,i}+T_{2,i}+T_{3,i}
    &= \tilde O(d).    
\end{align}

If the offline optimal is not monotonic, in each coordinate, we can write
\begin{align}
    T_{1,i} + T_{2,i} + T_{3,i}
    &= \tilde O(1),
\end{align}
by following similar arguments for the corresponding case in the proof of Theorem \ref{thm:ec-d}.

Finally we sum across all $|\cP| = O(d n^{1/3}C_n^{2/3})$. The case $C_n \le 1/n$ can be handled similar to that of the exp-concave case.
\end{proof}

\section{Technical Lemmas} \label{sec:tech}

We start by describing a partitioning procedure  namely \code{generateBins}.

    \begin{mdframed}
		    \code{generateBins}: Inputs - the offline optimal sequence.
			\begin{enumerate}
				\setlength\itemsep{0em}
                \item[Step 1] Initialize $\mathcal Q \leftarrow \Phi$. Starting from time 1, spawn a new bin $[i_s,i_t]$ whenever $\sum_{j=i_s+1}^{i_t+1} \|\bs u_j - \bs u_{j-1} \|_1 > B/\sqrt{n_i}$, where $n_i = i_t - i_s + 1$. Add the spawned bin $[i_s,i_t]$ to $\mathcal Q$.
                \item[Step 2] Initialize $\cP \leftarrow \Phi, \mathcal R \leftarrow \Phi$.
                \item[Step 3] For each bin $[i_s,i_t] \in \mathcal Q$:
                \begin{enumerate}
                    \item Let $\Delta \bs s_i = \bs s_{i_t} - \bs s_{i_s-1}$. $\bs \Gamma_i^+ = \sum_{j=i_s}^{i_t} \bs \gamma_j^+$. $\bs \Gamma_i^- = \sum_{j=i_s}^{i_t} \bs \gamma_j^-$.
                    
                    \item If for each $k \in [d]$, the sequence $\bs u_{k}$ is monotonic in $[i_s,i_t]$, then remove $[i_s,i_t]$ from $\mathcal Q$ and add it to $\cP$.

                    \item If there exists one coordinate $k \in [d]$ such that $\bs s_{i_s-1}[k] \in [-1,-1/4]$ and $\bs s_{i_t}[k] \in [0,1]$ and $\bs \gamma^+_j[k] = 0 \: \forall j \in [i_s,i_t]$, then remove $[i_s,i_t]$ from $\mathcal Q$ and add it to $\cP$. Goto Step 3.
                    
                    \item If there exists one coordinate $k \in [d]$ such that $\bs s_{i_s-1}[k] \in [-1/4,0]$ and $\bs s_{i_t}[k] \in [1/4,1]$ and $\bs \gamma^+_j[k] = 0 \: \forall j \in [i_s,i_t]$, then remove $[i_s,i_t]$ from $\mathcal Q$ and add it to $\cP$. Goto Step 3.
                    
                    \item If there exists one coordinate $k \in [d]$ such that $\bs u_{k}$ is non-monotonic in $[i_s,i_t]$ and $\bs s_{i_s-1}[k] \in [-1/4,1]$ and $\bs s_{i_t}[k] \in [-1/4,1]$ and $\bs \gamma^+_j[k] = 0 \: \forall j \in [i_s,i_t]$ then:
                    \begin{enumerate}
                        \item Initialize $z \leftarrow i_s$. Remove $[i_s,i_t]$ from $\mathcal Q$.
                        
                        \item if $\bs u_{i_s}[k] = -B$, then split $[i_s,i_t]$ into $[i_s,a]$ and $[a+1,i_t]$ where $a$ is the first time point within $[i_s,i_t]$ such that $\bs u_a[k] > -B$. Add $[i_s,a-1]$ to $\mathcal Q$. Set $z \leftarrow a$.
                        
                        \item Let $j$ be the first time in $[z,i_t]$ such that $\bs s_{j-1}[k] = -1$ with $\bs u_j[k] < \bs u_{j-1}[k]$. Add $[z,j-1]$ and $[j,i_t]$ to $\cP$. Goto Step 3.
                    \end{enumerate}
                    
                    \item If there exists one coordinate $k \in [d]$ such that $\bs u_{k}$ is non-monotonic in $[i_s,i_t]$ and $\bs s_{i_s-1}[k], \bs s_{i_t}[k] \in [-1/4,1/4]$ and $\bs \gamma^+_j[k] = 0 \: \forall j \in [i_s,i_t]$ then:
                    \begin{enumerate}
                        \item Initialize $z \leftarrow i_s$. Remove $[i_s,i_t]$ from $\mathcal Q$.
            
                        \item if $\bs u_{i_s}[k] = -B$, then split $[i_s,i_t]$ into $[i_s,a]$ and $[a+1,i_t]$ where $a$ is the first time point within $[i_s,i_t]$ such that $\bs u_a[k] > -B$. Add $[i_s,a]$ to $\mathcal Q$. Set $z \leftarrow a + 1$.
                        
                        \item Let $j$ be the first time in $[z,i_t]$ such that $\bs s_{j-1}[k] = -1$ with $\bs u_j[k] < \bs u_{j-1}[k]$. . Add $[z,j-1]$ and $[j,i_t]$ to $\cP$. Goto Step 3.
                    \end{enumerate}
            
                    \item If there exists one coordinate $k \in [d]$ $\bs u_{k}$ is non-monotonic in $[i_s,i_t]$ and such that $\bs s_{i_s-1}[k] \in [-1,-1/4]$ and $\bs s_{i_t}[k] \in [-1,1/4]$ and $\bs \gamma^+_j[k] = 0 \: \forall j \in [i_s,i_t]$ then:
                    \begin{enumerate}
                        \item Initialize $z \leftarrow i_t$. Remove $[i_s,i_t]$ from $\mathcal Q$.
            
                        \item If $\bs u_{i_t}[k] = -B$, then split $[i_s,i_t]$ into $[i_s,a]$ and $[a+1,i_t]$ where $a$ is the last time point within $[i_s,i_t]$ such that $\bs u_a[k] > -B$. Add $[a+1,i_t]$ to $\mathcal Q$. Set $z \leftarrow a$.        
                        
                        \item Let $j$ be the last time in $[i_s,z]$ such that $\bs s_{j-1}[k] = 1$ with $\bs u_j[k] > \bs u_{j-1}[k]$.. Add $[i_s,j-1]$ and $[j,z]$ to $\cP$. Goto Step 3.
                        
                    \end{enumerate}
            
                    \item If there exists a coordinate $k \in [d]$ such that $\bs u_{k}$ is non-monotonic in $[i_s,i_t]$ and $\bs s_{i_s-1}[k] \in [-1/4,1]$ and $\bs s_{i_t}[k] \in [-1,1/4]$ and $\bs \gamma^+_j[k] = 0, \forall j \in [i_s,i_t]$ then:
                    \begin{enumerate}
                        \item Initialize $p \leftarrow i_s-1$. Remove $[i_s,i_t]$ from $\mathcal Q$.
            
                        \item If $\bs u_{i_s}[k] = -B$, then let  $p$ be the largest point in $[i_s,i_t]$ such that $\bs u_t[k] = -B \: \forall t \in [i_s,p]$. Add $[i_s,p]$ to $\mathcal Q$.
                        
                        \item Let $j$ be the first point in $[p+1,i_t]$ with $\bs s_{j-1}[k] = -1$ with $\bs u_{j-1}[k] > -B$ and $\bs u_{j}[k] < \bs u_{j-1}[k]$. Add $[p+1,j-1]$ to $\cP$.
                        
                        \item If $\bs u_r[k]$ is monotonic in $[j,i_t]$, add  $[j,i_t]$ to $\mathcal Q$. Goto Step 3.
                        
                        \item Initialize $q \leftarrow i_t + 1$.
                        
                        \item If $\bs u_{i_t}[k] = -B$, let $q$ be smallest point in $[j,i_t]$ such that $\bs u_r[k] = -B \: \forall r \in [q,i_t]$. Add $[q,i_t]$ to $\mathcal Q$.
                        
                        \item Let $h$ be the last time point in $[j,q-1]$ such that $\bs s_{h-1}[k] = 1$ with $\bs u_{h}[k] > \bs u_{h-1}[k]$. Add $[j,h-1]$ to $\cP$.
                        
                        \item If $h < q-1$, add $[h,q-1]$ to $\cP$.
                        
                        \item Goto Step 3.
                        
                    \end{enumerate}
            
                    \item If there exists one coordinate $k \in [d]$ such that $\bs s_{i_s-1}[k] \in [0,1]$ and $\bs s_{i_t}[k] \in [-1,-1/4]$ and $\bs \gamma^-_j = 0 \: \forall j \in [i_s,i_t]$, then remove $[i_s,i_t]$ from $\mathcal Q$ and add it to $\cP$. Goto Step 3.        
            
                    \item If there exists one coordinate $k \in [d]$ such that $\bs s_{i_s-1}[k] \in [1/4,1]$ and $\bs s_{i_t}[k] \in [-1/4,0]$ and $\bs \gamma^-_j = 0 \: \forall j \in [i_s,i_t]$, then remove $[i_s,i_t]$ from $\mathcal Q$ and add it to $\cP$. Goto Step 3.
                    
                    \item If there exists one coordinate $k \in [d]$ such that $\bs u_{k}$ is non-monotonic in $[i_s,i_t]$ and $\bs s_{i_s-1}[k], \bs s_{i_t}[k] \in [-1,1/4]$  and $\bs \gamma^-_j=0 \: \forall j \in [i_s,i_t]$ and there exists a coordinate $j \in [i_s,i_t]$ such that $\bs s_{j-1}[k] = 1$ and $\bs u_{j-1}[k] < B$ then:
                    \begin{enumerate}
                        \item Initialize $p \leftarrow i_s$. Remove $[i_s,i_t]$ from $\mathcal Q$.
                        
                        \item If $\bs u_{i_s}[k] = B$, then let  $p$ be the largest point in $[i_s,i_t]$ such that $\bs u_t[k] = B \: \forall t \in [i_s,p]$. Add $[i_s,p]$ to $\mathcal Q$.
                        
                        \item Let $j$ be the first point in $[p+1,i_t]$ such that $\bs s_{j-1}[k] = 1$ with $\bs u_{j-1}[k] < \bs u_{j}[k]$. Add $[p+1,j-1]$ to $\cP$. Add $[j,i_q]$ to $\mathcal Q$. Goto Step 3.
                        
                    \end{enumerate}

                    \item If there exists one coordinate $k \in [d]$ such that $\bs u_{k}$ is non-monotonic in $[i_s,i_t]$ and $\bs s_{i_s-1}[k], \bs s_{i_t}[k] \in [-1/4,1]$ and $\bs \gamma^-_j = 0 \: \forall j \in [i_s,i_t]$ and there exists a $j \in [i_s,i_t]$ such that $\bs u_j[k] - \bs u_{j-1}[k] = -1$ and $u_j[k] < B$ then:
                    \begin{enumerate}
                        \item Initialize $z_1 \leftarrow i_s, z_2 \leftarrow i_t$. Remove $[i_s,i_t]$ from $\mathcal Q$.
                        
                        \item If $\bs u_{i_s}[k] = B$, then let  $p_1$ be the last point in $[i_s,i_t]$ such that $\bs u_t[k] = B \: \forall t \in [i_s,p_1]$. Set $z_1 \leftarrow p_1+1$. Add $[i_s,p]$ to $\mathcal Q$.
                        
                        \item If $\bs u_{i_t}[k] = B$, then let  $p_2$ be the smallest point in $[i_s,i_t]$ such that $\bs u_t[k] = B \: \forall t \in [p_2,i_t]$. Set $z_2 \leftarrow p_2-1$. Add $[p_2,i_t]$ to $\mathcal Q$.
                        
                        \item Let $j$ be the last point in $[z_1,z_2]$ such that $\bs s_{j-1}[k] = -1$ and $\bs u_j[k] < B$ with $\bs u_{j-1}[k] > \bs u_{j}[k]$. Add $[z_1,j-1]$ and $[j,z_2]$ to $\cP$. Goto Step 3.
                        
                    \end{enumerate}
                    
                    \item If there exists one coordinate $k \in [d]$ such that $\bs u_{k}$ is non-monotonic in $[i_s,i_t]$ and $\bs s_{i_s-1}[k] \in [-1,1/4]$ and $\bs s_{i_t}[k] \in [-1/4,1]$ and $\bs \gamma^-_j = 0 \: \forall j \in [i_s,i_t]$ then:
                    \begin{enumerate}
                        \item Initialize $p \leftarrow i_s-1$. Remove $[i_s,i_t]$ from $\mathcal Q$.
                        \item If $\bs u_{i_s}[k] = B$, then let  $p$ be the last time point such that $\bs u_t[k] = B \: \forall t \in [i_s,p]$. Add $[i_s,p]$ to $\mathcal Q$.
                        \item Let $j$ be the first point in $[p+1,i_t]$ such that $\bs s_{j-1}[k] = 1$ with $\bs u_{j-1}[k] < B$ with $\bs u_{j-1}[k] < \bs u_{j}[k]$. Add $[p+1,j-1]$ to $\cP$.
                        
                        \item If $\bs u_r[k]$ is monotonic in $[j,i_t]$, add $[j,i_t]$ to $\mathcal Q$. Goto Step 3.
                        
                        \item Initialize $q \leftarrow i_t + 1$.
                        
                        \item If $\bs u_{i_t}[k] = B$, let $q$ be smallest point in $[j,i_t]$ such that $\bs u_r[k] = B \: \forall r \in [q,i_t]$. Add $[q,i_t]$ to $\mathcal Q$.
                        
                        \item Let $h$ be the last time point in $[j,q-1]$ such that $\bs s_{h-1}[k] = -1$ with $\bs u_{h-1}[k] > \bs u_{h}[k]$. Add $[j,h-1]$ to $\cP$.
                        
                        \item If $h < q-1$, add $[h,q-1]$ to $\cP$.
                        
                        \item Goto Step 3.
                    \end{enumerate}
                    
                \end{enumerate}
                \item[Step 4] Return $\cP$.
            	\end{enumerate}
    
    \end{mdframed}

\begin{lemma} \label{lem:num-bins-ec}
The partitioning routine \code{generateBins} halts. Further we have $|\cP| = O(d n^{1/3}C_n^{2/3})$.
\end{lemma}
\begin{proof}
We need to argue that the loop in Step 3 halts.

FACT1: Notice that in the loop of Step 3,  we add a bin to $\mathcal Q$ only if $\bs u[\tilde k]$ is monotonic in that bin for a coordinate $\tilde k$. Once such a bin is added, in the later steps we do not create new bins across the previous coordinate $\tilde k$. 

FACT2: Step 1 ensures that within each bin we consider, a TV of at-most $B$ will only be incurred. Due to Lemma \ref{lem:ppt-1d}, this TV constraint implies that both $\bs \gamma^+[k]$ and $\bs \gamma^-[k]$ \emph{cannot} be simultaneously non-zero within any bin $[i_s,i_t]$. Consequently we consider all possible configurations of such TV constrained bins in Steps 3(b-m). See Table \ref{table:truth} for a comprehensive summary.

Combining the previous two facts, we conclude that any time point in $[n]$ will be into some bin in $\cP$ or $\mathcal R$ in at-most $d$ (maybe non-consecutive) iterations of the loop in Step 3.

By using similar arguments as in proof of Lemma \ref{lem:part-sq}, we have $|\mathcal Q| = O(n^{1/3}C_n^{2/3})$ after Step 1 gets finished. Due to FACT1, the loop in Step 3 can split a bin that was originally present in $\mathcal Q$ at the end of Step 1 into at-most $O(d)$ sub-bins. Hence  $|\cP|$ can be $O(d n^{1/3}C_n^{2/3})$ after Step 3.
\end{proof}

\begin{figure}[p]
    \centering
    \includegraphics[width=0.8\textwidth]{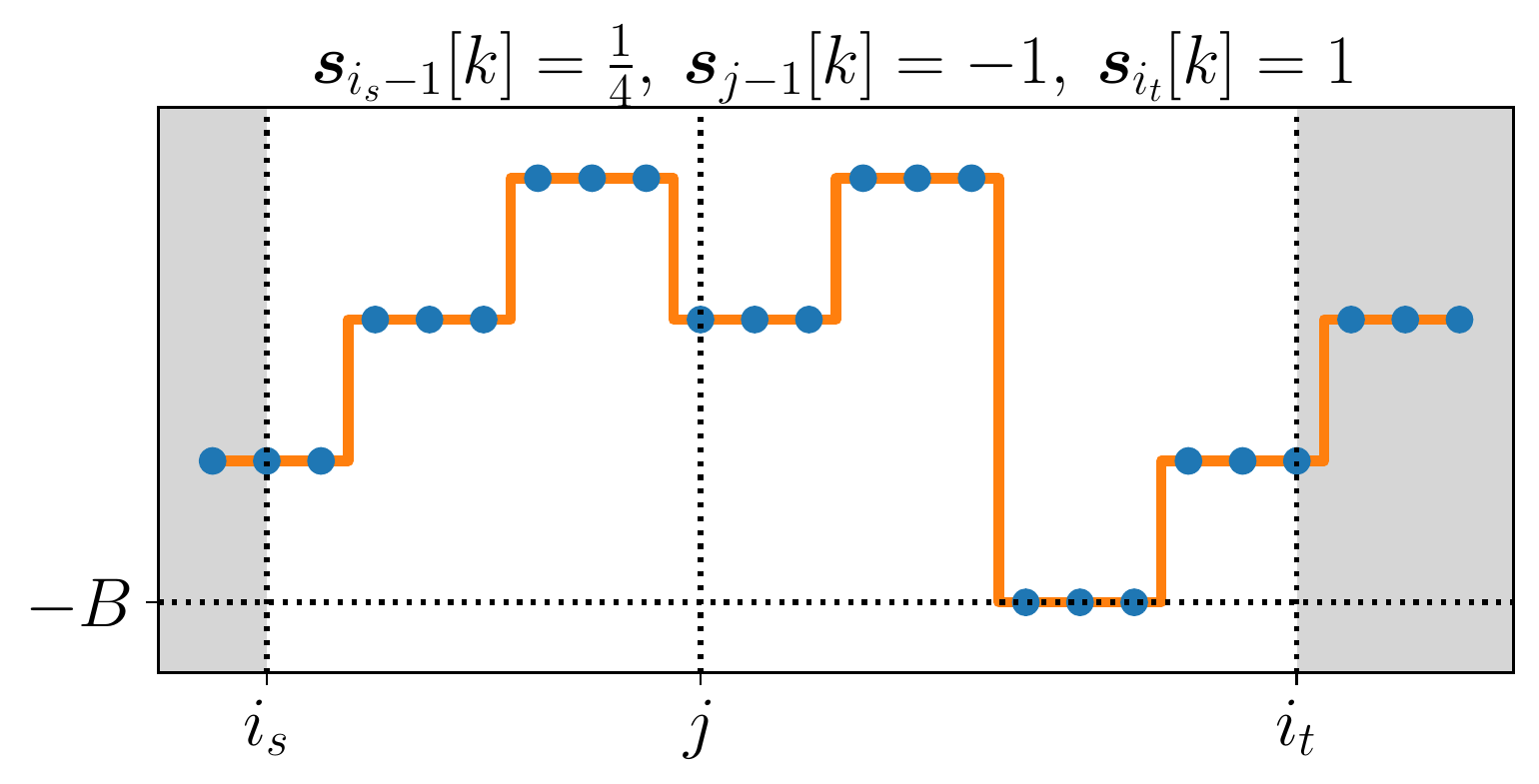}
    \caption{\emph{An example of a configuration corresponding Step 3(e) of \code{generateBins}. Here $z=i_s$.}}    
    \label{fig:ec-e}
        \includegraphics[width=0.8\textwidth]{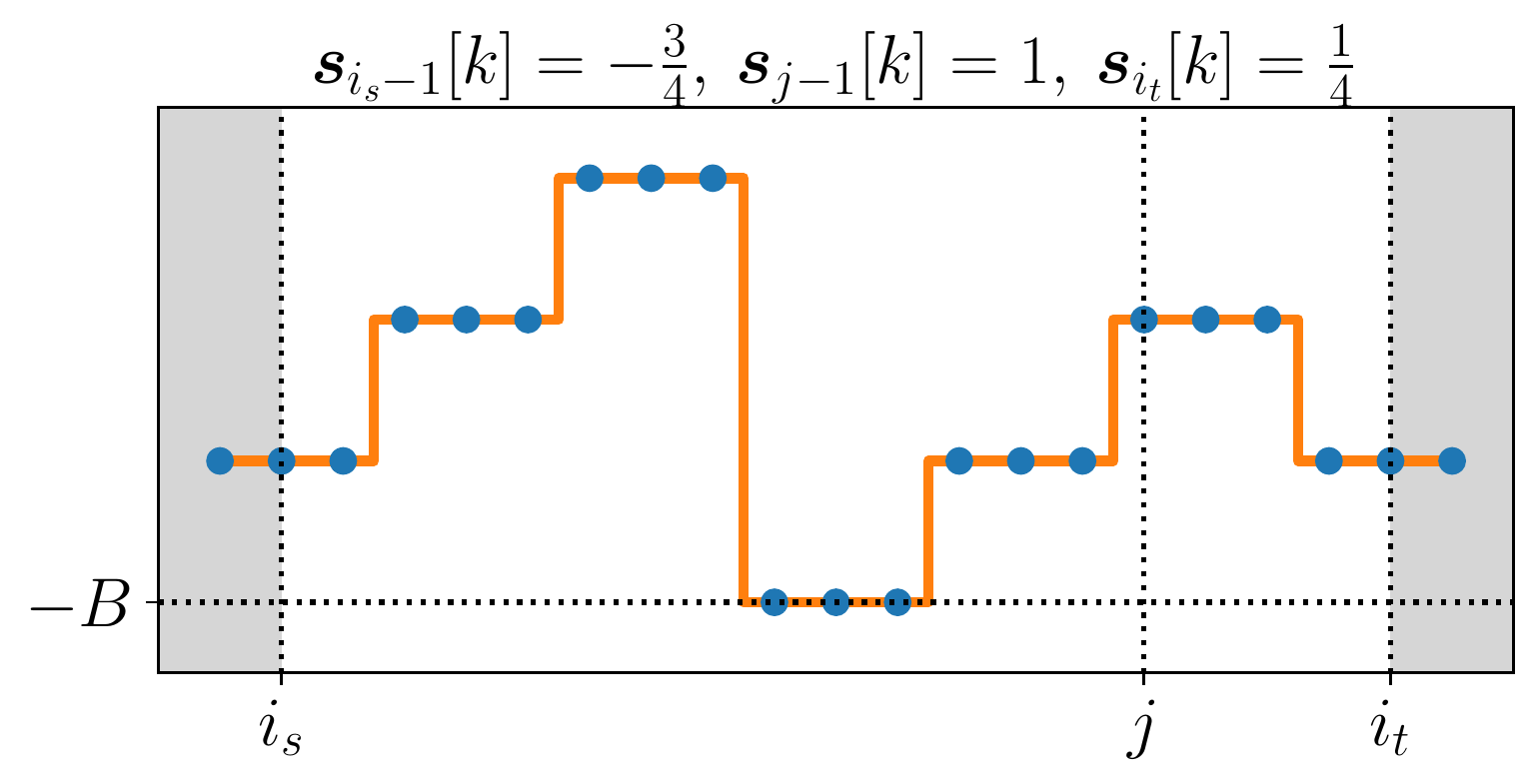}
    \caption{\emph{An example of a configuration corresponding Step 3(g) of \code{generateBins}. Here $z = i_t$.}}    
    \label{fig:ec-g}
        \includegraphics[width=0.8\textwidth]{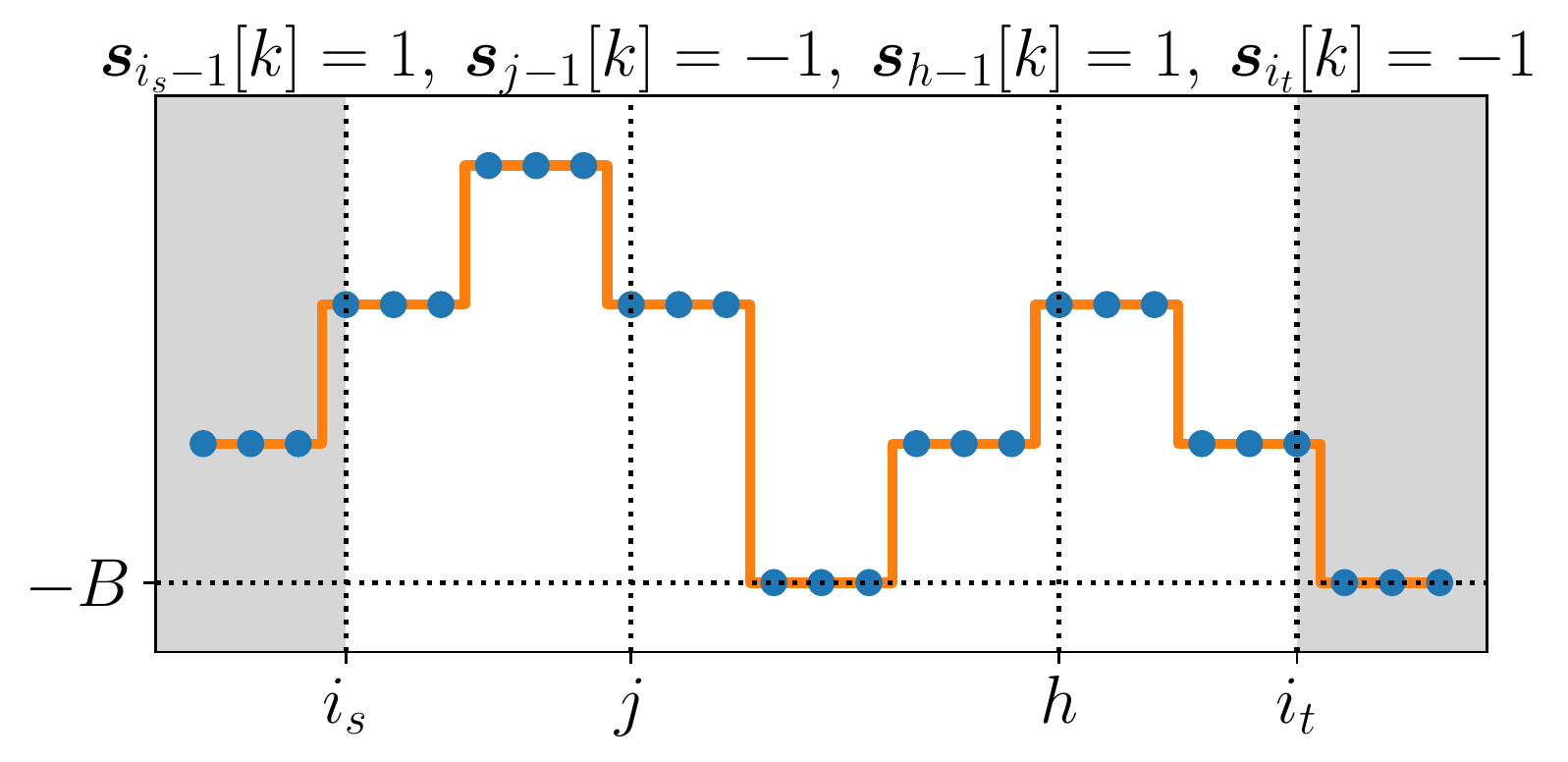}
    \caption{\emph{An example of a configuration corresponding Step 3(h) of \code{generateBins}. Here $p+1 = i_s,\: q = i_t+1$. }}
    \label{fig:ec-h}
\end{figure}

\begin{lemma} \label{lem:props}
Let $\cP$ be the partition produced by \code{generateBins}. Consider a bin $[i_s,i_t] \in \cP$. Using the notations of Lemma \ref{lem:t2-ec-d}, the bin $[i_s,i_t]$ satisfy one of the following properties.
\begin{itemize}
    \item Property 1:  Across each coordinate $k \in [d]$, the sequence $\bs u_j[k], j \in [i_s,i_t]$ is non-decreasing or non-increasing. Or,
    \item Property 2: $ \|\lambda \Delta \bs s_i + \bs \Gamma^-_i - \bs \Gamma^+_i\|_2 \ge \lamda/4$.
\end{itemize}
\end{lemma}
\begin{proof}
To prove the properties satisfied by each bin in $\cP$ we inspect the steps in \code{generateBins} and verify the stated properties. Below when we refer the coordinate $k$, we mean the same coordinate that is used by the corresponding steps in \code{generateBins}. For a bin $[a,b]$, we also recall the notations 
$\Delta \bs s_{a \rightarrow b}$, $\bs \Gamma^+_{a \rightarrow b}$ and $\bs \Gamma^-_{a \rightarrow b}$. We use the short hands $\Delta \bs s_i, \bs \Gamma^+_i, \bs \Gamma^-_i$ as in Lemma \ref{lem:t2-ec-d} for a bin referred by $[i_s,i_t]$.

\begin{enumerate}
    \item For the bin added in Step 3(c) we have $\Delta \bs s_i[k] > 1/4$. Also $\bs \Gamma^-_i[k] - \bs \Gamma^+_i[k] = \bs \Gamma^-_i[k] \ge 0$. Hence Property 2 is verified.
    \item Step 3(d) can be verified as above.
    
    \item In Step 3(e), we add $[z,j-1]$ and $[j,i_t]$ to $\cP$. By construction, the $\bs u[k]$ solution do no attain the value $-B$ in $[z,j-1]$. Hence $\bs \Gamma^-_{z \rightarrow j-1} - \bs \Gamma^+_{z \rightarrow j-1} = 0$. Since $\bs s_{z-1}[k] \in [-1/4,1]$ and $\bs s_{j-1}[k] = -1$, we have $\left|\lamda \Delta \bs s_{z \rightarrow j-1}[k] + \bs \Gamma^-_{z \rightarrow j-1}[k] - \bs \Gamma^+_{z \rightarrow j-1}[k]\right| > \lamda/4$. Hence Property 2 is verified for $[z,j-1]$. For the bin $[j,i_t]$ we have $\bs \gamma^+_r[k] = 0, \: \forall r \in [j,i_t]$. Since $\bs s_{j-1}[k] = -1$ and $\bs s_2[k] \in [-1/4,1]$, we have $\Delta \bs s_{j \rightarrow i_t}[k] \ge 1/4$. So $\lamda \Delta \bs s_{j \rightarrow i_t}[k] + \bs \Gamma^-_{j \rightarrow i_t}[k] - \bs \Gamma^+_{j \rightarrow i_t}[k] \ge  \lamda \Delta \bs s_{j \rightarrow i_t}[k] \ge \lamda/4$. Thus Property 2 is verified for $[j,i_t]$. See Fig. \ref{fig:ec-e} for an example of this configuration.
    
    \item Step 3(f) can be verified using similar arguments as above.
    
    \item In Step 3(g) we add $[i_s,j-1]$ and $[j,z]$ to $\cP$. Since $\bs s_1[k] \in [-1,-1/4]$ and $\bs s_{j-1}[k] = 1$. So we have $\Delta s_{i_s \rightarrow j-1} \ge 0$ and hence Property 2 is satisfied for $[i_s,j-1]$. By construction there $\bs u[k]$ do not attain the value $-B$ in $[j,z]$. So $\bs \Gamma^-_{j \rightarrow z}[k] - \bs \Gamma^+_{j \rightarrow z} = 0$. Since $\bs s_z[k] \in [-1,1/4]$ and $\bs s_{j-1}[k] = 1$, we conclude that Property 2 is satisfied for $[j,z]$. See Fig. \ref{fig:ec-g} for an example of this configuration.
    
    \item By construction of Step 3(h) $\bs \gamma^-_r[k] = 0 \: \forall r \in [p+1,j-1]$. Thus $\bs \Gamma^-_{p+1 \rightarrow j-1}[k] - \bs \Gamma^+_{p+1 \rightarrow j-1}[k] = 0$. Since $\bs  s_{j-1}[k] = -1$ and $\bs s_p[k] \in [-1/4,1]$, we can conclude that Property 2 is verified for $[p+1,j-1]$. Since $\bs s_{h-1}[k] = 1$ we have $\lamda \Delta \bs s_{j \rightarrow h-1}[k] + \bs \Gamma^-_{j \rightarrow h-1}[k] - \bs \Gamma^+_{j \rightarrow h-1}[k] \ge \lamda \Delta \bs s_{j \rightarrow h-1}[k] \ge \lamda/4$. Thus Property 2 is verified for $[j,h-1]$. By construction, $\bs u[k]$ do not attain the value $-B$ in $[h,q-1]$. Hence $\bs \Gamma^-_{h \rightarrow q-1} - \bs \Gamma^+_{h \rightarrow q-1} = 0$. Since $\bs s_{q-1}[k] \in [-1,1/4]$ and  $\bs s_{h-1}[k] = 1$, Property 2 is satisfied for $[h,q-1]$. See Fig. \ref{fig:ec-h} for an example of this configuration.
    
\end{enumerate}
The Properties stated in the Lemma can be verified for all bins that get added to $\cP$ in steps 3(i-m) using similar arguments as above.
\end{proof}

\begin{proof}\textbf{of Lemma} \ref{lem:part-ec-d}.
The proof is completed by the partitioning produced by \code{generateBins} and results of Lemmas \ref{lem:num-bins-ec} and \ref{lem:props}.
\end{proof}

\begin{table}[h]
\centering
\begin{tabular}{|c|c|c|c|}
\hline
$\mathbf{s}_{i_s-1}[k]$ & $\mathbf{s}_{i_t}[k]$ & $\mathbf{\Gamma}^+_i[k] = 0$ & $\mathbf{\Gamma}^-_i[k] = 0$ \\ \hline
{[}-1,-1/4{]}                                  & {[}-1,-1/4{]}                                & (g)                                                                              & (k)                                                                              \\
                                               & {[}-1/4,0{]}                                 & (g)                                                                              & (k)                                                                              \\
                                               & {[}0,1/4{]}                                  & (g)                                                                              & (k)                                                                              \\
                                               & {[}1/4,1{]}                                  & (c)                                                                              & (m)                                                                              \\ \hline
{[}-1/4,0{]}                                  & {[}-1,-1/4{]}                                & (h)                                                                              & (k)                                                                              \\
                                               & {[}-1/4,0{]}                                 & (f)                                                                              & (l)                                                                              \\
                                               & {[}0,1/4{]}                                  & (f)                                                                              & (l)                                                                              \\
                                               & {[}1/4,1{]}                                  & (d)                                                                              & (l)                                                                              \\ \hline
{[}0,1/4{]}                                  & {[}-1,-1/4{]}                                & (h)                                                                              & (k)                                                                              \\
                                               & {[}-1/4,0{]}                                 & (f)                                                                              & (k)                                                                              \\
                                               & {[}0,1/4{]}                                  & (f)                                                                              & (k)                                                                              \\
                                               & {[}1/4,1{]}                                  & (e)                                                                              & (l)                                                                              \\ \hline
{[}1/4,1{]}                                  & {[}-1,-1/4{]}                                & (h)                                                                              & (i)                                                                              \\
                                               & {[}-1/4,0{]}                                 & (h)                                                                              & (j)                                                                              \\
                                               & {[}0,1/4{]}                                  & (h)                                                                              & (l)                                                                              \\
                                               & {[}1/4,1{]}                                  & (e)                                                                              & (l)                                                                              \\ \hline
\end{tabular}
\caption{\emph{Various configurations of a non-monotonic coordinate within a bin $[i_s,i_t]$ and their assignments to the corresponding steps of \code{generateBins} routine for the cases $\bs \gamma^+_j[k] = 0$ for all $j \in [i_s,i_t]$ and $\bs \gamma^-_j[k] = 0$ for all $j \in [i_s,i_t]$.}}
\label{table:truth}
\end{table}
\end{document}